\documentclass[10pt,journal,compsoc]{IEEEtran}
\ifCLASSINFOpdf
\else
\fi

  \usepackage[nocompress]{cite}

  \usepackage{cite}

\usepackage{hyperref}
\newcommand{\RomanNumeralCaps}[1]{\MakeUppercase{\romannumeral #1}}
\usepackage{graphicx}
\usepackage{multirow}
\usepackage{array}

\usepackage[dvipsnames]{xcolor}
\usepackage{mathtools}
\usepackage{amsmath}
\usepackage[nointegrals]{wasysym}
\usepackage{array}
\usepackage{subcaption}
\usepackage{algorithmic}
\usepackage{listings}
\usepackage{footnote}
\usepackage{lipsum}
\usepackage{textgreek}
\usepackage{colortbl}
\usepackage{colortbl,hhline}
\usepackage{xcolor}
\usepackage{placeins}
\usepackage[para,online,flushleft]{threeparttable}
\usepackage{hyperref}
\usepackage{ifmtarg}
\usepackage[]{mdframed}
\usepackage{tikz,lipsum,lmodern}
\usepackage{multicol} 
\usepackage{setspace} 
\usepackage[many]{tcolorbox}
\usepackage{adjustbox}
\usepackage{epstopdf}
\usepackage{nicematrix}
\usepackage{tablefootnote}
\usepackage{cases}
\usepackage{float}
\usepackage{siunitx}
\usepackage[italic,symbolgreek]{mathastext}
\usepackage{textgreek}
\usepackage[printonlyused]{acronym}

\usepackage{pdflscape}
\usepackage{adjustbox}

\usepackage{lipsum}

\hypersetup{
    colorlinks=true,
    linkcolor=blue,
    filecolor=magenta,      
    urlcolor=cyan,
    pdftitle={Overleaf Example},
    pdfpagemode=FullScreen,
    }
\usepackage[ruled,vlined]{algorithm2e}
\graphicspath{{figures/}}
\usepackage{enumitem}
\usepackage{color}
\usepackage{amssymb}
\usepackage{amsfonts}
\usepackage{tabularx}
\usepackage{booktabs}
\usepackage{makecell}
\usepackage{longtable}
\usepackage{cite}
\usepackage{graphicx}
\usepackage{textcomp}
\usepackage{xcolor}
\usepackage{booktabs}
\usepackage{fancybox,framed}
\usepackage{multirow}
\usepackage[utf8]{inputenc}
\usepackage[acronym]{glossaries}
\usepackage{csvsimple}
\usepackage{chngcntr}
\usepackage{mathtools}
\usepackage{breqn}
\usepackage{tabulary}
  \graphicspath{{figures/}}

\usepackage{pifont}
\newcommand{\xmark}{\ding{55}}%
\usepackage{dsfont}
\usepackage[para,online,flushleft]{threeparttable}
\newwrite\malwrite
\openout\malwrite=mal-temp.tex
\newcount\malc \malc=0
\def\malfile{mal-result.tex}

\def\prettyRunes#1{%
  \advance\malc by 1%
  \write\malwrite{#1}%
  \IfFileExists{\malfile}{%
    {\armfamily\mdseries
\par 
{\normalfont Before (input \the\malc): }{\color{blue}#1}\par
{\normalfont After (output \the\malc): }{\color{green}
\input{\malfile}}%
}\par\smallskip
    }
    {}
  }

\makeatletter
\def\algbackskip{\hskip-\ALG@thistlm}
\makeatother

\usepackage{xr}
\makeatletter
\newcommand*{\addFileDependency}[1]{
  \typeout{(#1)}
  \@addtofilelist{#1}
  \IfFileExists{#1}{}{\typeout{No file #1.}}
}
\makeatother

\newcommand*{\myexternaldocument}[1]{
    \externaldocument{#1}
    \addFileDependency{#1.tex}
    \addFileDependency{#1.aux}
}

\myexternaldocument{File2}

\begin{document}

\title{A Review of Deep Learning for Video Captioning}

\author{Moloud~Abdar,
         Meenakshi~Kollati, 
         Swaraja~Kuraparthi,
         Farhad~Pourpanah,~\IEEEmembership{Member,~IEEE,}
         Daniel McDuff,
         Mohammad Ghavamzadeh, 
         Shuicheng Yan,~\IEEEmembership{Fellow,~IEEE,}
         Abduallah Mohamed, 
         Abbas Khosravi,~\IEEEmembership{Senior Member,~IEEE,}
         Erik~Cambria,~\IEEEmembership{Fellow,~IEEE,} Fatih Porikli,~\IEEEmembership{Fellow,~IEEE}

\IEEEcompsocitemizethanks{
\IEEEcompsocthanksitem This work was partially supported by the Australian Research Council’s Discovery Projects funding scheme (project DP190102181 and DP210101465).\protect 
\IEEEcompsocthanksitem M. Abdar and A. Khosravi are with the Institute for Intelligent Systems Research and Innovation (IISRI), Deakin University, Australia (e-mails: m.abdar1987@gmail.com \& m.abdar@deakin.edu.au \& abbas.khosravi@deakin.edu.au).\protect
\IEEEcompsocthanksitem M. Kollati and S. Kuraparthi are with the Department of Electronics and Communication Engineering, Gokaraju Rangaraju Institute of Engineering \& Technology, Hyderabad, Telangana, India (e-mails: mkollati@gmail.com \& kswarajai@gmail.com).\protect
\IEEEcompsocthanksitem F. Pourpanah is with the Department of Electrical and Computer Engineering, and Ingenuity Labs Research Institute, Queen’s University, Kingston, ON, Canada (e-mail: farhad.086@gmail.com).\protect
\IEEEcompsocthanksitem D. McDuff is with the University of Washington in Seattle, WA, USA (e-mail: dmcduff@uw.edu).\protect
\IEEEcompsocthanksitem M. Ghavamzadeh is with Google Research, Mountain View, CA, USA (e-mail: ghavamza@google.com).\protect
\IEEEcompsocthanksitem S. Yan is with the Sea AI Lab., Singapore (e-mail: shuicheng.yan@gmail.com).\protect
\IEEEcompsocthanksitem A. Mohamed is with the Meta Reality Labs in Seattle, WA, USA (e-mail: abduallahadel@meta.com).\protect
\IEEEcompsocthanksitem E. Cambria is with the School of Computer Science and Engineering, Nanyang Technological University, Singapore (E-mail: cambria@ntu.edu.sg).\protect
\IEEEcompsocthanksitem F. Porikli is with Qualcomm AI Research, USA (E-mail: fporikli@qti.qualcomm.com).\protect
}

\markboth{IEEE TRANSACTIONS ON }%
{Moloud Abdar\MakeLowercase{\textit{et al.}}: }

}


\IEEEtitleabstractindextext{
\begin{abstract}
Video captioning (\ac{VC}) is a fast moving, cross-disciplinary area of research that bridges work in the fields of computer vision, natural language processing (\ac{NLP}), linguistics and human-computer interaction. In essence, \ac{VC} involves understanding a video and describing it with language. Captioning is used in a host of applications from creating more accessible interfaces (e.g., low-vision navigation) to video question answering (\ac{V-QA}), video retrieval and content generation. This survey covers deep learning-based \ac{VC}, including but, not limited to, attention-based architectures, graph networks, reinforcement learning, adversarial networks, dense video captioning (\ac{DVC}), and more. We discuss the datasets and evaluation metrics used in the field, and limitations, applications, challenges, and future directions for \ac{VC}.
\end{abstract}
}

\maketitle

\IEEEdisplaynontitleabstractindextext

\IEEEpeerreviewmaketitle








%

\section{Introduction} \label{sec:introduction}
Automated descriptions of visual content can be very useful, from creating more accessible content to providing efficient methods of searching and indexing large media corpora to helping generate novel media. Image and video captioning~\cite{vinyals2016show, you2016image, sharma2018conceptual, gu2017non, guo2020non} is the task of generating textual descriptions of visual content. While describing what we see is a very natural task for humans, it is not trivial to create artificially intelligent algorithms that do the same. Researchers in the fields of computer vision (CV) and natural language processing (NLP)~\cite{gandhi2022multimodal} have not yet been able to completely solve the challenge of converting low-level visual features to higher-level semantic or symbolic abstractions~\cite{xiao2020video1}. 
\ac{VC} is more complex than image captioning, not only because videos contain many frames, thus carrying significantly more information than a still image, but also because it is necessary to extract information from the temporal dimension as well as the spatial dimension. Furthermore, videos are often accompanied by audio, meaning that multimodal understanding might be necessary to interpret the content correctly.

With the advent of the internet and online media platforms (e.g., YouTube, Facebook, TikTok), there has been an explosion of content created and shared. Videos are often associated with metadata in the form of simple tags or complete paragraphs. These data provide opportunities for improving automatic descriptions of visual content. The availability of huge datasets (e.g., Activitynet, \ac{MPII-MD} and \ac{S-MiT}) has boosted research in computer vision, including work on Dense Video Captioning (\ac{DVC}), Video Question Answering (\ac{V-QA}),~\ac{CBVR}~\cite{zhao2021pyramid}, and paragraph captioning~\cite{lei2020mart, song2021towards}. The qualitative and quantitative performance of \ac{VC} systems has improved dramatically since the advent of deep learning (DL)~\cite{lecun2015deep, wang2020recent, pourpanah2022review}. Neural models scale well with data and large parameter models can be successfully trained without overfitting.

\ac{VC} has applications across a plethora of fields, from \ac{AI}-assisted medical diagnostics to storytelling through videos to \ac{V-QA}~\cite{zeng2017leveraging, lei2018tvqa} to lip-reading to scene understanding and autopilot assistance~\cite{chen2021memory}. While \ac{VC} might appear a purely digital task it has implications for robotic systems that physically interact with humans and comprehend their surroundings~\cite{song2018self, xiong2018move}. A number of fields have leveraged and contributed to the literature on \ac{VC}, in this review we cover work in the domains of \ac{HRI}, video indexing, video tagging, video description, video recommendation systems, video summarization, visual information retrieval, and accessibility~\cite{chen2016video}.


\begin{figure}[ht]
\centering
    \includegraphics[width=0.485\textwidth]{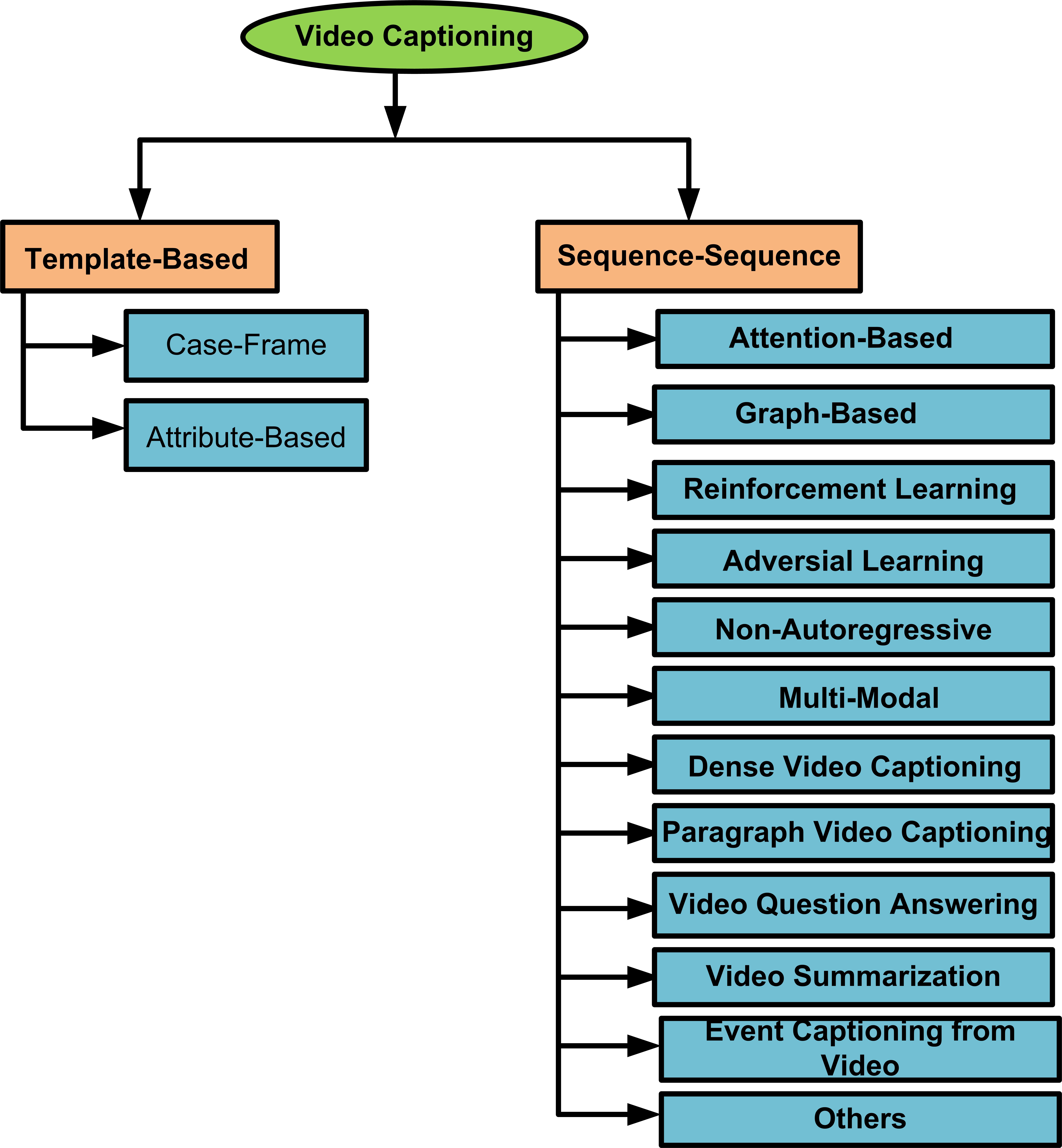}
    \caption{A taxonomy of deep-learning based \ac{VC} methods. }
   \label{fig:capt}
\end{figure}

\begin{figure}[ht]
\centering
    \includegraphics[width=0.49\textwidth]{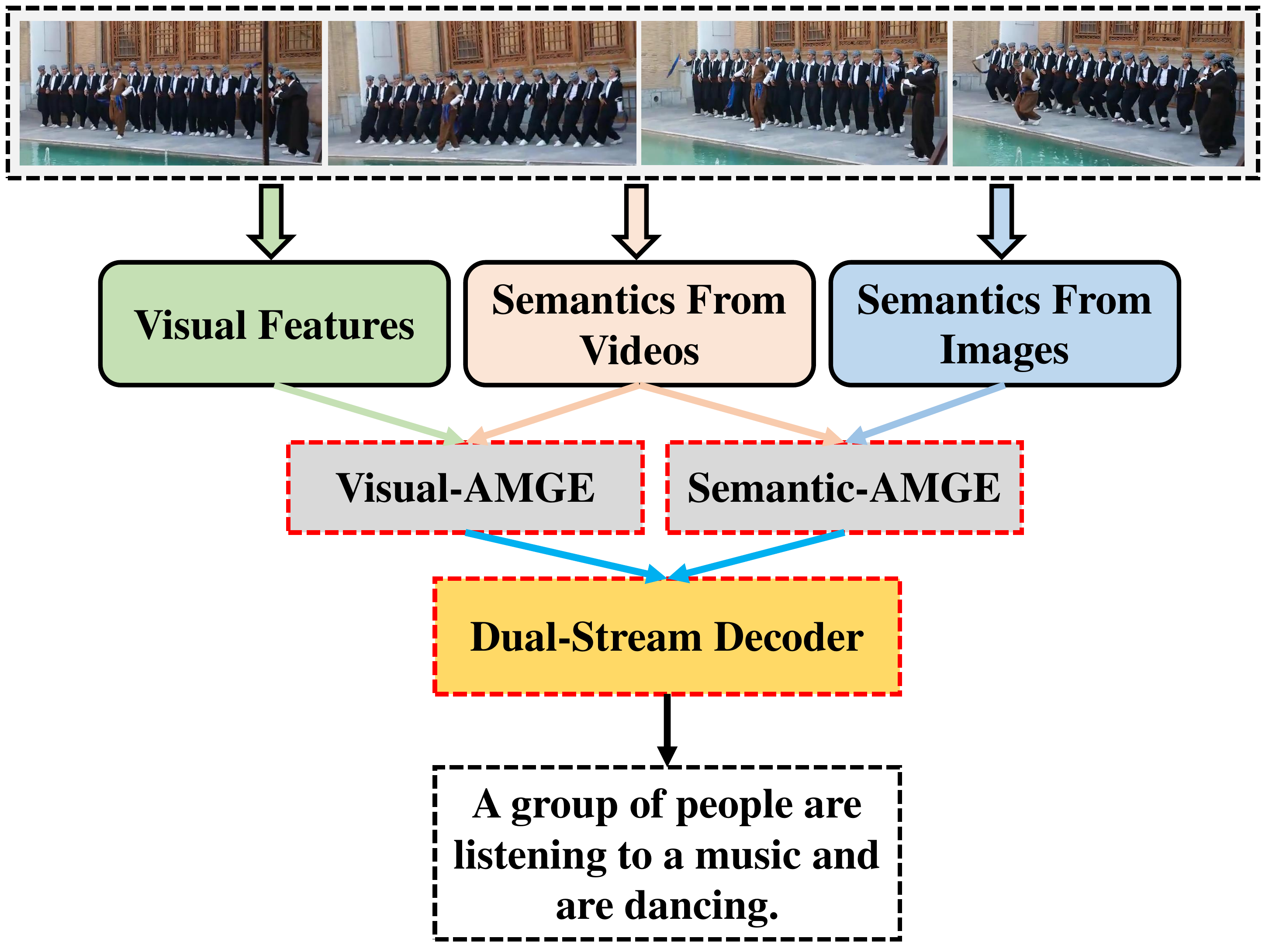}
    \caption{DS-\ac{RNN} for \ac{VC} with Attentive Multi-Grained
Encoder (AMGE) module~\cite{xu2018dual}.}
   \label{fig:DS-RNN}
\end{figure}

 Early \ac{VC} schemes used template-based approaches~\cite{kojima2002natural, babariya2019meaning,guadarrama2013youtube2text,krishnamoorthy2013generating,thomason2014integrating}.
Template methods, extract semantic concepts such as a subject, object, and verb from a set of visual classifiers, and then generate captions through a language model which assigns predicted triplets to predefined sentence templates.
Rohrbach et al.~\cite{rohrbach2013translating} proposed a Conditional Random Field (CRF) to learn the relationship between different components in the source video. They found it was possible to generate a rich semantic representation of the visual content, including objects and activity labels. Some of the drawbacks of template-based methods are that they are inflexible and are poor in modeling the diversity and expressiveness of language when the vocabulary is large~\cite{khurana2021video}. Empirical results show they are not good at generalizing outside the training set, and thus, have limited capacity in handling unseen data. Furthermore, as the visual classifiers are not trained with the language model they are not able to learn concepts end-to-end from low-level image features.\par
\begin{figure*}
\centering
        \begin{subfigure}[b]{0.46\textwidth}
            \centering
            \includegraphics[width=\textwidth]{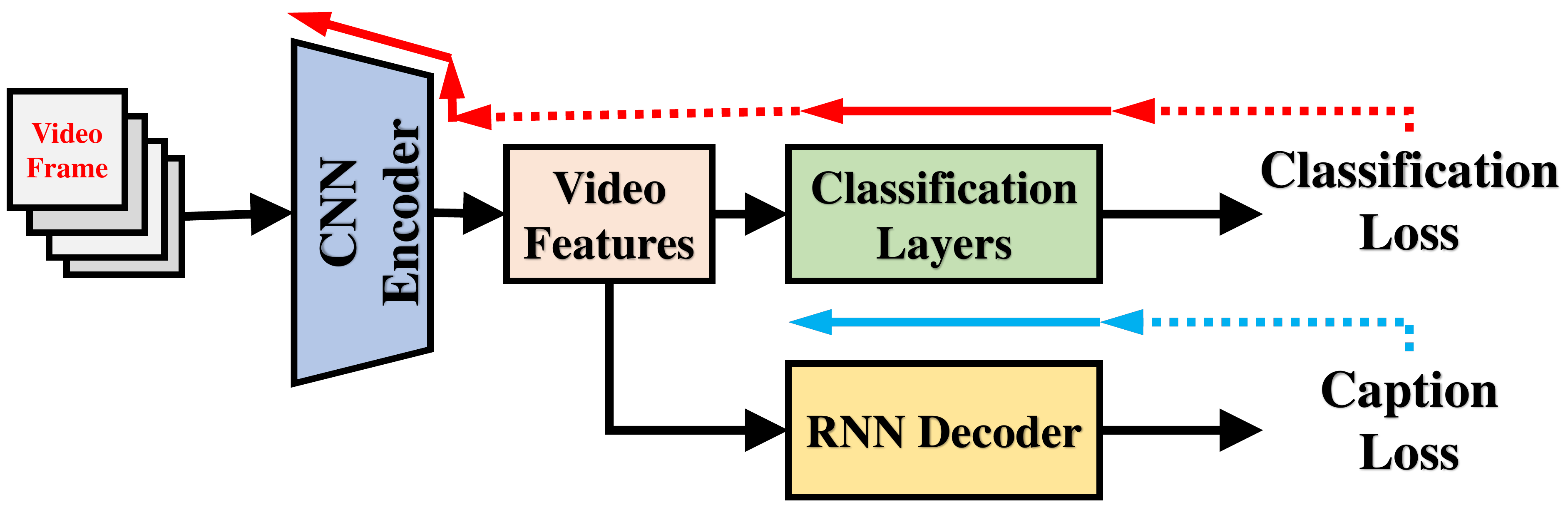} 
            \caption{Disjoint training}
            \label{fig:ETEA}
    \end{subfigure} 
     \begin{subfigure}[b]{0.46\textwidth}
            \centering
            \includegraphics[width=\textwidth]{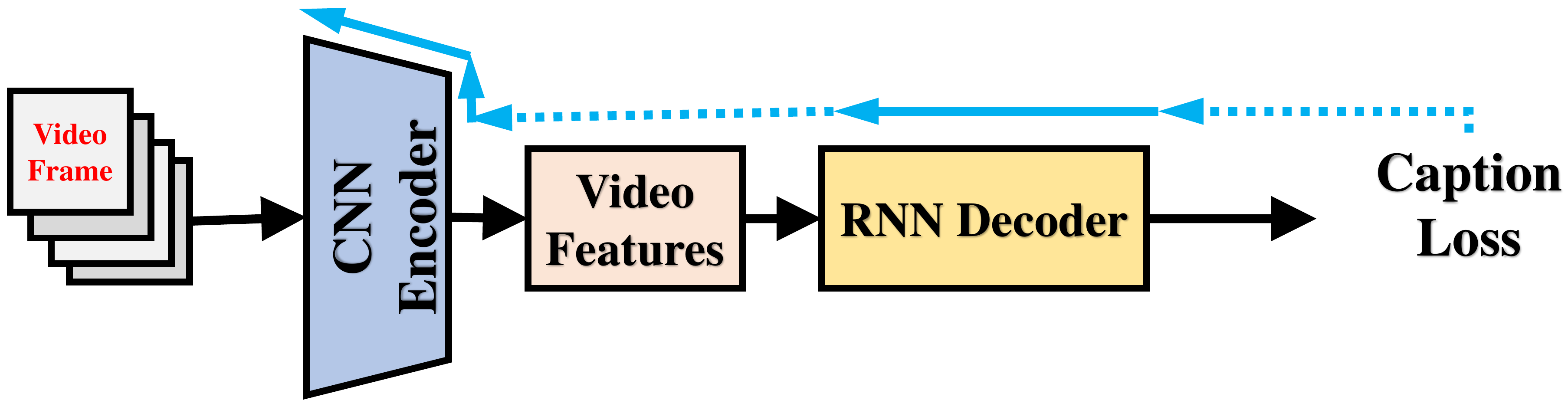}
            \caption{End-to-end training}
            \label{fig:ETEB}
    \end{subfigure} 
    \caption{Disjoint training vs end-to-end training methods for \ac{VC}~\cite{olivastri2019end}.}\label{fig:ETETRR}
\end{figure*}

In deep learning-based \ac{VC}, encoder-decoder pipelines are frequently employed~\cite{gao2020fused}. In these models, the encoder extracts visual features, for example, using a \ac{2D-CNN}~\cite{venugopalan2014translating} or linear embedding. The feature vector can be further reduced via \ac{MP}, a \ac{TE}, or via \ac{SEM}, before being fed to the decoder~\cite{aafaq2021empirical}. Traditionally, a \ac{CNN} is used in the encoder (video), whereas a \ac{RNN} is employed as the decoding phase (for language generation)~\cite{olivastri2019end}. However, as can be seen from Fig.~\ref{fig:ETETRR}, newly proposed state-of-the-art methods train the encoder and decoder parts separately.

Encoder-decoder frameworks that employ \ac{3D-CNN} layers typically better capture spatio-temporal information than a model that applies a \ac{2D-CNN} to each frame separately. Recurrent Neural Networks (\ac{RNN}) or Dual-Stream \ac{RNN}s (DS-\ac{RNN} shown in Fig.~\ref{fig:DS-RNN}) can be utilized in the encoding stage and perform a similar function (see~\cite{xu2018dual}). Language decoders utilize pre-trained word embeddings such as Word2Vec, GloVe, and fastText to convert semantically similar words into similar vector representations based on the feature encoding. Since language is sequential an \ac{RNN} is then used for decoding an output sequence~\cite{goyal2018deep}.
In recent years, Dynamic Video Captioning (\ac{DVC}) has emerged to deal with captioning real-time videos. "Live" video captioning typically involves the analysis of longer, unsegmented videos. Videos are often composed of a variety of events with some elements that are irrelevant to the desired caption. To address \ac{DVC}~\cite{deng2021sketch, chang2022event,aafaq2022dense}, systems have been designed to describe events using story-telling techniques~\cite{deng2021sketch}, which are useful in applications such as \ac{CBIR} and video recommendation~\cite{pu2020multimodal}.\par
Recently, several review papers~\cite{islam2021exploring, jain2022video, moctezuma2022video} have covered \ac{VC} developments in the period between 2020 and 2022. Islam et al.~\cite{islam2021exploring} produced a survey on the state-of-the-art deep learning based \ac{VC} approaches, evaluating the various benchmark datasets and summarizing the evaluation metrics. Jain et al.~\cite{jain2022video} surveyed \ac{VC} by compiling, studying, and summarizing the results of the existing \ac{VC} techniques in the literature. Moctezuma et al.~\cite{moctezuma2022video} presented another review paper on a subset of \ac{VC} papers published between 2015 and 2022, and discussed recent evaluation metrics and datasets. Compared to the other review papers~\cite{islam2021exploring,jain2022video, moctezuma2022video}, our work presents a thorough review of methods in \ac{VC}, evaluates recent techniques, datasets, and evaluation metrics. We cover relevant papers on \ac{VC} and related topics from 2015 to 2022, and discuss the main research gaps existing in the field and within related subfields. Table~\ref{Comp} presents a comprehensive comparison of our review paper on \ac{VC} and the existing review papers in the literature.\par
\color{black}
\begin{table*}[] 
\centering
\caption{Comparison between our paper and other recent surveys on \ac{VC}. It should be noted that this review focuses on \ac{VC} only and not image captioning. } \label{Comp}
\scriptsize
\begin{adjustbox}{width=0.999\textwidth}
\begin{tabular}{l l c c c c c c c}
\toprule
&  & \multicolumn{7}{c}{Study}                      \\ \cmidrule{3-9} 
&  & 
Ours (2023) &\multicolumn{1}{c}{\makecell{Khurana and Deshpande \\~\cite{khurana2021video} (2021)}}& \multicolumn{1}{c}{\makecell{Islam et al. \\ 
\cite{islam2021exploring} (2021)}}  & \multicolumn{1}{c}
{\makecell{Jain et al. \\~\cite{jain2022video} (2022)}}  & \multicolumn{1}{c}{\makecell{Li et al. \\~\cite{li2019visual} (2019)}} & \multicolumn{1}{c}{\makecell{Aafaq et al. \\~\cite{aafaq2019video} (2019)}} & \multicolumn{1}{c}
{\makecell{Moctezuma et al. \\~\cite{moctezuma2022video} (2022)}} \\ \midrule  
\multicolumn{1}{l}{\multirow{11}{*}{Techniques}} & Template-based   & \checkmark  & \multicolumn{1}{c}{\checkmark}         & \multicolumn{1}{c}{\checkmark} & \multicolumn{1}{c}{\checkmark} & \multicolumn{1}{c}{\checkmark}& \multicolumn{1}{c}{\checkmark} &  \multicolumn{1}{c}{\xmark}  \\ \cmidrule{2-9} 
\multicolumn{1}{l}{} & Attention-based     & \checkmark  & \multicolumn{1}{c}{\checkmark}  & \multicolumn{1}{c}{\checkmark} & \multicolumn{1}{c}{\checkmark} & \multicolumn{1}{c}{\checkmark}& \multicolumn{1}{c}{\checkmark} & \multicolumn{1}{c}{\checkmark}   \\ \cmidrule{2-9} 
\multicolumn{1}{l}{}  & Reinforcement learning     & \checkmark  & \multicolumn{1}{c}{\checkmark}         & \multicolumn{1}{c}{\checkmark} & \multicolumn{1}{c}{\checkmark} &\multicolumn{1}{c}{\checkmark}& \multicolumn{1}{c}{\checkmark} & \multicolumn{1}{c}{\checkmark}   \\ \cmidrule{2-9} 
\multicolumn{1}{l}{} & Adversarial learning & \checkmark  &\multicolumn{1}{c}{\xmark}         & \multicolumn{1}{c}{\xmark} & \multicolumn{1}{c}{\xmark} & \multicolumn{1}{c}{\xmark}& \multicolumn{1}{c}{\xmark} & \multicolumn{1}{c}{\checkmark}    \\ \cmidrule{2-9} 
\multicolumn{1}{l}{}   & Graph-based & \checkmark  &\multicolumn{1}{c}{\checkmark}         &\multicolumn{1}{c}{\checkmark} & \multicolumn{1}{c}{\xmark} &\multicolumn{1}{c}{\xmark}& \multicolumn{1}{c}{\xmark} & \multicolumn{1}{c}{\xmark} \\ \cmidrule{2-9} 
\multicolumn{1}{l}{}   & Non-autoregressive  & \checkmark  &\multicolumn{1}{c}{\xmark}         &\multicolumn{1}{c}{\xmark} & \multicolumn{1}{c}{\xmark} & \multicolumn{1}{c}{\xmark}& \multicolumn{1}{c}{\xmark} & \multicolumn{1}{c}{\xmark}   \\ \cmidrule{2-9} 
\multicolumn{1}{l}{}   & \ac{DVC} &\multicolumn{1}{c}{\checkmark} & \multicolumn{1}{c}{\checkmark} & \multicolumn{1}{c}{\checkmark} & \multicolumn{1}{c}{\xmark}& \multicolumn{1}{c}{\checkmark} & \multicolumn{1}{c}{\checkmark} & \checkmark   \\ \cmidrule{2-9} 
\multicolumn{1}{l}{} & \ac{V-QA}  & \checkmark  &\multicolumn{1}{c}{\checkmark}         &\multicolumn{1}{c}{\checkmark} & \multicolumn{1}{c}{\xmark} & \multicolumn{1}{c}{\xmark}& \multicolumn{1}{c}{\xmark} & \multicolumn{1}{c}{\xmark}    \\ \cmidrule{2-9} 
\multicolumn{1}{l}{}  & Video summarization  & \checkmark  &\multicolumn{1}{c}{\xmark}         &\multicolumn{1}{c}{\xmark} & \multicolumn{1}{c}{\xmark} & \multicolumn{1}{c}{\checkmark}& \multicolumn{1}{c}{\xmark} &   \multicolumn{1}{c}{\xmark}  \\ \cmidrule{2-9} 
\multicolumn{1}{l}{} & Paragraph \ac{VC} & \checkmark  &\multicolumn{1}{c}{\xmark}         & \multicolumn{1}{c}{\xmark} & \multicolumn{1}{c}{\xmark} & \multicolumn{1}{c}{ \checkmark}& \multicolumn{1}{c}{\xmark}  &  \multicolumn{1}{c}{\xmark}  \\ \cmidrule{2-9} 
\multicolumn{1}{l}{}   &  Event captioning from video & \checkmark  &\multicolumn{1}{c}{\checkmark}  & \multicolumn{1}{c}{\checkmark} & \multicolumn{1}{c}{\xmark} & \multicolumn{1}{c}{\xmark}& \multicolumn{1}{c}{\xmark}  & \multicolumn{1}{c}{\checkmark}  \\ \midrule
\multicolumn{1}{l}{Datasets}         &\multicolumn{1}{c}{}           & \checkmark    &\multicolumn{1}{c}{\checkmark}         & \multicolumn{1}{c}{\checkmark} & \multicolumn{1}{c}{\checkmark} & \multicolumn{1}{c}{\checkmark}& \multicolumn{1}{c}{\checkmark} & \multicolumn{1}{c}{\checkmark}    \\ \midrule
\multicolumn{1}{l}{Applications}         &\multicolumn{1}{c}{}       & \checkmark    &\multicolumn{1}{c}{\checkmark}         & \multicolumn{1}{c}{\checkmark} & \multicolumn{1}{c}{\xmark} & \multicolumn{1}{c}{\xmark} & \multicolumn{1}{c}{\xmark}  & \multicolumn{1}{c}{\checkmark}  \\ \midrule
\multicolumn{1}{l}{Technical Research Gaps}  &\multicolumn{1}{c}{}   & \checkmark     &\multicolumn{1}{c}{\xmark}         & \multicolumn{1}{c}{\checkmark} & \multicolumn{1}{c}{\xmark} & \multicolumn{1}{c}{\checkmark}& \multicolumn{1}{c}{\checkmark} & \multicolumn{1}{c}{\xmark}   \\ \midrule
\multicolumn{1}{l}{Evaluation Metrics}     &\multicolumn{1}{c}{}     & \checkmark      &\multicolumn{1}{c}{\xmark}         & \multicolumn{1}{c}{\xmark} & \multicolumn{1}{c}{\xmark} & \multicolumn{1}{c}{\checkmark}& \multicolumn{1}{c}{\checkmark} & \multicolumn{1}{c}{\checkmark}  \\ \bottomrule
\end{tabular}
\end{adjustbox}
\end{table*}

\begin{figure*}[ht]
\centering
    \includegraphics[width=0.98\textwidth]{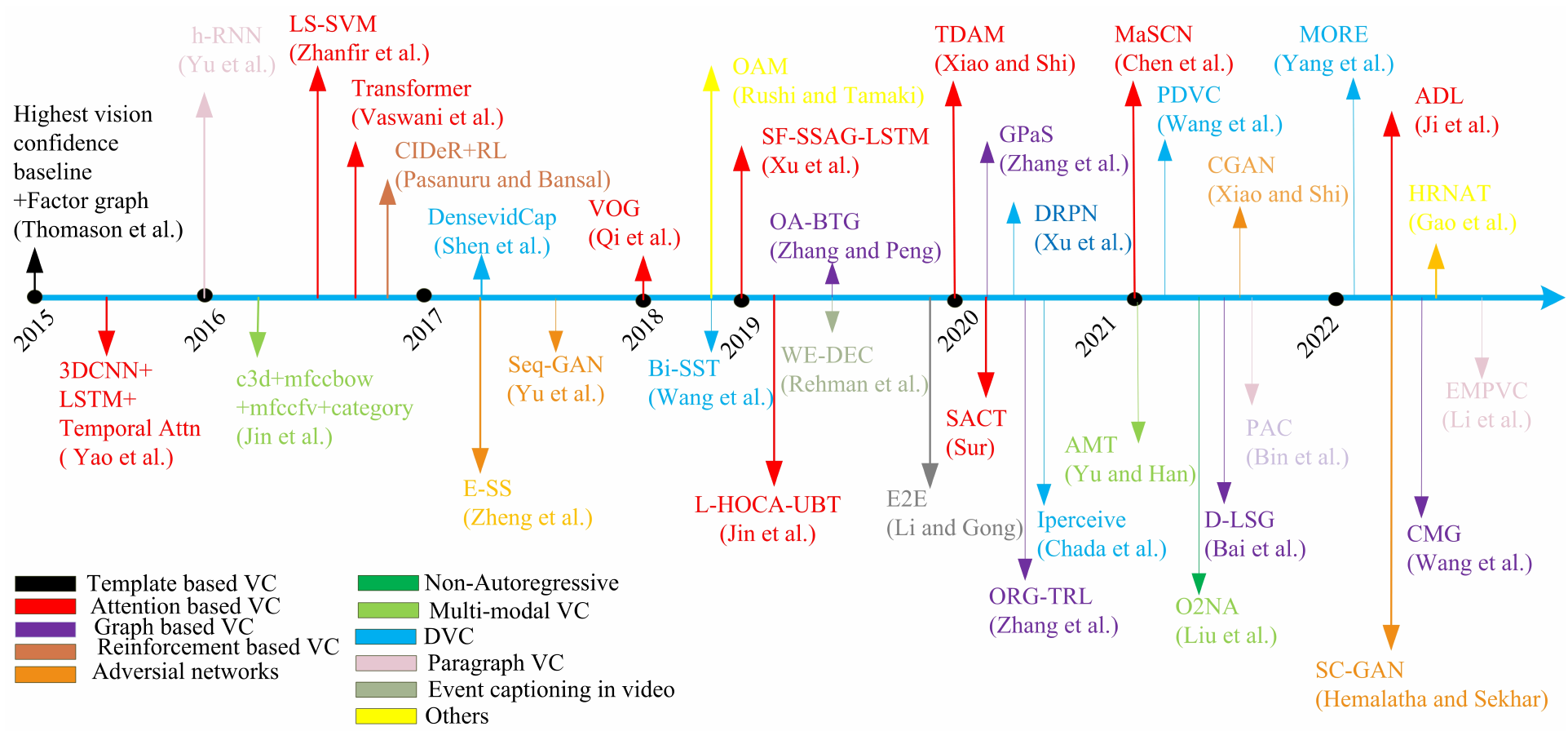}
   \caption{A chronological overview of the most relevant deep learning-based \ac{VC} methods.}
    \label{fig:TAXO}
\end{figure*}


\subsection{Main Contributions}
\label{Sec:sec:LRCON}
The major contributions of this review are as follows:\par
\begin{enumerate}
\item Provide the most comprehensive survey to date of deep learning methods for \ac{VC}, \ac{DVC}, \ac{T2I}, and \ac{T2V}, incorporating the recent advances in deep learning and highlighting their applications. 
\item Provide a thorough comparison of the existing methods on several publicly available datasets, with captioning evaluated based on parameters discussed in Section~\ref{Sec:sec:ECORRE}.
\item Unlike other reviews, we cover everything from template-based, attention-based, graphical, reinforcement learning, adversarial, Dense Video Captioning, and non-autoregressive \ac{VC} techniques to V-QA, video summarization, paragraph \ac{VC}, and event captioning from video (see Table~\ref{fig:TAXO}). 
\item For each explicit task, the contemporary landscape of deep learning is summarized, and the technical milestones are highlighted.
\item We conclude our review with discussions on open research challenges and opportunities for future studies.
 \end{enumerate}

The remainder of this survey has the following structure. We first cover the foundations of deep learning-based \ac{VC}; including datasets, evaluation metrics, and optimization (e.g., loss) formulations (Section~\ref{Sec:PLE}). We then categorize \ac{VC} methods (Section~\ref{Sec:PM}) and present a discussion on the applications of \ac{VC} (Section~\ref{Sec:App}). The main results, research gaps, and future research directions of \ac{VC} methods are presented towards the end (Section~\ref{Sec:Diss}). Finally, we conclude the review in Section~\ref{Sec:Co}.





\section{Deep Learning Video Captioning Foundations}
\label{Sec:PLE}

The main goal of \ac{VC} is to find/generate a sentence (or a sequence of words) for a given video. Assume a video sequence $\mathcal{V}=\{\mathbf{v}_1, \mathbf{v}_2, \ldots, \mathbf{v}_n\}$ with $n$ frames, where $\mathbf{v}_i \in \mathbb{R}^{w\times h \times c}$ represents the $i^{th}$ video, and $w$, $h$, and $c$ indicate width, height and the number of channels, respectively. The goal is to learn a model \textbf{\normalfont$\varPhi$} to map the frame sequences into the word vector space, i.e., $\varPhi: \mathcal{V} \to \mathcal{Y}$, where $\mathcal{Y}=\{\mathbf{y}_1, \cdots, \mathbf{y}_m\}$ consists of $m$ words, $\mathbf{y}\in \mathbf{R}^N$ being a one-hot vector and $N$ indicating the size of the word vocabulary. \ac{VC} can be considered as a sequence-to-sequence learning task, or similarly as an encoding of video features and decoding of text. \ac{VC} mainly is composed of an encoder and decoder. The encoder deals with the multi-modality of the input data video and audio. The decoder uses the embedding and generate the caption. Below, we explore the different options of both then encoders and decoders.

\subsection{Encoders}

Encoders in \ac{VC} often involve extracting features from the input video frames via a pre-trained \ac{CNN}. The extracted features are then utilized as input to an \ac{RNN} decoder to generate the caption. The choice of CNNs has a profound impact on the overall performance of the video captioning model. Popular pre-trained model architectures are C3D~\cite{wang2019self}, VGG-16~\cite{caron2019unsupervised}, VGG-19~\cite{jaworek2019melanoma}, Inception-v3~\cite{xie2018pre}, and Inception ResNet-v2~\cite{szegedy2017inception}. Videos comprise segments with boundaries; each segment may contain multiple scenes, actions, and objects. The segments vary from frame to frame. The segments dictate the underlying structure of the video and knowledge of the structure is important in enhancing the network's understanding of the video~\cite{xu2019semantic}. Most of the prior work has used features solely from video frames. However, multi-modal information such as audio, plays a crucial role in differentiating videos from a mere sequence of images. Extracting multi-modal information, can result in better feature representations, which in-turn help to improve the accuracy of the video captions. Multi-modal encoding is not trivial in part due to the challenge of combining latent representations from different modalities (audio, speech, video, image, and text sources)~\cite{deng2018deep}.

\subsection{Decoders}

The main problem with \ac{RNN}s used in the decoder is that they suffer from vanishing gradients. In most of the \ac{VC} algorithms, different types of \ac{RNN}s are used, such as \ac{LSTM}~\cite{jin2019low,nakamura2021sensor,mohanta2021boosting} networks and~\ac{GRU}. \ac{LSTM}s help overcome the vanishing gradient problem~\cite{hochreiter1997long} and have shown promising results in dealing with sequences of frames, learning long-range temporal patterns~\cite{goyal2018deep}. \ac{GRU}s are simple and require fewer parameters for training. They are mainly used for modeling short sequences and are less likely to result over-fitting~\cite{shi2020video}.

The accuracy of sequence-sequence \ac{VC} models can be enhanced using pre-trained weights~\cite{ramachandran2016unsupervised} and by introducing additional \ac{LSTM} layers.
Recently, transformers~\cite{dosovitskiy2020image, touvron2021training, yang2020learning} have been employed in \ac{VC} to capture long-range dependencies.  The use of transformers~\cite{vaswani2017attention} can accelerate training speed. Given queries $q$ $\in$ $\mathbb{R}^{t_{q} \times d_{k}}$, keys $k$ $\in$ $\mathbb{R}^{t_{v} \times d_{k}}$ and values $v$ $\in$ $\mathbb{R}^{t_{v} \times d_{v}}$. The set of queries, keys and values are represented in matrices--$q$, $k$ and $v$. In this equation, $t_{q}$ is the length of the queries and $t_{v}$ is the length of the keys and values. $d_{k}$ is the dimension of $q$ and k and $d_{v}$ is the dimension of $v$ and $\sqrt{d}$ is the scaling factor. The attention output is
\begin{equation}\label{eqq3}
f =A\left(q,k,v \right) =\text{softmax}\left(\dfrac{qk^{T}}{\sqrt{d}} \right) v.
\end{equation}

In \ac{MHA}~\cite{jin2020sbat} mechanisms, multiple heads are used, where the attention from all the heads are computed in parallel. \ac{MHA} facilitates the network obtaining attention output from different heads at different positions. For the
$j^{th}$~head, the attention output can be written as: 
\begin{equation}\label{eqq5}
 head_{J}=A\left(qW_{J}^{q},kW_{J}^{k},vW_{J}^{v} \right),
\end{equation}
\begin{equation}\label{eqq4}
\text{MHA}\left(q,k,v \right) =\left[head_{1}, head_{2} \cdots head_{h}  \right] W^{O},
\end{equation}
\noindent where $W_{J}^{q}$,~$W_{k}^{k}$,~$W_{q}^{q}$ are the projection (weight) matrices for the $j^{th}$ head, $W^{O}$ $\in$ $\mathbb{R}^{d \times d}$, and $d_{h}$ is the dimension of the output features. The attention function in Eq.(\eqref{eqq3}) can be directly applied on a set of $n$ queries $q$ $\in$ $\mathbb{R}^{n \times d}$ to obtain $F$ $\in$ $\mathbb{R}^{n \times d}$. The \ac{FFN} performs non-linear operations on the output features of the \ac{MHA}. If the dimension of the input vector $X$ $\in$ $\mathbb{R}^{n \times d}$ and the output vector is $F$ $\in$ $\mathbb{R}^{n \times d}$. The output vector $F$ is computed using Eq.(\eqref{eqq6}) in~\cite{yu2021accelerated} as:
\begin{equation}\label{eqq6}
F =\text{FFN}\left( X\right) =\text{max}\left(0, XW_{1}+b_{1} \right) W_{2}+b_{2}.
\end{equation}

Therefore, unlike with an \ac{LSTM} and \ac{GRU} which use serial tokens, transformers take parallel tokens. Transformers are better at capturing global dependencies, making them more effective at modeling complex sequences like videos. Captioning accuracy can be further improved in sequence-sequence \ac{VC} models by incorporating several techniques such as attention-based, graph-based, reinforcement-based, adversarial networks, non-autoregressive, \ac{DVC}, paragraph \ac{VC} and interactive \ac{V-QA} as shown in Fig.~\ref{fig:capt}. The chronological sequence of the representative methods is presented in the taxonomy of Fig.~\ref{fig:TAXO}.



\subsection{Datasets}
\label{Sec:sec:DATA}
The central goal in \ac{VC} is to develop models that transform videos into text. The performance of the network models is highly dependent on the knowledge learned from the training data. If these models are trained on one dataset and tested on another dataset from a different domain, the performance of the models is likely to decline dramatically~\cite{xu2020survey}. Therefore, large-scale visual datasets for specific tasks are critical. The datasets used in \ac{VC} describe open domain videos~\cite{chen2011collecting}, human activities~\cite{caba2015activitynet,ma2017grounded}, cooking recipes~\cite{iashin2020better, zhou2018towards}, movie~\cite{amirian2021automatic,xu2019semantic, wang2021end, wu2018multi}. Table~\ref{DATA} provides a summary of the most important datasets used in \ac{VC}.

\subsubsection{Open-domain databases}

\paragraph{\ac{MSVD}} The \ac{MSVD} dataset~\cite{amirian2021automatic,xu2019semantic, wang2021end, wu2018multi} (otherwise known as YouTube2Text~\cite{chen2011collecting}) is a widely used dataset in \ac{VC}. It comprises 1,970 videos and 80,839 English captions written by \ac{AMT} workers. Each video clip is associated with 40 descriptions. The clips used in \ac{MSVD} contain a single activity and are expressed with multi-lingual captions such as English, Turkish, and Chinese. For benchmarking, the training, validation, and testing splits contain 1,200, 100, and 670 videos, respectively~\cite{jin2019low,chen2020delving}. To prevent lexical bias, audio is removed from all the videos. In \ac{VC}, it is important to be mindful of biases and actively seek to use more neutral or inclusive language. However, a large number of the videos in \ac{MSVD} are of very poor quality and it presents challenges even for state-of-the-art model architectures~\cite{regneri2013grounding}. Further, the videos are short and captioned with only a single sentence~\cite{regneri2013grounding}.  

\paragraph{YouTube Highlight Dataset} The YouTube Highlight Dataset~\cite{sun2014ranking} contains videos from six domains (“skating”, “gymnastics”, “dog”, “parkour”, “surfing”, and “skiing") and the videos are variable in length. The total duration of the dataset is 1,430 minutes. The data are split in half for training and testing. This dataset poses challenges as (1) it comprises videos captured via handheld devices (2) the start and end of a specific highlight are determined by the subject (3) videos such as interviews and slideshows are included in the dataset. The videos in the test set were evaluated by \ac{AMT} workers. 

\paragraph{MSR-VTT}
Though there is an increased interest in \ac{VC}, the existing databases have limited variability and complexity. Often simple and focused on limited tasks such as e.g., cooking~\cite{das2013thousand } they are not applicable to all content. There are few large-scale datasets as videos are difficult to gather, annotate and organize. The MSR-VTT dataset was introduced to address these limitations. Compared to other datasets used in \ac{VC} such as MSVD~\cite{amirian2021automatic}, YouCook~\cite{das2013thousand},
M-VAD~\cite{torabi2015using} TACoS~\cite{regneri2013grounding}, and MPII-MD~\cite{rohrbach2013translating}, the
MSR-VTT~\cite{xu2016msr} benchmark is the largest in terms of the number of clip-sentence pairs, where each video clip is annotated with multiple sentences. MSR-VTT is particularly valuable for training large parameter models (e.g., \ac{RNN}s). The database contains videos from 257 popular queries in 20 representative categories. The creators of the database utilized over 3,400 worker hours to gather video-text pairs, annotated and summarized to boost the research in \ac{VC}. The dataset containd over 40 hours of video content, 10,000 video clips, and 200K clip-sentence pairs in total. The dataset is partitioned into training, testing, and validation sets of percentages of 65\%:30\%:5\%, respectively. 

\paragraph{\ac{VTW}}: The \ac{VTW} dataset~\cite{zeng2016generation} was designed for video title generation. The dataset spans 213.2 hours, 10 times longer than the videos in \ac{MSVD}. The dataset comprises 10k open-domain videos with annotation created by editors rather than \ac{AMT} workers. 

\paragraph{TGIF} 
Tumblr GIF~\cite{li2016tgif} is a large-scale video database of animated GIFs. TGIF comprises 100K animated examples collected from Tumblr, and 120K natural language
sentences annotated via crowd-sourcing. The crowdsource workers are gathered from only English-spoken countries like Australia, Canada, New Zealand, the UK and the USA. The 100K are split into 90K for the training and 10K for testing. 

\paragraph{VATEX}
The existing databases in \ac{VC} are heavily biased towards the English language, more multilingual datasets are necessary. 
VATEX~\cite{wang2019vatex} is more diverse compared to the voluminous MSR-VTT~\cite{xu2016msr} dataset. It contains both English and Chinese descriptions at a large scale, which can facilitate research into multilingual captioning. Another important property is that VATEX\cite{wang2019vatex} has the largest number of clip-sentence pairs with each video clip annotated with multiple unique sentences and every caption is unique in the corpus. VATEX contains more complete yet representative video content, covering 600 human activities in total, it is lexically richer and thus can enable more natural and diverse caption generation. It comprises 41,250 videos and 825,000 captions in two languages: English and Chinese. The captions in this dataset have 206,000 English-Chinese parallel translation pairs. To portray a large number of human activities, existing Kinetics-600 videos are reused in the database that contains 600 different human action classes. The official dataset partition is 25,991 for training, 3,000 for validation, and 6,000 for testing.

\subsubsection{Activity-based video datasets}
Video action classification and captioning have received a lot of research interest though the progress has been limited because of the dearth of large databases. Some databases~\cite{kay2017kinetics, goyal2017something} lack control over pose variations, motion, and other scene properties that might be important for learning fine-grained models. The datasets used for activity-based video captioning mainly focus on actors, objects, and their interactions.

\paragraph{Charades}
The videos in the Charades dataset~\cite{wei2020exploiting, sigurdsson2016hollywood} contain mundane activities taken indoors by 267 people from three continents. Charades offer 27,847 video descriptions, 66,500 temporally localized intervals for 157 action classes, and 41,104 labels for 46 object classes. Indeed, the scenes and activities captured in the dataset are quite diverse. This diversity provides unique challenges to \ac{VC}. The Charades dataset contains examples of how people interacte with objects and perform natural action sequences and presents the opportunity to learn common sense and contextual knowledge necessary for high-level reasoning and modeling tasks. 

\paragraph{Charades-Ego} The Charades-Ego dataset~\cite{sigurdsson2018charades} is similar to the Charades dataset and was collected by recruiting crowd workers on the Internet to record the videos in both first and third-person performing a given script of activities. The dataset contains 68,536 activities, 7,860 videos and 68.8 hours of first and third-person video. \ac{AMT} workers were asked to record two videos: one video acting out the script from a third-person perspective; and another collected while they perform the same script in the same way, with a camera fixed to their forehead. The creators of this dataset used a creative approach to capture examples from both first and third-point views concurrently and in a scalable fashion.  

\paragraph{ActivityNet Captions} The ActivityNet Captions dataset~\cite{krishna2017dense} comprises 20k videos and 100,000 sentences. The video caption on average describes 94.6\% of the entire video, thus making the dataset suitable for \ac{DVC}. To prove that ActivityNet Captions' captions mark semantically meaningful events, two different temporally annotated paragraphs from different workers were collected for each of the 4926 validation and 5044 test videos. Each pair of annotations was then tested to see how well they temporally corresponded to each other.

\paragraph{Something-Something V2} The dataset Something-Something V2 dataset~\cite{goyal2017something} is used for modeling fine-grained action recognition for video captioning tasks. Something-Something V2 contains 220,847 videos of 174 action categories. In the database collection, \ac{AMT} workers are asked to incorporate the action class and the objects involved. Something-Something V2 is one of the largest video datasets focused on human-object interactions.

\paragraph{Moments in Time dataset}: The Moments in Time dataset\cite{monfort2019moments} is a gathering of one million short videos labeled corresponding to an event happening within 3 seconds. The dataset comprises videos labeled from 339 different action classes, to make it possible for the models to understand the dynamics of actions in videos. It is one of the largest human-annotated video datasets capturing visual and audible short events produced by humans, animals, objects, or nature. The most commonly used verbs in the English language were chosen as a way to source the videos as a result there is a significant amount of diversity and intra-class variation. 

\paragraph{HowTo100m}
The HowTo100m dataset~\cite{miech2020end} contains more than 100 million uncurated instructional videos taken from various databases in the domains of action recognition (HMDB-51, UCF-101, Kinetics-700), text-to-video retrieval (YouCook2, MSR-VTT) and action localization (YouTube-8M Segments, Crosstalk). In the absense of manual annotation, the mismatch between narrations and videos is corrected by introducing a loss estimation Multiple Instance Learning (MIL) and Noise Contrastive Estimation (NCE), i.e., MIL-NCE, a multiple instances learning approach derived from noise contrastive estimation. After applying this loss function, the HowTo100M dataset provides solid visual representations that can outperform self-supervised and fully-supervised representations on downstream tasks. 

\paragraph{\ac{AViD} Dataset} While most datasets are biased and statistically confined to some regions and languages, \ac{AViD}\cite{piergiovanni2020avid} contains actions from diverse countries obtained using queries in many languages. The authors also consider privacy and licensing, with faces masked and all the videos having a creative commons license. 

\paragraph{\ac{S-MiT}} \ac{S-MiT} is a massive spoken video caption dataset~\cite{monfort2021spoken}, used mainly in video understanding to learn contextual information. It is formed from 3-second clips from the moments in time dataset~\cite{monfort2019moments}. In this dataset, spoken captions are aligned with video during the training phase. This is not possible with the other large video caption datasets and allows for spoken caption models to be analyzed with matching video information. The models trained on \ac{S-MiT} show better generalization in video retrieval problems. 

\subsubsection{Movie datasets}

\paragraph{MPII-MD} The MPII-MD dataset~\cite{rohrbach2015dataset} consists of videos from 94 Blu-ray Hollywood movies of diverse genres (e.g. drama, comedy, action). It comprises of 68K video-sentence pairs spanning over 73 hours. The dataset has been used to transcribe and align \ac{AD} and scripts. The dataset is freely available on the contributor’s website and the audio data has been preprocessed. Sections of the \ac{AD}s (which is mixed with the original audio stream) have been semi-automatically segmented and the segments interpreted via a crowd-sourced transcription service.

\paragraph{\ac{LSMDC}} The LSMDC dataset~\cite{rohrbach2017movie} is a fusion of two benchmark datasets, i.e., \ac{M-VAD} and MP\RomanNumeralCaps {2}-MD. It consists of video descriptions extracted from professionally generated \ac{DVS} tracks on popular movies. The \ac{LSMDC} dataset has a rich unique vocabulary tokens, i.e., 23,000. The dataset has been updated over time, \ac{LSMDC} 2015 dataset comprises 118K sentence-clips pairs and the duration of the video is about 158 hours. The training, validation, blind, and test sets contain 91,908, 6,542, 10,053, and 9,578 video clips, respectively. The \ac{LSMDC} 2016 contains a total of 128,000 clips, where 101,046 and 7,408 clips are used for training and validation, respectively.

\paragraph{\ac{M-VAD}}  The \ac{M-VAD} dataset~\cite{torabi2015using} comprises approximately 49,000 video clips from 92 movies. The alignment of descriptions to video clips was performed via an automatic procedure using \ac{DVS}~provided for the movies. It is particularly helpful mainly for the visually impaired. This dataset is the largest \ac{DVS}-derived movie dataset available requiring less human intervention in the post-processing. The whole dataset is 84.6 hours with 17,609 words. The dataset partition comprises of 38,949, 4,888 and 5,149 video clips for training, validation and testing~\cite{pasunuru2017multi, bolelli2018hierarchical,shin2016beyond,tan2020learning,wang2018hierarchical}. Each video is annotated with 20 references by \ac{AMT}.

\subsubsection{Cooking recipe datasets}

\paragraph{\ac{TACoS-Mlevel}} The \ac{TACoS-Mlevel} dataset~\cite{rohrbach2014coherent} describes routine cooking activities of 185 long videos, each with an average duration of 6 minutes. The dataset is annotated by multiple \ac{AMT} workers. Each video is partitioned into temporal intervals and each interval is annotated with a short sentence. The whole dataset is split into 16,145 distinct intervals and comprises 52,478 sentences. Considering a single video, there are about 87 intervals and 284 sentences. The importance of this dataset is that it describes video content with multiple levels of description. It comprises 15 sentence, 3-5 sentence and one sentence descriptions of each video. 

\paragraph{YouCook II} The YouCook II dataset~\cite{zhou2018towards} contains 2,000 videos with a total number of 14,000 clips, which were gathered from YouTube and include 88 different types of recipes. In this database, 9,600 clips are used for the training and 3,200 clips for validation. The dataset was annotated by \ac{AMT} workers
 and contains six different ways of cooking such as making breakfast, making sandwiches and preparing salads. The dataset was recorded from a third-person perspective. Three types of annotations are used, i.e., object tracking, actions, and human descriptions. The objects include dairy, condiments and the involved actions including putting objects down, seasoning food and pouring liquids. \ac{AMT} workers were directed to view the cooking video multiple times and to describe the video caption in three sentences with a minimum of 15 words.

\begin{table*}
\centering
\caption{Summary of datasets used in \ac{VC}.} \label{DATA}
\resizebox{0.98\textwidth}{!}{\begin{tabular}{>{\centering}p{1.7cm}>{\centering}p{1cm}>{\centering}p{1.2cm}>{\centering}p{1.7cm}>{\centering}p{1.5cm}>{\centering}p{1.5cm}>{\centering}p{1.6cm}>{\centering}p{1.6cm} c c c c }
\toprule
Dataset & Theme & \#Videos & Source & \#Clips & \#Captions & \#Words & \# Vocabulary &  \#Duration/clip & Time duration & Multi-- Sentence & Spoken \tabularnewline
\midrule
MSVD & Open & 1,970 & \ac{AMT} & 1,970 & 70,028 & 607,339 & 13,010 & 4-10 & 53 & \xmark & \xmark \tabularnewline
\midrule
MSR-VTT & Open & 7,180 & \ac{AMT} & 10,000 & 20,000 & 1,856,523 & 29,316 & 10-30 & 41.2 & \xmark  & \xmark \tabularnewline
\midrule
MVAD & Movie & 92 & \ac{DVS} & 499,886 & 55,904 & 50,926 & 17,609 & 62 & 84.6 & \checkmark & \xmark \tabularnewline
\midrule 
ActivityNet & Open & 20,000 & \ac{AMT} & 100,000 & 100,000 & 134,000 & 15,564 & 10 & 846 & \xmark & \xmark \tabularnewline
\midrule
Youcook2 & Cooking & 2,000 & \ac{AMT} & 15,400 & 15,400 & 121,418 & 2,583 & 20 & 15.9 & \checkmark & \xmark  \tabularnewline
\midrule
Charades & Human & 9,848 & \ac{AMT} & 10,000 & 27,380 & 607,339 & 13,000 & 30 & 82.01 & \xmark  & \xmark  \tabularnewline
\midrule
\ac{TACoS-Mlevel} & Cooking & 185 & \ac{DVS} & 14,105 & {--} & 2,000 & {--} & 307 & 15.9 & \xmark  & \xmark  \tabularnewline
\midrule
MPII-MD & Movie & 94 & \ac{DVS} + script & 68,337 & 68,375 & 653,467 & 24,549 & 39 & 73.6 & \xmark  & \xmark  \tabularnewline
\midrule
VATEX & Human & 41,300 & \ac{AMT} & 41,300 & 413,000 & 499,4768 & 44,103 & {--} & {--} & \checkmark & \xmark \tabularnewline
\midrule
LSMDC & Movie & 200 & \ac{DVS} + \ac{AD} & 128,000 & {--} & {--} & 23,000 & 4 & 150 & \checkmark & \xmark  \tabularnewline
\midrule
\ac{S-MiT} & Open & 515,912 & \ac{AMT}  & 515,912 & 515,912 & 5,618,064 & 50,570 & {--} & {--} &\checkmark  & \checkmark \tabularnewline
\bottomrule
\end{tabular}}
\end{table*}

\subsection{Evaluation Metrics}
\label{Sec:sec:EVAM}
Model performance evaluation by humans is extremely valuable; however, it is also costly, time-consuming, and often subjective. Therefore, automatic evaluation tools are helpful and necessary for \ac{VC}. The fidelity of evaluation metrics depends on how close the metrics reflect human judgment. The evaluation metrics of \ac{VC} fall into two main groups: evaluation of correctness and evaluation of diversity. 

\subsubsection{Evaluation of Correctness}
\label{Sec:sec:ECORRE}
The frequently used evaluation metrics of correctness
are \ac{BLEU} [B@n]~\cite{papineni2002bleu}, \ac{METEOR} [M]~\cite{denkowski2014meteor},~\ac{ROUGE-L}[R]~\cite{lin2004rouge},~\ac{CIDEr} [C]~\cite{vedantam2015cider, gao2022hierarchical}, \ac{WMD}~\cite{rubner2000earth,kusner2015word} and \ac{SPICE}~\cite{anderson2016spice}. These metrics are used for many NLP tasks, including text summarization, \ac{IC}, text similarity measurement, and the document analysis (see Table~\ref{tabbb3}). A good evaluation metric should give satisfactory results even when there are words replaced with their synonyms, redundant words are add to the text, word order is changed or sentences are abbreviated without changing the meaning~\cite{aafaq2019video}.

\begin{table} 
\caption{Evaluation metrics of text similarity and the research field in which they were initially adopted.}
\label{tabbb3}
\begin{adjustbox}{width=0.5\textwidth}
\begin{tabular}{l c}

\toprule
Evaluation metric of similarity & Research area from where it is taken\tabularnewline
\midrule
\ac{BLEU}, \ac{METEOR} & \ac{MT}\tabularnewline
\ac{CIDEr}, \ac{SPICE} & \ac{IC} \tabularnewline
\ac{ROUGE-L} & Document summarization\tabularnewline
\ac{WMD} & Document similarity\tabularnewline
\bottomrule
\end{tabular}
\end{adjustbox}
\end{table}

\paragraph{\ac{BLEU}[B@n]} This metric measures how good a machine translation coincides with a set of human-generated translations (reference translations)
by counting the percentage of n-grams in the machine translation overlapping with the references~\cite{lin2004rouge}. 

\ac{BLEU}[B@n] is then computed as follows~\cite{papineni2002bleu}:
\begin{equation}
  \ac{BLEU}[B@n] =BF \times exp (\sum_{n=1}^{N} w_{n}\log p_{n}),
\label{eq66}  
\end{equation}

\noindent where $BF$ is the brevity penalty factor, $p_{n}$ is the geometric mean of the modified n-gram precision up to length $N$ and $w_{n}$ is the weight of n-gram precision, where the sum of $w_{n}$ terms is equal to 1. Assume $cl$ be the length of the machine translation and $rl$ be the length of the reference translation. The $BF$ in Eq.(\eqref{eqNEW111}) is given by:
\begin{align}
     BF = \Bigg\{
                \begin{array}{ll}
                  1, & \text{if}~~ cl > rl\\
                  e^{( 1-\dfrac{rl}{cl})} & \text{if}~~ cl \le  rl
                \end{array}. 
             \label{eqNEW111}
\end{align}  

\noindent\doublebox{\begin{minipage}[t]{1\columnwidth - 2\fboxsep - 7.5\fboxrule - 1pt}%
Human-generated sentence 1: John is middle aged.\\
Human-generated sentence 2: John is thirty five years old.\\
Machine-generated sentence: John has thirty five years.\\
The number of unigrams in the machine-generated sentence: 5.\\
The unigrams covered in human-generated sentences: 'John','thirty,' five', 'years'.\\
So, BLUE score is: 4/5 = 0.8.
\end{minipage}}
\\
\\
The evaluation metric \ac{BLEU}[B@n] is based on precision and does not use recall. This metric considers words with synonyms as different words. The metric penalizes even very small variation in words~\cite{wei2020exploiting}. This shortcoming is addressed by \ac{METEOR} which is, as a result, generally closer to human judgments.
\vspace{-0.0cm}

\paragraph{\ac{ROUGE-L} [R]}
\ac{ROUGE-L} [R] measures the quality of a summary by comparing system-generated summaries with reference summaries. The metric considers n-gram matching, word pairs between the machine-generated summary and the ideal summaries framed by human subjects. ROUGE-N is an n-gram recall between a system-generated summary and a set of human-generated summaries. ROUGE-N is calculated as follows:
\begin{multline} \label{eq11}
ROUGE-N = \\ \frac{\sum_{s \in \text{Reference summaries}} \sum_{s \in \text{n-gram}}Count_{match}(n-gram)}{\sum_{s \in n-gram} Count(n-gram)},
\end{multline}
\noindent where $n$ stands for the length of the n-gram, $Count(n-gram)$, and $Count_{match}(n-gram)$ is the maximum number of n-gram overlaps in a candidate summary and a set of reference summaries. Rouge is a recall-based evaluation metric as the denominator of the equation is the total sum of the number of n-grams occurring on the reference summary side. In Eq.(\ref{eq11}), as more reference summaries are included in the metric, the number of matching n-grams in the denominator increases. Every time a reference is added to the pool, the space of distinct summaries is enhanced. ROUGE-N evaluates different aspects of text summarization, by controlling the type of reference added to the reference pool. In Eq.(\ref{eq11}), the numerator sums all the reference summaries which assign more weight to match n-grams occurring in multiple references. ROUGE-N gives higher scores to a machine-translated summary that has more words that overlap with the reference summaries. There are several variants of ROUGE (see~\cite{lin2004rouge}). ROUGE-L is a \ac{LCS}-based \ac{F} to compute the correlation between two text summaries $A$ of length $a$ and $B$ of length $b$ assuming $A$ is a reference summary and $B$ is a model generated summary.

\begin{equation}
\label{eq12}
 R_{LCS}=\frac{A+B}{a},
\end{equation}
\begin{equation}
\label{eq13}
 P_{LCS}=\frac{A+B}{b},
\end{equation}
\begin{equation}
\label{eq14}
 F_{LCS}=\frac{(1+\gamma^{2} R_{LCS} P_{LCS})}{ R_{LCS}+ \gamma^{2} P_{LCS}},
\end{equation}
\noindent where $LCS(A,B)$ is the length of a \ac{LCS} of $A$ and $B$, and $\gamma$ = $\dfrac{P_{LCS}}{R_{LCS}}$. \par
\noindent\doublebox{\begin{minipage}[t]{1\columnwidth - 2\fboxsep - 7.5\fboxrule - 1pt}%
Human-generated sentence 1: \underline{the gunman} kill police.\\
Human-generated sentence 2: Police kill \underline{the gunman}. \\
Human-generated sentence: Police killed the gunman. \\
The number of unigrams in the machine-generated sentence: 5. \\
The number of matching bigrams in human-generated sentences 1 and 2 is the same.\\ So, ROUGE-2 is the same for the human-generated sentence 1 and 2.\\
The number of matching unigrams in human generated sentence 1 covering in the machine-generated sentence is two that is "the", and "gunman". \\
The number of matching unigrams in human-generated sentence 2 covered in the machine-generated sentence is three that is "the", "kill", and "gunman.\\
In the first sentence ROUGE-2 score is = 2/5 = 0.4. \\
In the second sentence ROUGE-2 score is = 3/5 = 0.6. \\
The ROUGE-L score favors sentence two as it has greater meaning than sentence one.
\end{minipage}}


\paragraph{\ac{METEOR}} The \ac{METEOR} metric considers both precision and recall. Precision $P$ is calculated as the ratio of the number of uni-grams in the machine-generated translation that overlap (to uni-grams in the human-generated translation) and the total number of uni-grams in the machine-generated translation. Recall \emph{R} is calculated as the ratio of the number of uni-grams in the machine-generated translation that are overlap (to uni-grams in the reference translation) and the total number of uni-grams in the reference translation. The harmonic mean, $H_{m}$, of precision and recall is given by~\cite{banerjee2005meteor} as follows:
\begin{equation}
H_{m}=\frac{10PR}{R+9P}.
\label{eq8}
\end{equation}
The penalty used in \ac{METEOR}~\cite{banerjee2005meteor} is the form
\begin{equation}
Penalty=0.5 \times \left( \frac{\# Chunks}{ \# \text{Unigrams matched}} \right)^{3}.
\label{eq9}
\end{equation} 
The \ac{METEOR} score $M_{s}$ is computed from Eqs. \eqref{eq8} and \eqref{eq10}, as follows:
\begin{equation}
M_{s}=H_{m} \times \left( 1- Penalty \right).
\label{eq10}
\end{equation} 
\noindent\doublebox{\begin{minipage}[t]{1\columnwidth - 2\fboxsep - 7.5\fboxrule - 1pt}%
\# Computation of METEOR score	
\\
\\
\includegraphics[scale=0.7]{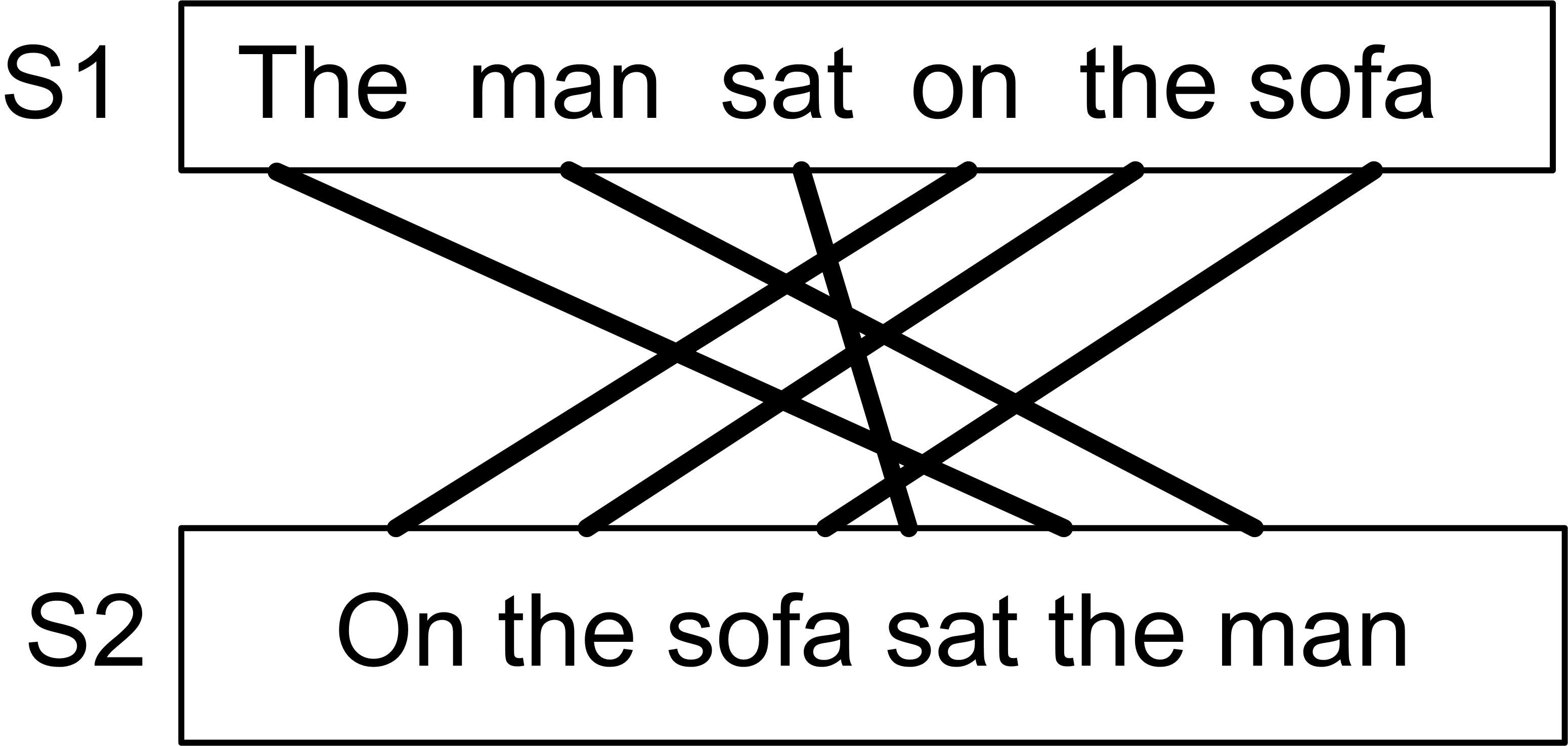}
\\
	S1 is human-generated sentence and S2 is machine-generated sentence.\\ S2 is jumbled version of S1. Since all the unigrams are matched.\\
	The precision P=6/6=1 and R=6/6=1\\
	The $H_{m}$ given by \eqref{eq8} is $H_{m}$=$\frac{10 \times 1 \times 1}{1+ 9 \times 1 }$=$1$.
	Since two subsequent unigrams "The", and "man" matched the no of chunks is 2.\\
	The penalty in \eqref{eq9} = $ 0.5 \times ( \frac{2}{6})^{3}$=$0.000013$\\
	The Meteor score in \eqref{eq10} is 1*(1-0.000013)=0.9997
\end{minipage}}

\paragraph{\ac{CIDEr}}
The \ac{CIDEr} metric~\cite{vedantam2015cider} is another evaluation score that was introduced in 2015. It is based on the correlation of a machine-produced caption with a set of ground truth captions written by different human subjects. This metric generally leads to high agreement with human consensus. When using \ac{BLEU}, analysis is carried out with five descriptions; however, this is sub-optimal when measuring how a ``majority" of humans would designate a video. Therefore, in computing \ac{CIDEr}, forty to fifty descriptions per video in the database are used. While computing n-gram words, they are first mapped to their stem or root forms. The n-grams common in the all-human references will be given less weight as they are less informative. To encode this, a \ac{TF-IDF}~\cite{christian2016single} weighting for each n-gram is used. \ac{TF-IDF} is a statistical measure that is used to evaluate how relevant n-gram is to a video in a collection of videos in the database. This is done by multiplying two metrics: how many times an n-gram appears in a video and the inverse document frequency of the n-gram across a set of videos in the database. It operates by increasing proportionally to the number of times an n-gram appears in a video but is offset by the number of videos that contain the n-gram. Thus, words that are common in every \ac{VC}, such as: this, what, and if, rank lower even though they may appear many times. However, if the n-gram appears many times in one video, while not appearing many times in others, it probably means that it is significant in some way and is given more weight. The frequency of n-gram $w_{k}$ occurring in reference sentence $s_{i,j}$ is denoted by $h_{k}s_{i,j}$ and $h_{k}c_{i,j}$ for candidate sentence. \ac{TF-IDF} weighting for each n-gram $g_{k}s_{i,j}$ is formulated~\cite{vedantam2015cider}, as:

\begin{multline}
g_{k}\left(s_{i,j} \right) = \\ \dfrac{g_{k}\left(s_{i,j} \right) }{\sum_{w_{k} in \mho} h_{l}\left(s_{i,j} \right)}\ln\left(\dfrac{\left|I \right| }{\sum_{I_{p} \in I} \text{min}\left(1, \sum_{q} h_{k}\left(s_{p,q}\right)  \right) }\right),
\end{multline}

\noindent where $\mho$ is the vocabulary of all n-grams and $I$ is the set of all videos in the dataset. The $CIDEr_{n}$ score for n-grams of length n is derived considering the average cosine similarity between the candidate sentence and the reference sentences, which incorporates both precision and recall:
\begin{equation}\label{eq18}
CIDER_{n}\left(c_{i},S_{i}\right) =
\frac{1}{m}{\sum_{j}\frac{g^{n}\left(c_{i} \right){g^{n}\left(S_{ij} \right)}}{\left\|g^{n}\left(S_{ij} \right)\right\|\left\|g^{n}\left(S_{ij} \right)\right\|  }}.
\end{equation}

\paragraph{\ac{WMD}}

\begin{figure}
	\centering
	\includegraphics[width=0.9\linewidth]{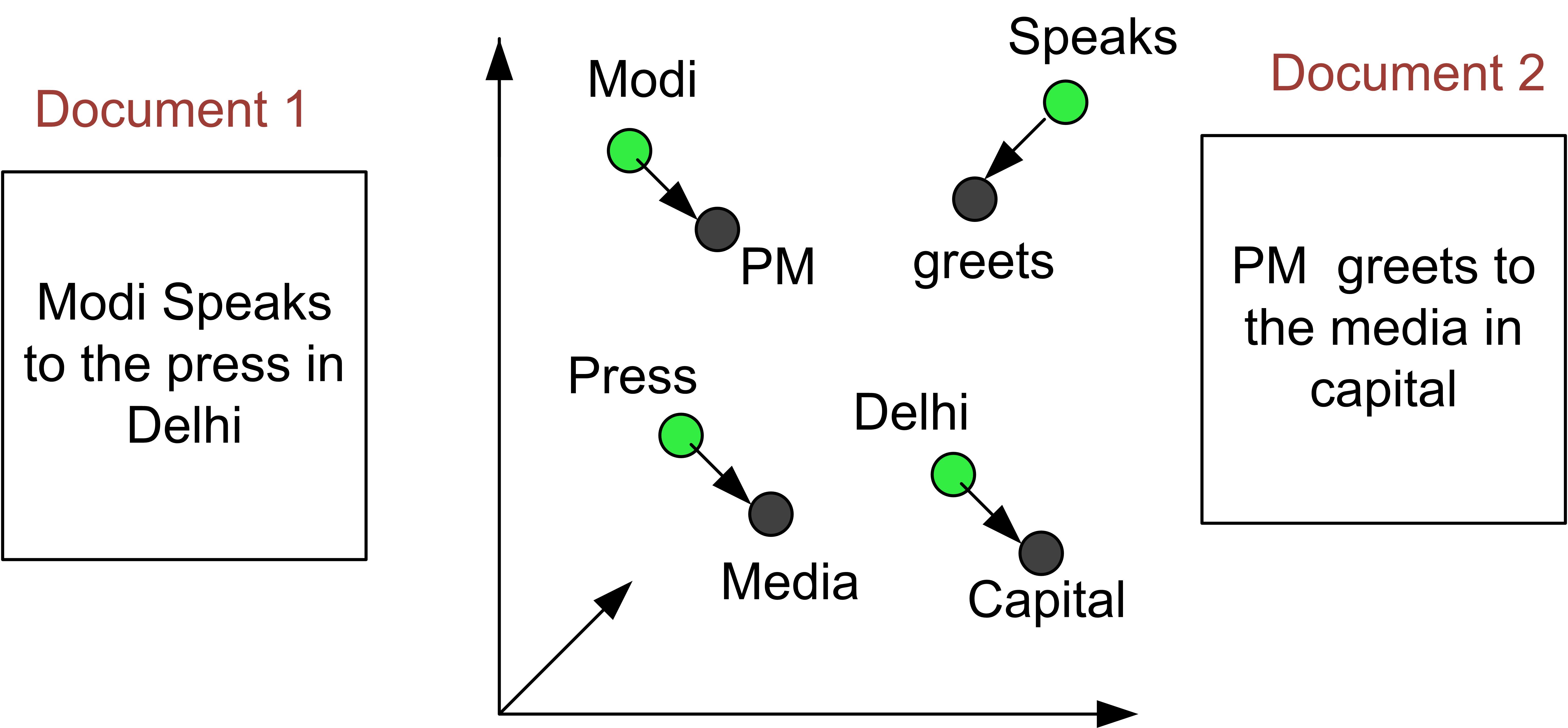}
	\caption{word2vec embedding}
	\label{fig:wordembedding}
\end{figure}

The \ac{WMD} metric is designed for the document retrieval~\cite{yang2019simple} based on document similarity. The widely used techniques for document similarity are the \ac{BoW} and \ac{TF-IDF}. As shown in Fig.~\ref{fig:wordembedding}, \ac{BoW} and \ac{TF-IDF} can not distinguish between two sentences that are semantically similar, but with different words. To solve this \ac{WMD} was proposed by Kusner et al.~\cite{kusner2015word}. The authors used word2vec embeddings and considered the text documents as a weighted point cloud of embedded words. The distance between two text documents \emph{A} and \emph{B} is the minimum cumulative distance that words from document A need to travel to match exactly the point cloud of document B. The computation of \ac{WMD} involves considering the normalized \ac{BoW} by removing the stop words using the word2vec embedding. The distance of the two texts is measured as an \ac{EMD}~\cite{rubner2000earth}, which is frequently used in transportation to determine the cost of travel. The WMD metric makes use of the Euclidean distance in the word2vec embedding space to take into consideration semantic relatedness between pairs of words. The cost to relocate every word between captions is then used to determine the distance between two texts or captions. A special case of the \ac{EMD} is used to model the \ac{WMD}~\cite{rubner2000earth} prior to the linear optimization. \ac{WMD} is less susceptible to the order of words or changing of the synonym than BLUE, \ac{ROUGE-L}[R], and \ac{CIDEr}. Moreover, it matches closely the human decisions, like \ac{CIDEr} and \ac{METEOR}.

\paragraph{\ac{SPICE}}
One of the most recent performance metrics proposed for image and video descriptions is SPICE~\cite{anderson2016spice}. BLEU, \ac{ROUGE-L}[R], and \ac{CIDEr} are n-gram sensitive and poor when evaluating sentences that have similar meanings and contain different words. Suppose the sentence is ``a young girl is standing on the tennis count", \ac{SPICE} evaluation metric searches similar words such as (1) there is a girl, (2) girl is young, (3) girl is standing, (4) there is a court, (5) court is for tennis, and (6) girl is standing on the court. If each of the propositions is contained in the caption, then the caption is given more weight in this evaluation metric. \ac{SPICE} is computed in two stages. In stage one, a dependency parser is constructed considering the syntactic dependencies between the words, and in stage two, dependency trees are transformed to scene graphs. SPICE. With the help of a dependency parse tree, the semantic scene graph preserves objects, respective attributes, and their interconnections. Object classes O(c), relation types R(c), and attribute types A(c) are examples of semantic tokens that make up a scene graph tuple G(c) of a caption $C$.

\begin{equation}\label{eq144}
G\left(C\right) =\left\langle O\left( C\right), R\left( C\right), A\left( C\right)\right\rangle.
\end{equation}

SPICE is calculated from the F1-score seen between tuples of artificially produced descriptions and the real-world data, similar to METEOR, SPICE searches WordNet for synonyms and treats them as close matches. Although the SPICE score plays only a limited role in \ac{VC}, the precision of the parsing is clearly a potential performance barrier. For example, if the word ``swimming'' is parsed as an ``object'' in the sentence ``black cat swimming through the river," and the term "cat" is interpreted as an "attribute," the sentence fails and receives a very low score.
\color{black}

\subsubsection{Evaluation of Diversity}
\label{Sec:sec:ECDIVER}
Generally, machine-generated sentences are not completely correlated with human-generated sentences because there are many correct answers, as a result diversity-based evaluation metrics such as \ac{VS}, \ac{PNS}, \ac{DC}~\cite{nabati2020multi} and \ac{DCE}~\cite{xiao2019diverse} have been developed. \ac{VS} provides the number of distinct words in the generated caption. \ac{PNS} is the percentage of generated captions not taken in the training set. To measure the diversity, mutual overlap and n-gram diversity are considered. Let $S_{X}$ be the group of sentences illustrating the $X^{th}$ test video in the dataset. The n-gram diversity (DIV-n) for the $X^{th}$ test video is calculated as the ratio of the number of distinct n-grams in $S_{X}$ to the number of n-grams in $S_{X}$. Later, the mean value of n-gram diversities for the test videos gives n-gram diversity or DIV-n value. In the calculation of the mutual overlap metrics, evaluation metrics like BLEU@N, ROUGE-L, METEOR, or CIDEr can be used to calculate the correctness score between machine-generated and human-generated sentences. For the $X^{th}$ test video, the mutual overlap is formulated as:
\begin{equation}\label{eqzzz}
Mean_{CS}(S_{X})=\dfrac{1}{\left| S_{X}\right| }\sum_{s \in S_{X}}CS(\left\lbrace s  \right\rbrace, S_{x} \in s),
\end{equation}
\noindent where $CS(c, r)$ computes correctness score between a candidate generated sentence $c$ and the human generated sentences $r$ that is measured by a measure such as BLEU@N, METEOR, ROUGE-L or CIDEr. Here $\left| S_{X}\right|$ denotes the cardinality of $S_{X}$ set. The mutual overlap metric of the whole dataset is computed with mutual overlap metrics for all the test videos in the dataset. The diversity of generated captions is vital for \ac{DVC}. The diversity is computed by computing the correlation between pairs of captions or between one caption and a set of other captions. The semantic correlation is captured with \ac{LSA}.\color{black} \par
The diversity of generated captions is a key in \ac{DVC}. The evaluation is based on diversity opposite from the similarity of the captions. The solution to determine the similarity between pairs of captions, or between one caption to a set of other captions. The semantic relatedness of the sentence is measured with \ac{LSA}~\cite{deerwester1990indexing}. The diversity is computed by finding the cosine similarity of two LSA vectors of sentence~\cite{shen2017weakly}, as follows:
\begin{equation}\label{eq20}
D_{div}=\dfrac{1}{n} \sum_{s^{i}, s^{j}\in S \; ; i\ne j}\left(1-\left\langle s^{i}, s^{j} \right\rangle  \right), 
\end{equation}
\noindent where $S$ is sentence length with cardinality $n$ and $\left\langle s^{i}, s^{j} \right\rangle$ is the cosine similarity $i$ between the $s^{i}$ and $s^{j}$. The main bottleneck in~\cite{deerwester1990indexing} is that it does not consider the rationality of the sentence. The two very different sentences get high diversity scores but their descriptions may be wrong. The other bottleneck of~\cite{deerwester1990indexing} is that the proposed \ac{LSA} method is not able to capture polysemy. To address these problems,~\cite{xiao2019diverse} proposed a metric called \ac{DCE}. It is mainly formulated from two angles: diversity between sentences computed at the same time maintained sentences reasonableness. Instead of using the difference between sentences as in~\cite{shen2017weakly}, Jaccard similarity coefficient~\cite{real1996probabilistic} is used in determining \ac{DCE}, which is the best to deal with discrete data in modeling word level correlations. The use of \ac{BERT} also alleviates the problems associated with \ac{BoW} to produce sentence-level representation. The DCE can be computed as~\cite{xiao2019diverse}:
\begin{multline}
\label{eq21}
  DCE=\dfrac{1}{m_v}\sum_{k=1}^{m_v}\dfrac{1}{n}\sum_{s^{i}, s^{j}\in \mathbf{S} , i\ne j}  [M(s^i)+M(s^j)].[\delta(1-J(s^i,s^j))\\+(1-\delta)(1-\left\langle BT(s^i),BT(s^j)\right\rangle],
\end{multline}
\noindent where $S$ is the sentence set with cardinality $n$ (i.e., if each video comprises 10 captions, $n$ maybe 20 because each sentence is used to calculate the similarities with others), $m_{v}$ is the number of videos, $\delta$ is the adjustment coefficient, $M(s)$ is the METEOR score of candidate sentence $s$, $BT(s)$ is the sentence vector encoded by BERT and $\left\langle \right\rangle$ represents the Jaccard and cosine similarity.

\subsection{Training Loss of Video Captioning}
One of the key considerations in machine learning is the computation of the training loss and \ac{VC} is no exception. Suppose $\mathcal{V}$ is video and its ground-truth caption $\mathbf{y}$=[$\mathbf{y}_{1}$,$\mathbf{y}_{2}$, $\cdots$ $\mathbf{y}_{m}$] from a dataset on which the model is being trained $D$, the function for loss is computed as:~\cite{ryu2021semantic}
\begin{equation}\label{L11}
\mathcal{L}=\mathcal{L}_{CE}+\lambda_{CA} \mathcal{L}_{CA},
\end{equation}
\noindent where $\mathcal{L}_{CE}$ is the cross entropy loss and $\mathcal{L}_{CA}$ is the cross attentive loss. The $\mathcal{L}_{CE}$ is obtained by taking the log-likelihood which is negative to bring about the correct caption as:

\begin{equation}\label{keyWWW}
 \mathcal{L}_{CE}=\sum_{\left( \mathcal{V},\mathbf{y}\right) \in D }\sum_{t}\left(-\log P\left(\mathbf{y}_{m}|\mathcal{V},\mathbf{y}_{1},\mathbf{y}_{2}, \cdots \mathbf{y}_{m} \right) \right).
\end{equation}

In order to improve the sentence semantic coherence, in addition to $\mathcal{L}_{CE}$, $\mathcal{L}_{CA}$ loss is also used. In order for a semantic group to have a consistent meaning for each of its members, it must have frames that have a high correlation with the relevant phrases. To obtain this, a negative video that does not overlap with the input video's caption is chosen at random from a collection of videos. The frames of the negative video are provided as erroneous candidates for the semantic aligner. The positive relevance score $\alpha_{i,j,t}^{positive}$ between a phrase $P_{i}$ and an input frame $\mathcal{V}_{j}$, and the relevance score which is negative is computed between a phrase $P_{i}$ and negative frame $\mathcal{V}_{j}^{neg}$. Later the obtained positive and negative scores are normalized by implementing the softmax.
Let $Prob_{ca}(se_{i,t})$=$\sum_{1}^{m}\alpha_{i,j,t}^{positive}$ indicates the probability that the semantic group $se_{i,t}$ will not have any frame which is negative. $Prob_{ca}(se_{i,t})$ increases with a positive relevance score compared to the negative relevance score and hence it is named contrastive attentive loss. The contrastive loss is given by Eq.(\ref{CLEQU}).
\begin{equation}\label{CLEQU}
 \mathcal{L}_{CA}=\sum_{\left( \mathcal{V},\mathbf{y}\right) \in D }\sum_{t} \sum_{i}^{m_{t}}\left( log \ Prob_{ca}(se_{i,t} \right),
\end{equation}
\noindent where $m_{t}$ is the surviving phrases after filtering with the phase suppressor.
\color{black}
Most \ac{VC} methods use a single decoder in training. While, Lin et al.~\cite{lin2021augmented}, for the first time, trained multiple decoders to boost the accuracy by reducing mimicry loss using Augmented Partial Mutual Learning (APML). The mutual learning strategy is used to transfer knowledge between various types of decoders such as \ac{LSTM}, \ac{GRU}, and transformers. Unlike the work of~\cite{hinton2015distilling}, which trains two models, i.e., teacher and student, in parallel and the presentation of the student model is enhanced under the teacher's guidance, in the mutual learning all the decoders act as teachers and try to improve their performance by minimizing the mimicry loss. In addition, Chen et al.~\cite{chen2020semantics} improved the accuracy of the \ac{VC} by minimizing a sentence-length-modulated loss function. 
\color{black}

\section{Review of Video Captioning Methods}
\label{Sec:PM}
In this section, the main deep learning methods and findings in \ac{VC} research are discussed (refer to Fig.~\ref{fig:capt} for the taxonomoy of methods).


\subsection{Attention-based methods} 

Attention-based models (e.g., transformers) that use deep learning~\cite{cao2019image,lu2021chinese} are heavily utilized in \ac{NLP} applications such as machine language translation~\cite{bahdanau2014neural, wang2021multi,singh2021attention}, dialogue generation~\cite{shang2015neural}, machine reading comprehension~\cite{seo2016bidirectional}, natural language inference~\cite{li2017end}, and \ac{VC}~\cite{xiao2020video,chen2021multi}. In sequence-to-sequence networks, the context of preceding words and the encoder's fixed output are used by the decoder to construct the subsequent word. Therefore, it is possible that the overall context of the encoded information is lost, making the output extremely reliant on the most recent hidden cell state. By storing the context from the start to the completion of the sequence, the attention mechanism's adoption fixes the problem~\cite{rahman2021video}. In view of this, the attention mechanism is used in \ac{VC}, which learns when and what portions of the video the decoder has to attend to. The incorporation of attention mechanisms helps the model in locating relevant regions of the input. Since not every frame in a video is equally important to the video, the attention mechanism on the decoder side determines a weight distribution for each frame. There are several attention mechanisms used in \ac{VC} such as \ac{Soft-Attn},~\ac{Hard-Attn},~\ac{Global-Attn},~\ac{Local-Attn},~\ac{Text-Attn},~ \ac{Temp-Attn},~\ac{Self-Attn},~\ac{MHA}, and~\ac{Stacked-Attn}. 

\begin{table*}[htbp]
\scriptsize
\caption{Summary of attention-based \ac{VC} techniques.}\label{tab5}
\centering
\begin{threeparttable}
\begin{tabular}{>{\centering}p{2.8cm}>{\centering}p{0.6cm}>{\centering}p{4cm}>{\centering}p{3cm}>{\centering}p{3cm}}
	\toprule 
	Study & Year & Method & Dataset & Attention type\tabularnewline
	\midrule
	Yao et al.~\cite{yao2015describing} & 2015 & \makecell{\ac{VC}-based \\spatio--temporal attention\tnote{1}} & Youtube2Text,DVS & Spatiotemporal-Attention \tabularnewline
	\midrule 
	Li et al.~\cite{li2017mam}  & 2017 & MAM+RNN\tnote{2} &  \ac{MSVD},Charades & Multi-level Attention  \tabularnewline
	\midrule 
  Yu et al.~\cite{yu2017supervising} & 2017 & GLEN\tnote{3} &  VAS,LSMDC,Hollywood2  & Spatio-Temporal Attention  \tabularnewline
	\midrule 
 Chen et al.~\cite{chen2018tvt} & 2018 &  TVT\tnote{4}&  \ac{MSVD},\ac{MSR-VTT} & \ac{MHA}  \tabularnewline
\midrule
 Li et al.~\cite{li2018multimodal} & 2018 & Multimodal Architecture\tnote{5} &  \ac{MSVD},\ac{MSR-VTT} & Dynamic Attention  \tabularnewline
\midrule
 Yan et al.~\cite{yan2019stat} & \text{2019} & STAT\tnote{6} &  \ac{MSVD},\ac{MSR-VTT} &  Spatial-temporal Attention \tabularnewline 
\midrule
 Dong et al.~\cite{dong2019not} & 2019 & HA\_IL\tnote{7} & \ac{MSVD},\ac{MSR-VTT} &  Hierarchical Attention \tabularnewline 
\midrule
 Zhu and Jiang~\cite{zhu2019attention} & 2019 & DenseLSTM\tnote{8} & \ac{MSVD},\ac{MSR-VTT} & \ac{Self-Attn} \tabularnewline

\midrule
 Sah et al.~\cite{sah2020understanding} & 2020 & MHB-GA\tnote{9} & \ac{MSVD},\ac{MSR-VTT},\ac{M-VAD}  & \text{Gaussian attention} \tabularnewline
\midrule
 Sur~\cite{sur2020sact} & 2020 &  SACT\tnote{10} & ActivityNet,YouCookII  & \text{Multinomial Attention} \tabularnewline
\midrule
 Hori et al.~\cite{hori2021optimizing} & 2021 & LLAVC\tnote{11}& ActivityNet & \text{MHA} \tabularnewline 
 \midrule
 Perez-Martin et al.~\cite{perez2021attentive} & \text{2021} & AVSSN\tnote{12}  & \ac{MSR-VTT} & \ac{MHA} \tabularnewline
\midrule
Tu et al.~\cite{tu2021enhancing} & \text{2021} & Visual tags--TTA\tnote{13}& \ac{MSVD},\ac{MSR-VTT} & Textual Temporal Attention \tabularnewline
\midrule
Nakamura et al.~\cite{nakamura2021sensor} & \text{2022} & \makecell{DMA\tnote{14}} & MMAC captions & Dynamic Attention \tabularnewline
\midrule
Lin et al.~\cite{lin2022swinbert} & 2022 & SWINBERT & \makecell{\ac{MSVD},YouCookII,\ac{MSR-VTT},\\TVC,VATEX}  & Sparse Attention \tabularnewline %
\midrule
 Dai et al.~\cite{deb2022variational} & 2022 & VSLAN\tnote{15} & \ac{MSVD}+\ac{MSR-VTT} & Joint Hierarchical Attention \tabularnewline
 \midrule
 Song et al.~\cite{song2022contextual} & 2022 & CANet\tnote{16} & \ac{MSVD},EmVidCap,EmVidCap-S & Contextual Attention \tabularnewline
	\bottomrule
\end{tabular} %
\begin{tablenotes}
\item[1] \ac{3D-CNN}+\ac{RNN} encoder+LSTM decoder with spatio-tamporal attention;
\item[2] Multi-level Attention Model based \ac{RNN}; 
\item[3] Gaze Encoding Attention Network with Multi modal \ac{GRU};
\item[4] Two-view transformer with \ac{MHA}
\item[5] Multi modal architecture for \ac{VC} with memory networks and an attention mechanism;
\item[6] Spatial-Temporal Attention Mechanism for Video Captioning;
\item[7] Hierarchical Attention Model with Information Loss;
\item[8] Attention-based Densely Connected Long Short-Term Memory;
\item[9] Multi--stream Hierarchical Boundary with Gaussian Attention;
\item[10] Self-aware multi-space Feature Composition Transformer for Multinomial Attention;
\item[11] Low latency \ac{VC} with transformer with \ac{MHA};
\item[12] Attentive Visual Semantic Specialized Network with Temporal Attention;
\item[13] Visual Tags and Textual Temporal Attention;
\item[14] Dynamic Modal Attention;
\item[15] Variational Stacked Local Attention Network;
\item[16] Contextual Attention Network.
\end{tablenotes}
\end{threeparttable}		
\end{table*}
\subsubsection{Soft Attention}
The attention states used in \ac{Soft-Attn} are based on the previous hidden state of \ac{RNN} and visual features~\cite{yao2015describing, zanfir2016spatio,sah2020understanding}. The context vector is computed as the weighted sum of all previously hidden state features and computed using $V_{t}=\sum_{k=1}^{N}\alpha_{t,i}h_{i}^{e}$ where $h_{i}^{e}$ are previous hidden state of encoder, $\alpha_{t,i}$ are weights and $t$ is the time step. These weights $\alpha_{t,i}$ act as alignment mechanism which give higher weights to certain encoders' hidden states that allow them meet that decoder time step better, and are calculated as:
\begin{equation}\label{eq20HHH}
\alpha_{t,i} =\frac{exp \left(e_{t,i}\right) }{\sum_{j=1}^{N}exp\left(e_{t,j}\right)  }.
\end{equation}

The unnormalized relevance score is calculated using:
\begin{equation}\label{eq21SAAA}
e_{t,i} =w^{T}\text{tanh}(W_{a}h_{i}^{e} +W_{b}h_{t-1}^{d}+b),
\end{equation}
\noindent where \emph{w}, $W_{a}$, $W_{b}$ are learnable parameters. The context vector is given by weighted sum of the encoder hidden states and is given by:
\begin{equation}\label{eq22}
c_{t}=\sum_{j=1}^{N}\alpha_{t,i}h_{i}^{e}.
\end{equation}

\ac{Soft-Attn} is fully differentiable and can employ end-to-end backpropagation. The softmax weights non-zero probability to unimportant elements, which will deteriorate the attention given to the few significant elements. A number of \ac{VC} methods use \ac{Soft-Attn}~\cite{lei2021video,liu2020sibnet,olivastri2019end,chen2019boundary}.

\subsubsection{Hard Attention}
\ac{Hard-Attn}~\cite{xu2015show} selects a key region in video based on a multinoulli distribution and requires Monte Carlo sampling to train. It relies on the detection of objects and is trained by augmenting the approximate variational lower bound~\cite{gui2019semantic}. The~\ac{Hard-Attn} model is based on sequential sampling and it is non-differentiable. It relies on policy gradient reinforcement algorithms to compute attention weights.

\subsubsection{Global Attention + Local Attention} A \ac{Global-Attn}~\cite{jin2019recurrent} mechanism assesses the attention weights of the temporal features which are global (i.e., not localized to a particular region of the input). 
\ac{Local-Attn}~\cite{jin2019recurrent} mechanisms assess the attention weights of the object features which are constrained to local regions of the input. Jin et al.~\cite{jin2019recurrent} fused global temporal features and local object-based features in a complementary way to create a multimodal attention mechanism. Peng et al.~\cite{peng2021video} introduced tags in language decoding by using the global control of the text and local strengthening of it during training. In another study, Deb et al.~\cite{deb2022variational} proposed a variational stacked local attention network (VSLAN) which consists of a local attention network (LAN) and a feature aggregation network (FAN), exploits visual features from assorted pre-trained models for \ac{VC}. LAN attends related clips of a video using bilinear pooling. Then, FAN aggregates the features in a way that previously learned information is discounted from the preceding LAN. Both LAN and FAN are integrated into the decoder for captioning.
\subsubsection{Spatio--temporal attention} This category gives attention to salient regions within the frames in the video. In temporal attention, the attention mechanism focuses on keyframes. In Spatio-temporal attention, both spatial and temporal attention are exploited. Chen et al.~\cite{chen2016video} and Yao et al.~\cite{yao2015describing} developed a temporal attention method that utilizes attention weights to enable the decoder to concentrate only on a specific group of frames. The authors in~\cite{yan2019stat} proposed \ac{STAT} using \ac{2D-CNN}, \ac{3D-CNN} and \ac{RNN} as encoder and \ac{LSTM} as decoder. \ac{STAT} not only emphasizes on essential frames but also on critical areas within those frames, enabling the decoder to gather sufficient input data and perform precise decoding. In addition, Liu et al.~\cite{liu2018fine} formulated a spatio-temporal attention model by placing emphasis on the objects in the video at the fine-grained region level. A \ac{MAM} to encode the video features at the frame and region level is proposed in~\cite{li2017mam}. It is able to focus on the most correlated visual features in generating the correct caption. Region level attention layer is used to extract the salient regions in the video and the frame level attention layer is used to derive a subset of frames that is correlated to the video caption. In~\cite{yu2017supervising}, the authors proposed the Gaze Encoding Attention Network that leverages spatial and temporal attention. Spatial attention produces feature pools that are regulated by gaze maps, as well as temporal attention identifies a sample of feature pools for the generation of words by decoder modules. Additionally, in~\cite{chen2018less}, they incorporated a succession of visual features which are given as input into the \ac{LSTM} encoder, and a \ac{GRU} decoder assists in producing the sentence. This innovative work's fundamental idea is the utilization of a CNN and \ac{RL} to decide whether or not a frame has to be encoded. \par
\begin{figure}[ht!]
\centering
    \includegraphics[width=0.48\textwidth]{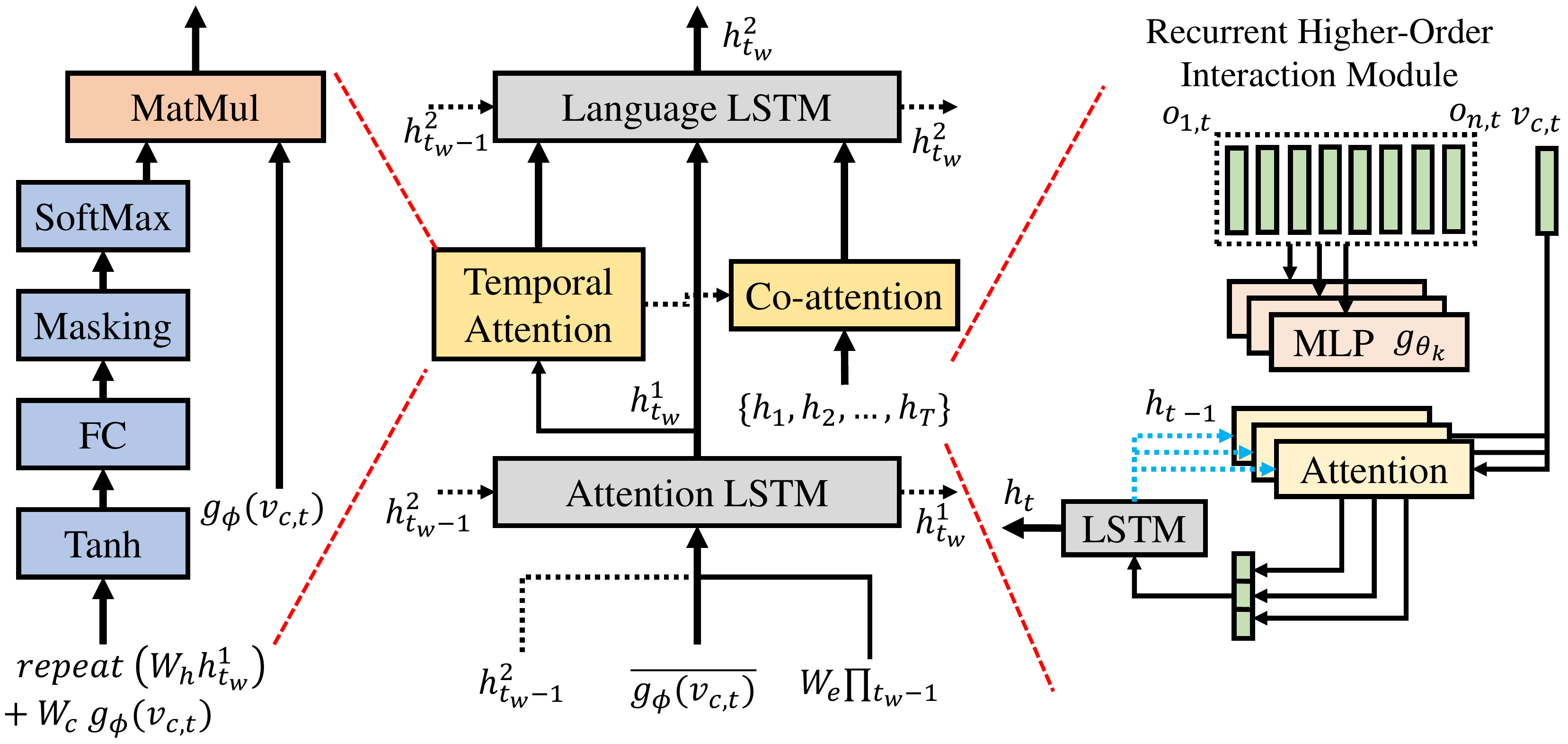}
    \caption{The SINet-Caption model used in~\cite{ma2017grounded}.}
    \label{fig:SINet}
\end{figure}

In another study, Chen et al.~\cite{chen2018tvt} proposed \ac{VC} based on Two-View Transformer and fusion blocks. The two fusion blocks used in the work are add-fusion and attentive fusion. There are two benefits to an attentive-fusion block over an add-fusion block. Firstly, the weights assigned to attention vary according to the circumstances of the current position. Secondly, by choosing various modalities like representation of frame, representation of motion, as well as previously generated words, the decoder is able to select a quality associated context that will jointly direct the description generation methodology. The researchers in~\cite{ma2017grounded} emphasized on fine-grained object interaction for video understanding using Attention.~\ac{LSTM} which is called SINet-Caption, as indicated in Fig.~\ref{fig:SINet}. The proposed model uses both fine-grained (object) and coarse- (overall image) visual representations for each video frame. Temporal and co-attention are employed for obtaining high-level object interactions in video caption generation. The authors in~\cite{dong2019not} used hierarchical layers of \ac{LSTM} and two attention layers. The two attention layers capture the information at the frame level and a novel deep \ac{VC} architecture which combines a textual memory, a visual memory, and an attribute memory in a hierarchical way to guide attention for efficient video representation extraction and semantic attribute selection~\cite{wang2018hierarchical}. Perez et al.~\cite{perez2021attentive} introduced a \ac{VC} by incorporating two attention models-temporal and adaptive attention. The temporal attention in the \ac{VC} focuses attention on the keyframes. The adaptive attention mechanism uses two LSTMs as a fusion gate. The fusion gates determine when to provide visual features and when to provide the semantic context information from the bottom and top \ac{LSTM} layers. Study~\cite{tu2021enhancing} proposed textual temporal-based attention for improving captioning by introducing pre--detected visual tags that not only belong to textual modality but also can convey visual information. In another research, hierarchically combines spatial and temporal attention in two different orders: (i) spatiotemporal (ST), and (ii) temporal-spatial (TS) attention. ST, first, applies spatial attention and linear pooling on features extracted from each frame and then applies temporal attention. 
In another study, STaTS~\cite{cherian2020spatio}, which hierarchically integrates spatial and temporal attention in two separate orders: (i) \ac{ST}, and (ii) \ac{TS} attention. \ac{ST} first performs linear pooling and spatial attention to the characteristics that were retrieved from each frame, after which it incorporates temporal attention. In addition, an LSTM-based ranking strategy is formulated to preserve the temporal order. However, all words in the caption may not depend on features that are varying temporarily. A proposed solution, called TS, addresses this problem by first applying temporal attention to identify specific frames to respond, followed by spatial attention to the spatial feature representations of selected frames.
\subsubsection{Self-Attention and Multi-head attention}
The \ac{Self-Attn} and \ac{MHA} are widely used in transformers as discussed in section~\ref{Sec:PLE}. In Eq.(\ref{eqq3}), If the three values $q$, $k$, and $v$ belong to the same feature, then it is called \ac{Self-Attn}, otherwise it is called cross-attention. Each scaled-dot product module is called a head and in \ac{MHA}, multiple heads are used to boost the captioning accuracy as given in Eq.(\ref{eqq5}) and Eq.(\ref{eqq4}). In~\cite{zhu2019attention}, self-attention dense \ac{LSTM} and latest guiding dense \ac{LSTM} are used to get differential weights in the attention mechanism and to speed up the captioning process~\cite{zhu2019attention}. The authors argued that if the attention schemes based on summation and concatenation to fuse previous hidden states, it would result same weight to predict the target word, and therefore the authors employed \ac{Self-Attn} scheme. Research~\cite{qi2021video} designed \ac{VC} based on a symmetric bidirectional decoder. The authors exploited \ac{Self-Attn} and \ac{MHA}. \ac{MHA} is used in the work for past and future predictions. A low latency online \ac{VC} is proposed based on a transformer with a multi-modal extension of video and audio in the work of~\cite{hori2021optimizing}. The flaw in \ac{MHA} is the attention affected by the similarity of the features in multiple frame sequences~\cite{sur2020sact}. The authors pinpointed the loopholes in \ac{MHA} that they were in a deficit of coherence and often considers unrelated features. The lack of sufficient procedures that can add to the contents of the frames is the cause. For that issue, Self-Aware Composition attention offers a solution. 
\subsubsection{Other Attention Mechanisms}
For multiple data frame-based problems, self-aware composition attention gave importance to classifying the regions of interest, but also on expanding the knowledge regarding the useful frames. This will lessen pollution and might even benefit in improved composition. The loopholes of the present \ac{VC} such as the existence of gaps in current semantic representations and the generated captions are prevented in~\cite{perez2021attentive} by proposing \ac{AVSSN}. \ac{AVSSN} can choose to decide when to use visual or semantic information in the language generation process.\color{black}
 The authors in~\cite{deng2021syntax} formulated a syntax-guided hierarchical attention network, which exploits semantic and syntax cues to bridge the gap between the visual and sentence-context features for captioning. A globally-dependent context encoder is introduced to extract the global sentence-context feature that gives flexibility in generating non-visual words. Later, hierarchical content attention and syntax attention are used in captioning. The semantic alignment between the vision and language in captioning is improved by introducing a semantic alignment refiner. This refiner minimizes the distance between the original feature and reproduces one which is expected to be as clear as possible.
 
\begin{figure}[ht]
\centering
    \includegraphics[width=0.4\textwidth]{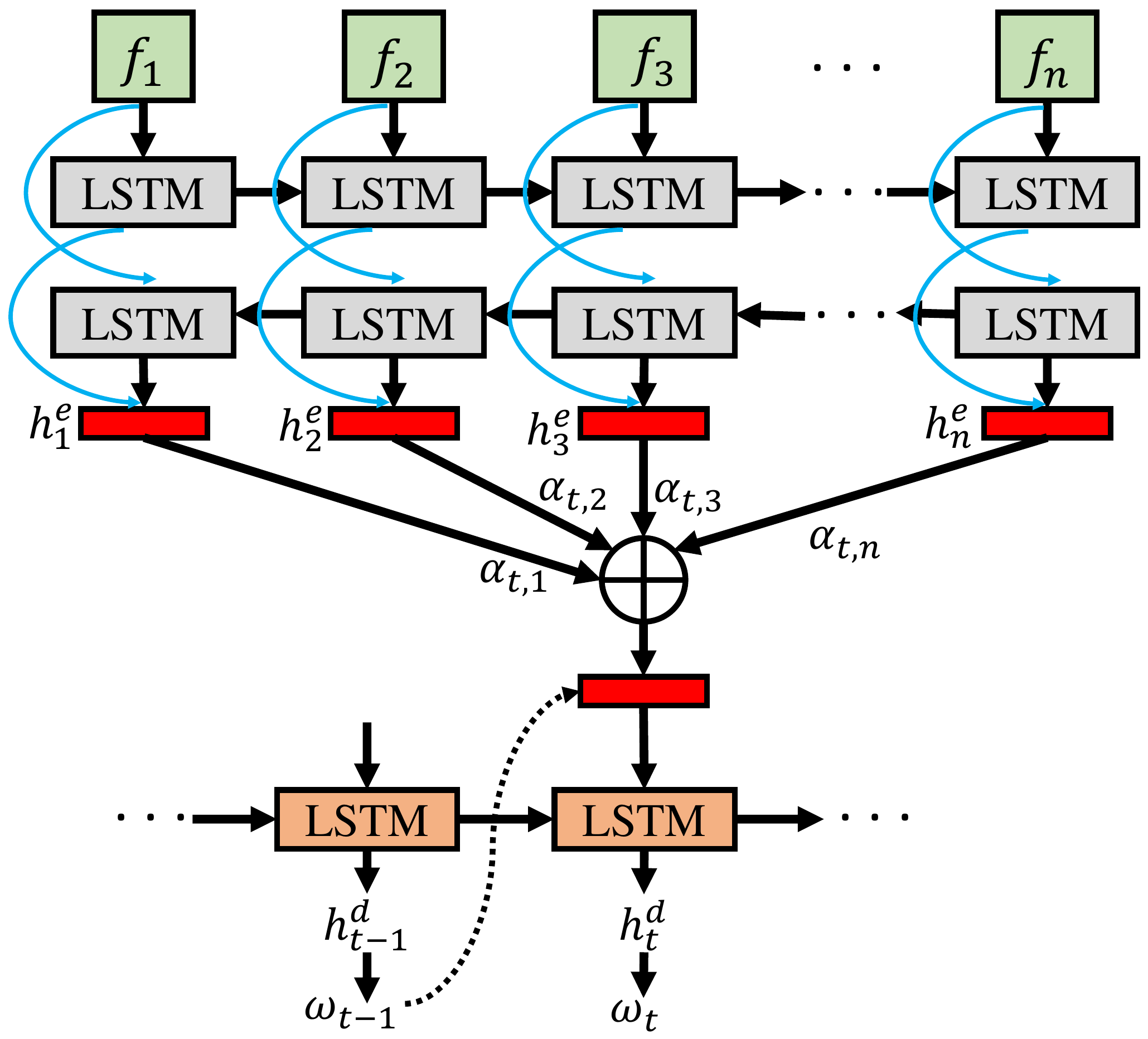}
    \caption{An attention-based sequence-to-sequence model used in~\cite{pasunuru2017multi}.}
    \label{fig:SEEEQQQ}
\end{figure}

A few studies have recently centered on dynamics fusion by exploiting multi-modal features. In accordance with the status of the model, task-driven dynamic fusion (TDDF)~\cite{zhang2017task} linearly combined heterogeneous data. The authors of~\cite{hori2017attention} suggested an attentional fusion (AF)-based modality-dependent hierarchical fusion network.
 In~\cite{pasunuru2017multi}, a many-to-many multi-task learning model with an attention mechanism (attention-based sequence-to-sequence model for \ac{VC}) was proposed (see Fig.~\ref{fig:SEEEQQQ}). The authors in~\cite{wu2018interpretable} proposed a trajectory-structured attentional encoder-decoder (TSA-ED) neural network framework for \ac{VC}. In the encoder-decoder framework, \ac{LSTM} is used and sentence description depends on the moving trajectory of objects in the video.

\begin{figure}[ht]
\centering
    \includegraphics[width=0.45\textwidth]{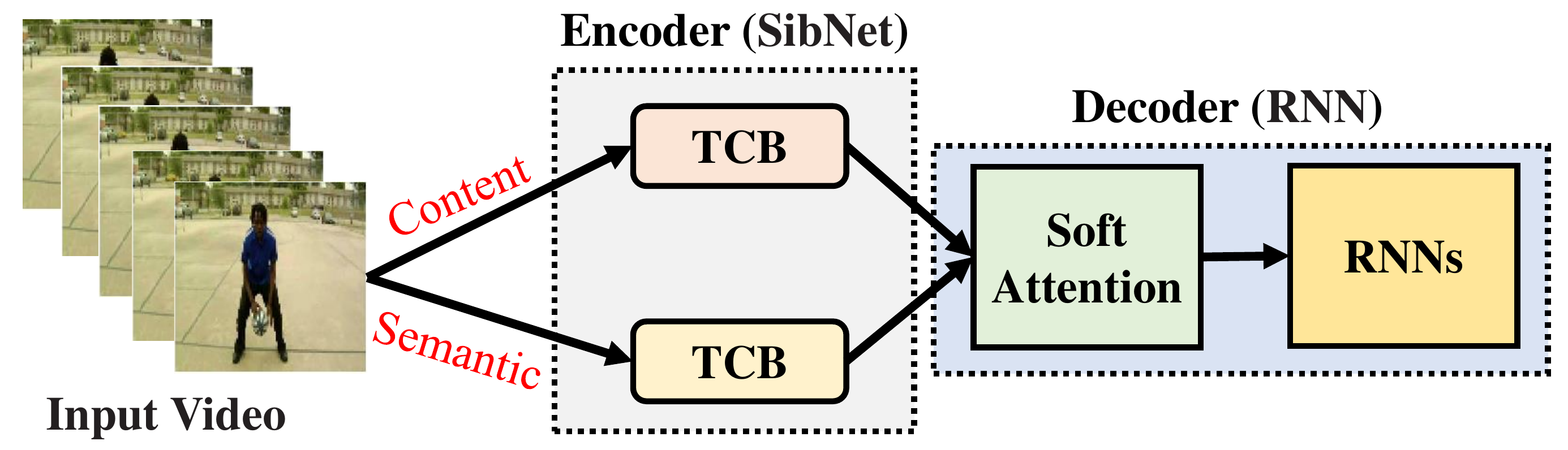}
    \caption{Sibling Convolutional Encoder (SibNet) used in~\cite{liu2020sibnet} where TCB stands for temporal Convolutional block.}
    \label{fig:SibNet}
\end{figure}

\begin{figure}[ht]
\centering
    \includegraphics[width=0.45\textwidth]{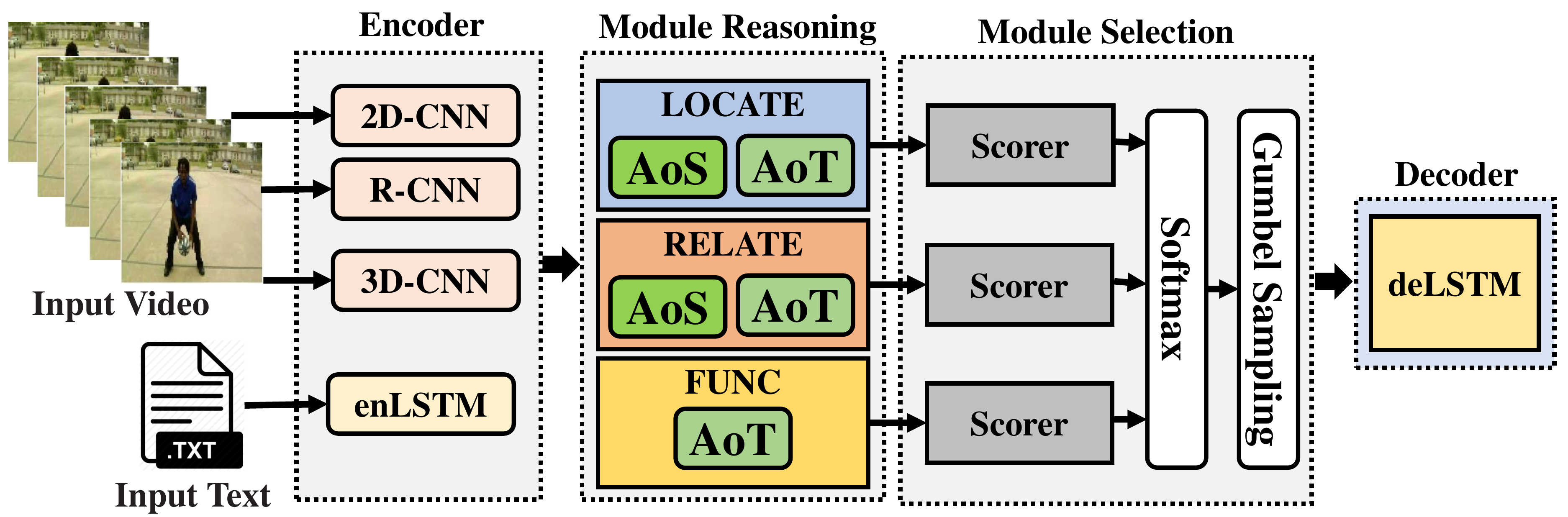}
    \caption{Reasoning Module Networks (RMN) with AoS (Attention over Space) and AoT (Attention over Time) used in~\cite{tan2020learning}.}
    \label{fig:AoT}
\end{figure}

\begin{figure*}
\centering
        \begin{subfigure}[b]{0.46\textwidth}
            \centering
            \includegraphics[width=\textwidth]{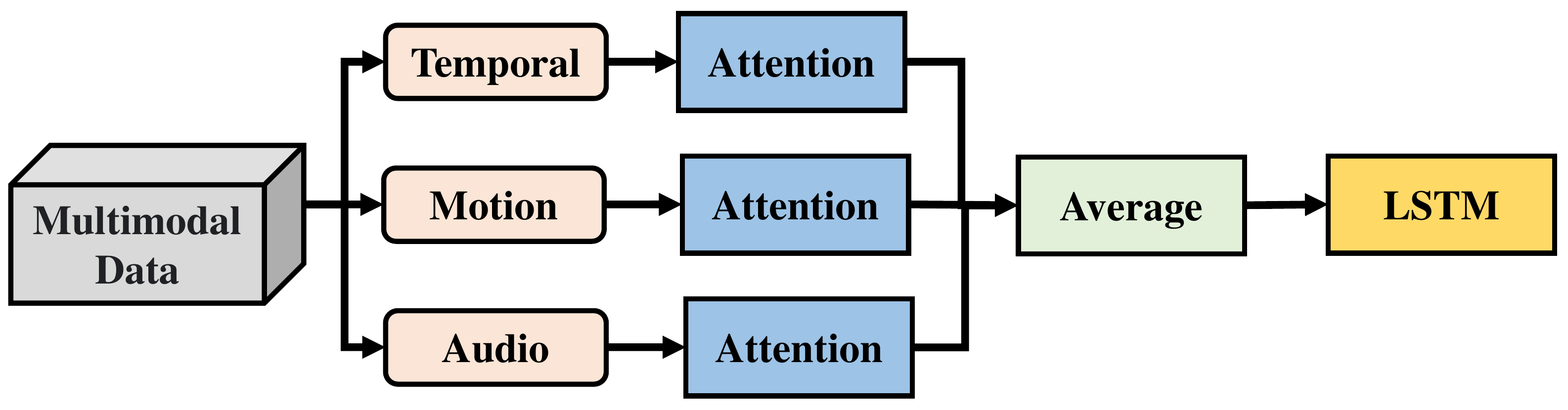}
            \caption{\ac{AAMF}}
            \label{fig:emd}
    \end{subfigure} 
     \begin{subfigure}[b]{0.46\textwidth}
            \centering
            \includegraphics[width=\textwidth]{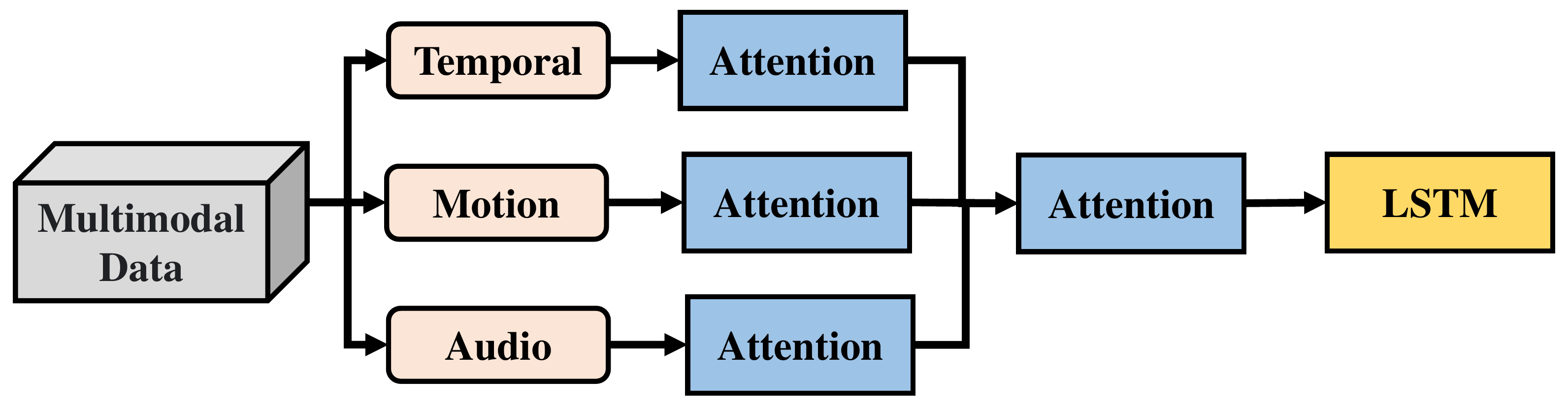}
            \caption{\ac{AMF}}
            \label{fig:AMF}
    \end{subfigure} \\
    \begin{subfigure}[b]{0.76\textwidth}
            \centering
            \includegraphics[width=\textwidth]{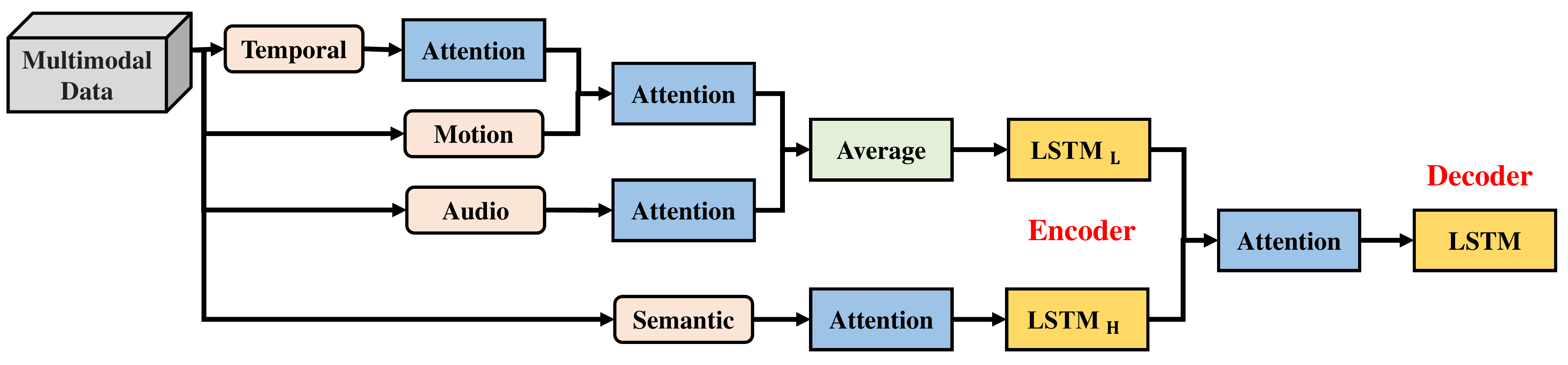}
            \caption{\ac{HATT}}
            \label{fig:HATT}
    \end{subfigure}
    \caption{Three well-known attention-based multimodal fusion methods for \ac{VC}~\cite{wu2018hierarchical}.}\label{fig:MUFUATT}
\end{figure*}

In addition, a variational \ac{POS} encoder (VaPEn) has been proposed to enforce diversity in sentences. The proposed method by Hou et al.~\cite{hou2019joint} consists of a video \ac{POS} tagging and visual cue translation that are jointly trained for \ac{VC}. It considers both visual perceptions and syntax representations for generating captions. Specifically, \ac{POS} tags are used for representing syntax structure. It also uses a mixture model to implement the visual perceptions. The mixture model translates visual cues into lexical words conditioned on the extracted syntactic structure. Also, there are different feature fusion with attention mechanism used in \ac{VC}. For example, \ac{HATT}~\cite{wu2018hierarchical} is a multi-modal fusion model based on hierarchical attention strategy. It fuses different modalities with attention manner to exploit the complementariness of multi-modal features. \ac{HATT} consists of three attention layers: \textit{(i)} low-level attention for dealing with temporal, motion, and audio features, \textit{(ii)} high-level attention for focusing on semantic labels, and \textit{(iii)} sequential attention layer for incorporating information obtained by low-level attention with high-level attention. Fig.~\ref{fig:MUFUATT} clearly shows the main differences between the three well-known attention-based multimodal fusion methods, i.e., \ac{AAMF}, \ac{AMF}, and \ac{HATT}. 

\ac{HRNAT}~\cite{gao2022hierarchical} consists of two components: \textit{(i)} hierarchical representation network (HRN), and \textit{(ii)} auxiliary tasks. HRN includes a hierarchical encoder module for representing multi-level visual concepts and a description generation module. The former obtains contextual features by employing different multilevel concepts, and the latter generates sequential descriptions. In addition, \ac{HRNAT} learns three auxiliary tasks to obtain rich semantics-aware, syntax-aware and content-aware cues in a self-supervised manner. Specifically, the semantics-aware cues of hierarchical semantic representation are learned by the encoder from the cross-modality matching task. While the syntax-aware and content-aware cues during a language generation process are obtained by the decoder. 

In~\cite{chen2021motion}, a recurrent region attention module is proposed for extracting multiple regions from each video frame. Then, a motion-guided cross-frame message passing is designed for encoding spatial information for \ac{VC}. Video representations are compacted flexibly by proposing an adjusted temporal graph decoder. It updates temporal relations between video features. Most methods that use pre-trained 2D CNN for extracting spatial information do not preserve the spatial structure. To alleviate this issue, motion-guided spatial attention (MGSA)~\cite{chen2021towards}, which is an adaptive spatial attention mechanism, was proposed. It aggregates the extracted spatial feature map by a spatial attention map. Specifically, motion is used as guidance for spatial attention, and a gated recurrent attention unit is devised for establishing the relation between temporal maps. In addition, a hierarchical visual textual graph is constructed to extract rich semantic information.
 \subsection{Graph-based Methods}
\label{Sec:GRA} 
Graphs are data structures consisting of nodes and the nodes are linked by edges. The nodes represent entities such as objects~\cite{wang2018videos} and actions~\cite{tsai2019video}, and the edges represent the interaction among the entities. Graphs are used in the real world for depicting objects and their association in various fields such as social networks, e-commerce networks, biology networks, traffic networks, etc.,~\cite{zhang2020deep}. The objects in the video and the relationships between them have been proposed for achieving the visual tasks of action recognition~\cite{wang2021actionclip}, and video retrieval~\cite{cao2022visual}. Existing \ac{VC} methods generate captions based on either a scene or object level by operating directly on raw pixels of \ac{2D-CNN} or \ac{3D-CNN} to capture the higher level of semantic interactions. Their main limitation is that they rely completely on the attributes of objects and ignore the cross-object transformations~\cite{wu2019learning} or object dynamics in temporal framework~\cite{wu2019learning,zhou2019grounded}. While graph-based methods operate on the correlation of higher-level semantic entities and their interactions\cite{hou2020joint}. MaskTrackRCNN~\cite{yang2019video} can concurrently perform object detection, instance segmentation, and object tracking. The method is applied to m-frame video to obtain the object trajectories. Graph-based methods employ tracking on \ac{VC} algorithms to extract the object trajectories in the temporal domain.\par
The following graph-based approaches are used in video captioning. They are given below:

\subsubsection{Temporal graph and Bidirectional Temporal graph based \ac{VC}}
 Since the objects in the video across consecutive frames are dynamically changing in shape and location over time, so it is required to “trace” the objects’ temporal trajectories for capturing the detailed trajectories of intra-frames of the video content. Since the objects do not appear in all the frames of the video, it gives rise to challenges for trajectory capture. To address this issue, object alignment in two directions is required. The bidirectional temporal alignment is used to describe two temporal trajectories for each object if objects are visible only in a few frames of video.
\ac{OA-BTG}~\cite{zhang2019object} is proposed to exploit the salient objects with their detailed temporal dynamics and represent them using discriminating temporal representations by performing object-aware local feature aggregation on detected object regions. The \ac{BTG} comprises both forward and reverse-order graphs along the temporal order, that capture temporal trajectories of complementary information for each salient object instance. The capture of the detailed temporal dynamics for objects in the global context accurately describes the fine-grained captions. Object aware aggregation for each salient object instance by using a learnable \ac{VLAD} model on the temporal trajectories to obtain discriminative representation. In object aggregation (OA), two \ac{VLAD} models are used to learn spatiotemporal correlations of objects at local and global levels to encode discriminative representations. The attention mechanism is a hierarchical structure comprising two attention mechanisms i.e., temporal attention and object attention. The temporal attention attends object regions at different time steps and represents $N$ object VLAD representations into one representation. Object attention is used to describe to discriminate the contribution of different object instances. Zhu et al.~\cite{zhu2020ovc} designed a video-based object-oriented \ac{VC} network (OVC)-Net via a temporal graph and detail enhancement. They used mask-RCNN for object isolation, segmentation, and tracking. In addition, an object-oriented temporal graph is constructed to understand the \ac{ST} evolution of the motion trajectory in different frames. They also used a detail enhancement module for capturing the distinct features among the different objects to enhance the accuracy of the \ac{VC}. This technique is the first scheme that generates holistic descriptions of detailed descriptions for specific objects under small samples. If the scenes and objects in the video are dynamic in a relatively long proposal, the existing methods cannot interpret the scene evolution within the temporal proposal framework. 

\subsubsection{Spatio-Temporal graph-based \ac{VC}}
Spatio-temporal graphs consider trajectories of objects in inter-frames and intra-frames. Zhang et al.~\cite{zhang2019object} utilized
a \ac{BTG} and object-aware aggregation in which a trainable \ac{VLAD} is employed for the object feature aggregation.
Xiao et al.~\cite{xiao2020video1} exploited spatial-temporal information of the videos in the framework of \ac{VC} by constructing \ac{TGN} and \ac{RGN}. \ac{TGN} focuses on the sequential information of frames while \ac{RGN} is designed to exploit the relationships among salient objects within a given frame. \ac{GCN} is used for encoding frames with their sequential information and building a region graph for utilizing object information. A stacked \ac{GRU} decoder with a coarse-to-fine structure for caption generation. \ac{OA-BTG}~\cite{zhang2019object} utilized
a \ac{BTG} and object-aware aggregation in \ac{VC}. Compared with it, \ac{OSTG} ~\cite{zhang2020video} further explores the intra-frame spatial relationships among objects along with a \ac{BTG}
to improve video captioning accuracy. A trainable \ac{VLAD} is employed for the object feature aggregation.
\subsection{Multi-granularity graph-based \ac{VC}}
GLMGIR~\cite{yan2019fine} is a graph-based fine multi-granularity interaction representation technique to model fine-granular actions. The model was designed and evaluated to model team interactions in sports. Three types of interactions are observed, 1) local interactions emphasizing joints of humans, 2) semi-local interactions which are among two or three players; 3) global interactions that reflect a strategic move in offense or defense. GLMGIR~\cite{yan2019fine} develop a systematic interaction representation learning method that encodes different granularity of interactions jointly and well explores their dependency structure. Multi-granular attention Module (MAM) is used to generate attention weights to choose the proper contextual temporal regions for the subsequent sentence generation. On the other hand, it also generates granularity weights that dynamically select the correct/proper granular feature that maps to the specific sentence. \ac{GMMP}~\cite{li2022graph} is meta learning model based on \ac{GCN} that uses \ac{POS} information for \ac{VC}. It uses \ac{GCN} to model the temporal dynamics in the video for capturing temporal structure from streaming frames. Specifically, a simple graph convolution (SGC)~\cite{wu2019simplifying} is used to reduce the complexity of the conventional \ac{GCN}. In addition, a multi-granularity \ac{POS} attention mechanism is introduced to use \ac{POS} information in both word and phrase levels for generating high-quality sentences. In order to avoid deteriorating performance during tests, \ac{RL} is used to optimize sentence-level metrics by adopting a meta-learning strategy.

\subsubsection{Semantic graph-based \ac{VC}}
Semantic graphs are used in \ac{VC} to produce meaningful captions. This problem is addressed in by Zhang et al.~\cite{zhang2020dense} who proposed the \ac{GPaS} framework. In the graph, the semantic words are used as nodes, and the interactions of objects are as edges by coupling a \ac{GCN} and a \ac{LSTM} for text summarization. Most video captioning algorithms suffer from the long-tailed problem. In long-tailed problems, content-specific words appear less frequently than common words and function words. Zhang et al.~\cite{zhgobject} proposed \ac{ORG} at the encoder to collect object interaction features to improve the visual representation for \ac{VC}. They employed a \ac{TRL} method to exploit the successful external language model (ELM) to integrate huge linguistic expertise into the caption model. The ELM produces more semantically similar word proposals which extend the ground truth words used for training to curtail the long-tailed problem. Bai et al.~\cite{bai2021discriminative} proposed video summarization by the latent semantic graph. The conditional graph used in this work is not in the conventional form of a semi-positive indefinite affinity matrix. Rather it takes region-level object proposals conditioned on \ac{ST} information of video frames. In this, a dynamic graph is used instead of a static graph that links enhanced object proposals with randomly initialized nodes. In other words, a large number of enhanced object proposals is summarized into high-level visual knowledge with dynamic graphs. The latent semantic visual words are then fed into the language decoder. A language decoder comprises an attention \ac{LSTM} network for weighting dynamic visual words and a language \ac{LSTM} network for producing semantically rich captions. If pre-trained detection models are used in \ac{VC}, it is possible to lose important semantic concepts, since the model is not trained on the video captioning datasets, it would miss some semantic concepts that are not defined during pre-training. Wang et al.~\cite{wang2021cross} proposed Cross-Modal Graph (CMG) that is constructed with the cross-modal meta details. In Cross-modal meta concepts, a weakly supervised learning method is used to identify the correspondence between visual regions of the given words of target captions. In this model three types of cross-modal graphs are used, one is graphs at video level, the second is the graph at frame level and the third is dynamically formed cross-meta graphs to predict better the object interactions and relations in the \ac{VC}. Object relation graph and multi-modal feature fusion \ac{ORMF}~\cite{yan2022multimodal} leverages the relationship between objects in the video and the correlation between multi-modal features. It constructs an object relation features graph using the similarity and Spatiotemporal relationship of objects in video and encodes the object relation using \ac{GCN}. In addition, a multi-modal feature fusion model is constructed for learning the relationship between features of different models and fusing the features. Li et al.~\cite{li2022long} used a long short-term graph (LSTG) to capture short-term semantic relations and long-term transform dependencies. The authors introduced a global gated reasoning module (GGRM) to prevent the relation ambiguity problem in the \ac{VC}. The \ac{Self-Attn} block is replaced with GGRM in the transformer block to fully exploit the object's relations for enhancing the captioning accuracy. In another study, Wang et al.~\cite{wang2022mivcn} proposed a semantic association graph for the analysis of multi-modal interactions. The semantic association graph has two modules: \textit{(i)} semantic association graph module (SAGM), and \textit{(ii)} multimodal attention constraint module (MACM). The SAGM builds the topological graph between semantic attributes and utilizes \ac{GRU} to learn long-range temporal patterns. MACM is used to capture complementary visual features and filter superfluous visual features. A summary and comparison of graph-based \ac{VC} methods are presented in the following and Table~\ref{GARABBASS}. 

\begin{table*}
\centering
\scriptsize
\caption{Summary of graph-based \ac{VC} methods.} \label{GARABBASS}
\begin{center}
\begin{threeparttable}
\begin{tabular}{p{2.5cm}>{\centering}p{0.9cm}>{\centering}p{3cm}cccccccc}
\toprule 
 \multirow{2}{*}{Study} & \multirow{2}{*}{Year} & \multirow{2}{*}{Method} & \multirow{2}{*}{Dataset} & \multicolumn{7}{c}{{\centering}Results} \tabularnewline
\cline{5-11} 
 &  &  &  & METEOR& ROUGE-L & CIDEr& B@1 & B@2 & B@3 & B@4\tabularnewline
\midrule 
\multirow{2}{2.3cm}{Zhang and Peng~\cite{zhang2019object}} & \multirow{2}{*}{2019} & \multirow{2}{*}{OA--BTG\tnote{1}} & MSVD & 36.2 & --& 90.6 &-- &  -- &  -- & 56.9\tabularnewline
\cmidrule{4-11} \cmidrule{5-11} \cmidrule{6-11} \cmidrule{7-11} \cmidrule{8-11} \cmidrule{9-11} \cmidrule{10-11} \cmidrule{11-11} 
 &  &  & MSR-VTT & 28.2 & --  & 46.9 & -- & --  & --  & 41.4\tabularnewline
\midrule 
Zhu et al.~\cite{zhu2020ovc} & \multirow{1}{*}{2020} & OVC-Net\tnote{2} & -- & 20.0 & 45.1 & 50.4 & 47.0 & 33.4 & 25.7 & 20.2\tabularnewline
\midrule 
Zhang et al.~\cite{zhang2020dense} & 2020 & GPaS\tnote{3} & Activitynet & 7.44 & -- & 13.00 & 13.73 & 6.02 & 2.54 & 0.93\tabularnewline
\midrule 
\multirow{2}{2cm}{Bai et al.~\cite{bai2021discriminative}} & \multirow{2}{*}{2021} &  \multirow{2}{*}{D-LSA\tnote{4}} & MSVD & 37.6 & 75.2 & 100.8 & -- & -- & -- & 60.9\tabularnewline
\cmidrule{4-11} \cmidrule{5-11} \cmidrule{6-11} \cmidrule{7-11} \cmidrule{8-11} \cmidrule{9-11} \cmidrule{10-11} \cmidrule{11-11} 
 &  &  & MSR-VTT & 28.8 & 62.3 & 51.2 & --  &--  & --& 44.6\tabularnewline
\midrule  
\multirow{2}{2cm}{Zhang et al.~\cite{zhang2020video}} & \multirow{2}{*}{2021} &  \multirow{2}{*}{OSTG\tnote{5}} & MSVD & 36.8 & {--} & 92.1 & {--} & {--} & {--} & 57.9\tabularnewline
\cmidrule{4-11} \cmidrule{5-11} \cmidrule{6-11} \cmidrule{7-11} \cmidrule{8-11} \cmidrule{9-11} \cmidrule{10-11} \cmidrule{11-11} 
 &  &  & MSR-VTT & 28.6 & {--} & 48.2 & {--} & {--}  & {--}  & 41.9\tabularnewline
\midrule 
\multirow{3}{2cm}{Li et al.~\cite{li2022graph}} & \multirow{3}{*}{2020} & \multirow{3}{*}{GMMP\tnote{6}} & MSVD & 36.5 & 72.6 & 85.6 &  &  &  & 57.2\tabularnewline
\cmidrule{4-11} \cmidrule{5-11} \cmidrule{6-11} \cmidrule{7-11} \cmidrule{8-11} \cmidrule{9-11} \cmidrule{10-11} \cmidrule{11-11} 
 &  &  & MSR-VTT & 28.9 & 63.8 & 54.9  & {--} & {--} & {--} & 45.1\tabularnewline
\cmidrule{4-11} \cmidrule{5-11} \cmidrule{6-11} \cmidrule{7-11} \cmidrule{8-11} \cmidrule{9-11} \cmidrule{10-11} \cmidrule{11-11} 
 &  &  & Charades & 21.8  & 42.5 & 30.3  & {--}  & {--} & {--} & 20.1\tabularnewline
\midrule 
Yan et al.~\cite{yan2022multimodal} & 2022 & ORMF\tnote{7} & MSVD & 37.6 & 73.4 & 93.1 & {--} & {--} & {--} & 56.4\tabularnewline
\midrule 
Wang et al.~\cite{wang2022mivcn} & 2022 & {MIVCN}\tnote{8} & MSVD & 36.4 & 79.1 & 92.1 & {--} & {--} & {--} & 56.8\tabularnewline
\bottomrule
\end{tabular}
\begin{tablenotes}
\item[1] Object-aware aggregation with bidirectional temporal graph;
\item[2] Object-oriented Video Captioning Network;
\item[3] Graph-based Partition-and-Summarization;
\item[4] Discriminative Latent Semantic Graph;
\item[5] Object-aware spatio-temporal Graph;
\item[6] Graph Convolutional Network (GCN) via Meta-learning with Multi-granularity Part-of-Speech;
\item[7] Multimodal Feature Fusion based on Object Relation;
\item[8] Multimodal Interaction Video Captioning Network based on Semantic Association Graph.
\end{tablenotes}
\end{threeparttable}
\end{center}
\end{table*}

\subsubsection{Challenges in graph-based \ac{VC}}
\begin{itemize}
\item
Increasing the number of convolution layers degrades the performance. So developing deeper structural patterns of convolution layers in \ac{GCN}s is still an open challenge for the researchers.
\item
Most graph-based video captioning~\cite{zhang2019graph} has relied on static graphs. However, in reality dependancies are often dynamic. To this end, learning static graphs may not provide optimal performance. 
\item
Most of the graph-based \ac{VC} algorithms are based on the aggregation of objects in inter and intra-frames of video and theoretically modeled with a one-dimensional Weisfeiler–Lehman graph isomorphism test. Graph isomorphism networks have reached their limit~\cite{xu2018powerful}. 
\item There is scope for graph neural network-based video captioning algorithms that employ recurrent GNNs, convolutional GNNs, graph autoencoders, and spatial-temporal GNNs.
\item In \ac{VC} object detection-based schemes might result in the generation of unnecessary
object bounding boxes or meaningless relationship pairs. Further, these schemes rely on a ranking of probability for outputting relationships, which will result in semantically redundant relationships~\cite{xu2020survey}.
\end{itemize}
\begin{tcolorbox}[enhanced,attach boxed title to top center={yshift=-3mm,yshifttext=-1mm},
  colback=blue!5!white,colframe=blue!75!black,colbacktitle=red!80!black,
  title=Summary,fonttitle=\bfseries,
  boxed title style={size=small,colframe=red!50!black} ]
  We have comprehensively covered the work on graph-based \ac{VC}. Most of the research used the \ac{MSVD} and/or \ac{MSR-VTT} datasets as benchmarks for evaluation. BLEU,~\ac{METEOR},~\ac{ROUGE-L},~\ac{CIDEr} are used for the evaluation of graph-based video captioning algorithms. The graph-based video captioning can be improved by considering inter-frame and intra-frame object interactions. Not only object features, but considering frame features will improve the performance of graph-based \ac{VC}. Instead of using static graphs, dynamic graphs may depict object interactions well in time and space. The captioning accuracy can be improved by minimizing the loss function. There are several challenges in the graph-based \ac{VC}.
\end{tcolorbox}

\color{black}
\color{black}
\subsection{Reinforcement Learning}
\label{Sec:RELE}
Combining the representational learning power of deep learning with existing \ac{RL} methods can be helpful in \ac{VC} applications~\cite{padakandla2021survey}. 
The main attraction of \ac{RL} is that it automatizes the process which requires no human interference~\cite{goodfellow2016deep}.
In \ac{RL}, an agent updates their positions based on rewards or penalties resulting from actions performed in a given environment. The definition of an \ac{RL} task has four main considerations, the policy, reward, value, and model with which the agents are interacting. The policy defines the agent's behavior at a given instance of time. It determines the actions to be taken in an environment. The reward function defines the goal of \ac{RL}. The values are predictions of the reward. Lastly, the model determines a set of actions to be taken to obtain a future course of action before they are actually experienced~\cite{sutton2018reinforcement}.

\begin{table*}
\centering
\caption{Summary of \ac{RL} task for \ac{VC}.}
\begin{threeparttable}
\resizebox{0.99\textwidth}{!}{\begin{tabular}{>{\raggedright}m{3.6cm}>{\centering}p{0.7cm}>{\centering}p{3.5cm} c c c c c}
\toprule  
\centering{Study} & Year & Method  & Dataset & METEOR & CIDEr & ROUGE-L & B@4\tabularnewline
\midrule
 \multirow{2}{*}{Pasunuru and Bansal~\cite{pasunuru2017reinforced}} & \multirow{2}{*}{2017} & \multirow{2}{*}{CIDEnt-RL\tnote{1}} $\quad$& \multirow{1}{*}{MSR-VTT} & 28.4 & 51.7 & 61.4 & 40.5\tabularnewline
\cmidrule{4-8} 
 &  &  & \multirow{1}{*}{MSVD} & 34.9 & 88.6 & 72.2 & 54.4\tabularnewline
\midrule
\multirow{1}{*}{Phan et al.~\cite{phan2017consensus}} & \multirow{1}{*}{2017} & \multirow{1}{*}{CST\_GT\_none}\tnote{2} & \multirow{1}{*}{MSR-VTT} & 29.1 & 49.7 & 62.4 & 44.1\tabularnewline
\midrule
\multirow{2}{*}{Wang et al.~\cite{wang2018video}} & \multirow{2}{*}{2018} & HRL\tnote{3} & \multirow{1}{*}{MSR-VTT} & 28.7 & 48.0 & 61.7 & 41.3\tabularnewline
\cmidrule{3-8} 
 &  & HRL-16 & \multirow{1}{*}{Charades} & 19.5 & 23.2 & 41.4 & 18.8\tabularnewline
\midrule
\multirow{2}{*}{Li and Gong\cite{li2019end}} & \multirow{2}{*}{2019} & \multirow{2}{*}{E2E\tnote{4}} & \multirow{1}{*}{MSVD} & 33.6 & 86.5 & 70.5 & 48.0\tabularnewline
\cmidrule{4-8} 
 &  &  & \multirow{1}{*}{MSRVTT} & 27.0 & 48.3 & 61.0 & 40.4\tabularnewline
\midrule
\multirow{3}{*}{Zhang et al.~\cite{zhang2019reconstruct}} & \multirow{3}{*}{2019} & \multirow{3}{*}{RecNet\textsubscript{(l+g)}\tnote{5}} & \multirow{1}{*}{MSR-VTT} & 27.7 & 49.5 & -- & 39.3\tabularnewline
\cmidrule{4-8} 
 &  &  & \multicolumn{1}{c}{MSVD} & 34.8 & 85.9 & -- & 52.9\tabularnewline
\cmidrule{4-8} 
 &  &  & ActivityNet & 10.61 & 38.88 & -- & 1.75\tabularnewline
\midrule
\multirow{3}{*}{Wei et al.~\cite{wei2020exploiting}} & \multirow{3}{*}{2020} & \multirow{3}{*}{N/A} & \multirow{1}{*}{MSVD} & 34.4 & 85.7 & -- & 46.8\tabularnewline
\cmidrule{4-8} 
 &  &  & \multicolumn{1}{c}{MSR-VTT} & 26.9 & 43.7 & -- & 38.5\tabularnewline
\cmidrule{4-8} 
 &  &  & Charades & 17.2 & 21.6 & -- & 12.7\tabularnewline
\midrule
Dong et al.~\cite{dong202multi} & 2021 & SRL\tnote{6} & {MSR-VTT} & 31.15 & 61.18  & - & 46.54 \\
\midrule
Praveen et al.~\cite{sv2021exploration}  & 2021 &  WAFTM\tnote{7} & MSVD & 34.2 & 92.43 & 70.8 & 50.4 \\ 
\bottomrule
\end{tabular}}
\begin{tablenotes}
\item[1] CIDEnt-reward \ac{RL}
model;
\item[2] Consensus-based Sequence Training with the Reward-Weighted Cross Entropy;
\item[3] Hierarchical~\ac{RL};
\item[4] End to End with Beam Search;
\item[5] Reconstruction Network;
\item[6] Semantic-Reinforced Learning;
\item[7] Weighted Additive Fusion Transformer with Memory Augmented Encoders.
\end{tablenotes}
\end{threeparttable}
\end{table*}

Two problems in captioning models exist when they are trained to maximize the likelihood of the next the word given the previous ground-truth input words~\cite{phan2017consensus}. This approach has two problems, 1) an objective mismatch problem and 2) an exposure bias problem. The objective mismatch occurs due to the objective function optimized at training time being different from the true objective. Mathematically, it is similar to minimizing a weighted squared error in which accuracy on the least probable captions is given higher priority and therefore poorly fit for applications such as \ac{VC}. The greedy output eventually returned by the system – is given the lowest priority of all during training which may result in erroneous captions. Another problem is the exposure bias problem~\cite{phan2017consensus} in which there is a large disparity between the distribution during training and testing time. During training, the model is only exposed to the word sequences from the training data, while at testing time, the model instead only has access to its own predictions; therefore, whenever the model encounters a state it has never been exposed to before, it may behave unpredictably for the rest of the output sequence. To tackle these two problems \ac{RL} based policy gradient is used in the framework of \ac{VC}.

\paragraph{\ac{VC} techniques employing policy--gradient \ac{RL}}
Pasunuru and Bansal~\cite{pasunuru2017reinforced} proposed \ac{RL} based on a policy gradient algorithm~\cite{williams1992simple}. In this work, an \ac{RNN} can be considered as an agent, which interacts with the complex environment (the textual and video context every time step). The parameters of the agent defined a policy, whose implementation marks the agent picking an action. In the sequence generation setting, an action refers to predicting the next word in the sequence at each time step. After taking an action the agent updates its internal state (the hidden units of \ac{RNN}). Once the agent has reached the terminal condition, it obtains a reward. The authors approximated the gradients via a single sampled word as in~\cite{ranzato2015sequence}. Hence, for training, the authors employed a mixed loss function, which is a weighted combination of the cross-entropy loss (XE) and \ac{RL} loss. The authors employed a decomposable, intra-sentence attention model of~\cite{ parikh2016decomposable}. The reward employed in the work, an entailment score to correct the phrase-matching metric (CIDEr), and the reward is used as a CIDEr-entailment score. The training error is minimized by taking the negative reward function as the objective function, which is given by:
\begin{equation}\label{key0}
L(\theta)= - \mathbb{E}_{w^{s}}  \sim  \rho_{\theta}\left[ r\left(w^{s} \right) \right], 
\end{equation}
\noindent where $w^{s}$ is the word to be sampled, $\theta$ is the model parameters, $\rho_\theta$ is the policy. The gradient of this loss function is defined as:
\begin{equation}\label{key222}
\nabla _{\theta}L(\theta)= - \mathbb{E}_{w^{s}}  \sim  \rho_{\theta}\left[ r\left(w^{s} \right) \nabla _{\theta}log\left( \rho_{\theta}\left(w^{s} \right)   \right) \right].
\end{equation}
In practice, gradients estimated based on \eqref{key222} are unstable and variance reduction is performed using a baseline $b$ as
\begin{equation}\label{key223}
\nabla _{\theta}L(\theta)= - \mathbb{E}_{w^{s}}  \sim  \rho_{\theta}\left[ r\left(w^{s}-b \right) \nabla _{\theta}log\left( \rho_{\theta}\left(w^{s} \right)   \right) \right].
\end{equation}
Phan et al.~\cite{ phan2017consensus } formulated a Consensus-based Sequence Training (CST) scheme to produce video captions. It is an alternative to \ac{RL} but uses the multiple existing training-set captions in a novel way. Firstly, CST performs an \ac{RL} like pre-training, but with captions from the training data replacing model samples. This alleviates the objective mismatch issue existing in cross-entropy loss estimation. Second, CST applies \ac{RL} for fine-tuning using average \ac{CIDEr} among training captions as the baseline b. This fine-tuning additionally removes the exposure bias problem. The two stages of CST allow objective mismatch and exposure bias to be assessed separately and together establish a new state-of-the-art task in \ac{VC}. Chen et al.~\cite{chen2018less} proposed a plug-and-play \ac{VC} (Picknet) based on \ac{RL}. The reward is given to a subset of frames based on two factors: \textit{(i)} maximizing the discrimination in visual features, and \textit{(ii)} minimizing the disparity between the generated caption and the ground truth. The rewards considered in this method are based on the language reward and visual diversity reward. The rewarded agents will be chosen and the corresponding latent representation of the encoder-decoder framework will be updated for future trials. The training is carried out in three phases. In the supervision phase, the encoder and decoder are pre-trained and in the second phase, \ac{RL} is used to pick key-frames. In the final phase, picknet and encoder-decoder framework are trained with the cross-entropy loss. The results show that the selection of key-frames does not deteriorate the captioning accuracy. End to End learning in video captioning is complex due to input video and output captions being lengthy sequences. The end-to-end video sequences are both memory-consuming and data-hungry, making it extremely hard to
train. Li and Gong~\cite{li2019end} proposed to multitask \ac{RL} approach for training the E2E \ac{VC} model. The \ac{RL} is introduced to regulate the search space of the E2E neural network, from which an E2E \ac{VC} model can be found and generalized to the testing phase. \color{black} Zhang et al.~\cite{zhang2019reconstruct} introduced a reconstructor block in the existing encoder-decoder framework. This block captures latent representations from the encoder to the decoder (forward flow) and similarly from the encoder to a decoder (backward flow). The reconstructor reconstructs the global or local framework of the video using the intermediate hidden state pattern of the decoder as input. Instead of using log-likelihood which is negative to bring about an accurate description sentence, the authors employed a reward function based on policy gradient using \ac{RL} algorithm. The reconstructor is formed by using \ac{LSTM}s. The encoder-decoder approach is viewed as an "agent" that communicates with the environment, while the video descriptions and sentences are regarded to form the "environment". The policy $\pi_{theta}$ predicts a word at each step $t$ for agent \ac{LSTM}s and then updates "state," which refers to the hidden states and cell states of the agent. The authors used \ac{CIDEr} as the reward. \color{black} \par
RNN encoder used in \ac{VC} are used to encode a video along the forward direction (from start to end), the backward direction, or in both. In the sequential \ac{RNN} structure, the current input signal can be attenuated according to the previous hidden state. Thus, in the presence of noisy information at the beginning of a sequence, it has a negative effect on encoding the key information that appears after the noise. To alleviate this problem, Shi et al.~\cite{shi2019watch} proposed an iterative video encoding scheme to predict the key frame and encode a video based on its keyframe, and then integrate it within the captioning model. They combined \ac{RL}-based training method to jointly train the video refocusing encoder with the captioning model in an end-to-end manner. Gui et al.~\cite{gui2019semantic} proposed a semantic-assisted encoder and decoder framework with \ac{RL}. In this work, the authors employed an attention mechanism at the decoder side which pays more attention to the key frames while producing every word and uses a \ac{CIDEr} based reward policy gradient \ac{RL} in training to produce refined captions, Xu et al.~\cite{xu2020deep} proposed a novel deep reinforcement polishing network \ac{RL} based \ac{VC} inspired by the proofreading in editing the text by human beings. They employed a reward in deep \ac{RL} to optimize the global quality of produced sentences. The semantic gap between visual and text is reduced by revising the word errors and grammar errors. by iteratively polishing the generated caption sentences. For better long-sequence generation, the long-term reward in deep \ac{RL} is adopted to directly optimize the global quality of caption sentences. To reduce the semantic gap between the visual domain and the language domain, the caption candidate is considered an additional cue for \ac{VC}, which is gradually updated by revising the word errors and grammar errors. Xiao and Shi~\cite{xiao2019diverse} proposed a model to select discriminant attributes for \ac{VC}. If attributes are poor, the generated captions may be erroneous. To alleviate the problem, and to enhance the model's ability to filter redundant attributes, a reward-based \ac{RL} loss is designed to update the decoding components as:\par
\begin{equation}
L_{re}=-\mathrm{E}\left(R_{T}|S_{\theta}\right) =- \sum_{w_{1}^{,}   \in \gamma}\pi_{\theta}\left(w_{1}^{,}| S_{0} \right)  Q_{\pi_{\theta}} \left( S_{0},w_{1}^{,}\right),
\end{equation}
\noindent where $S_{0}$ is the initial state, $R_{t}$ is the reward for a complete sequence, $w_{1}^{,}$ is the produced token, $\gamma$ is the vocabulary, $\pi_{\theta}$ is the generator policy which influences the action that selects the next token, and $Q_{\pi_{\theta}} \left( S_{0},w_{1}^{,}\right)$ indicates the action-value function of a sequence.
\color{black}

\paragraph{\ac{VC} techniques employing Manger--worker--critic algorithm \ac{RL}}
Wang et al.~\cite{wang2018video} proposed a hierarchical \ac{RL} algorithm. This algorithm comprises of three components: a low--level worker, a high-level manager, and an internal critic. The manager sets goals for the worker to accomplish, and the worker generates a word for each time step by following the goal set by the manager. In \ac{VC} task the manager asks the worker to generate fluent and meaningful sentences, and the worker generates the corresponding words in the next few time steps in order to fulfill the job. The internal critic determines if the worker has accomplished the goal and transmits a binary segment signal to the manager to help it update goals. The whole pipeline terminates once an end-of-sentence token is reached. In this model, both the manager and the worker are equipped with a soft attention module to capture better the temporal dynamics. In training, the manager policy is made deterministic, and the worker policy is stochastic.The manager policy and is denoted by $g_{t}$=$\mu \theta_{m}(S_{t})$ and worker policy is denoted by $\pi \theta_{w}(a_t,s_t,g_t)$. The worker and manger are trained individually, while training worker goal exploration is ceased. Similarly, when training the manager, the authors consider workers with the Oracle behavior policy. Therefore captions are generated by 
greedy decoding. Xu et al.~\cite{xu2020deep} formulated a reinforcement polishing network to polish the caption and grammar by introducing word denoise network and grammar checking network in the said model. The main problem in deep learning based \ac{RL} algorithms is that at encoder, which is impractical to define a trajectory-level reward function which is rich. In general, feedback is minimal and few activities have non-zero returns. Additionally, it is difficult to generate the reward by contrasting the current state with the preferred result. Hence, it is required to exploit sampling strategy to extract key frames in video that can generate meaningful sentences.

Ranzato et al.~\cite{ranzato2015sequence} provided a solution to problems for generative models in generation of text such as exposure bias including loss functions in the sequence model using \ac{MIXER}~model. In an incremental learning approach, the random strategy of \ac{RL} is changed with the cross-entropy trained model's optimal policy, which involves gradually introducing the model to more and more of its own predictions. Quian et al.~\cite{qian2021filtration} proposed a deep RL algorithm based on actor-critic. The fundamental idea behind actor-double-critic is that an agent's behavior is influenced by both their own personality and their exterior surroundings. The uncertain reward and inadequate feedback in training of conventional \ac{RL} is avoided and the advantage of actor-double-critic is that it gives steady feedback after every action. Sentences are produced by combining the key frames with the acCCN. The feature combination function in the \ac{CCN} provides a merging of visual features that results in an effective semantic modeling for the \ac{VC} issue.

Zhang et al.~\cite{zhang2019reconstruct} introduced reconstructor block in the existing encoder-decoder framework. This block captures latent representations from encoder to decoder (forward flow) and similarly from encoder to decoder (backward flow). The reconstructor reconstructs the global or local framework of the video using the intermediate hidden state pattern of the decoder as input. Instead of using log likelihood which is negative to bring about the accurate description sentence, the authors employed a reward function based on policy gradient using RL algorithm. The reconstructor is formed by using \ac{LSTM}s. The encoder-decoder approach is viewed as an "agent" that communicates with the environment, while the video descriptions and sentences are regarded to form the "environment". The policy $\pi_{ theta}$ predicts a word at each step $t$ for agent \ac{LSTM}s and then updates "state," which refers to the hidden states and cell states of the agent. The authors used \ac{CIDEr} as the reward.\par
In addition, Xiao and Shi~\cite{xiao2019diverse} proposed a model to select discriminant attributes for \ac{VC}. If attributes are poor, the generated captions may be erroneous. To alleviate the problem, and to enhance the models ability of filtering redundant attributes, a reward-based \ac{RL} loss is designed to update the decoding components as:\par
\begin{equation}
L_{re}=-\mathrm{E}\left(R_{T}|S_{\theta}\right) =- \sum_{w_{1}^{,}   \in \gamma}\pi_{\theta}\left(w_{1}^{,}| S_{0} \right)  Q_{\pi_{\theta}} \left( S_{0},w_{1}^{,}\right),
\end{equation}
\noindent where $S_{0}$ is the initial state, $R_{t}$ is the reward for a complete sequence, $w_{1}^{,}$ is the produced token, $\gamma$ is the vocabulary, $\pi_{\theta}$ is the generator policy which influences the action that selects the next token, and $Q_{\pi_{\theta}} \left( S_{0},w_{1}^{,}\right)$ indicates the action-value function of a sequence.

\subsection{Adversial Networks}
\label{Sec:ADAU}
\ac{GAN}~\cite{goodfellow2014generative} models are widely used in computer vision applications such as image-to-image translation~\cite{emami2020spa}, text-to-image synthesis~\cite{zhang2017stackgan}, medical image generation~\cite{frid2018gan}, synthetic image datasets~\cite{zhang2017physically}, text-to-image translation~\cite{bayat2020inverse} video prediction~\cite{hu2019generative} and
3D object generation~\cite{wu2016learning}. There are different flavors of \ac{GAN} such \ac{CGAN}\cite{mirza2014conditional}, stackGAN\cite{zhang2017stackgan}, SeqGAN~\cite{yu2017seqgan}, rankGAN\cite{juefei2019rankgan}, leaky GAN\cite{hou2022semi}, MAS-GAN~\cite{zhang2022unifying} that have been all be employed in various computer vision applications. \ac{GAN}s are unsupervised neural network comprising two networks, i.e., $G$ and $D$. The $G$ tries to produce real data given input is a random noise variable with the probability distribution of $p_{z}{(z)}$ and it maps it to $p_{g}$ over $x$ by a mapping function $G(z;\theta_{g})$ where $G$ is a neural network with model parameters $\theta_{g}$. The \ac{D}, $D(x; \theta_{D})$ outputs a binary signal to represent to distinguish the real data from fake data as accurately as possible. $\theta_{D}$ is the model parameters for the $D$. The $D$ is tried to maximize the possibility of assigning correct labels to both the training samples and the synthetic data coming from $G$. The $G$ is tried to minimize $log(1-D(G(z)))$ and the objective function $V(G,D)$ is
\begin{multline}
V(G,D)= \\
\underset{G}{min} \; \underset{D}{max}\mathbb{E}_{x\sim p(g)}\left|  logD(x)\right| \\ + \mathbb{E}_{z\sim p_{z}(z)}\left| log(1-D(G(z))\right| 
\label{gan2}
\end{multline}

The text descriptions in \ac{GAN}-based \ac{VC} models comes with two major hurdles due to the special nature of the representation of the sentences. In image generation, the transformation is from random noise to image via a deterministic continuous mapping, whereas in text generation, a sequential sampling procedure is required in which discrete tokens are sampled in each step~\cite{huszar2015not}. This procedure is non-differentiable which makes back-propagation infeasible; therefore, policy-gradient \ac{RL} is used. The second problem in the conventional \ac{GAN} setting the generator, $G$, receives feedback from the evaluator when an entire synthetic image is produced. This procedure leads to several problems in training, such as vanishing gradients and error propagation. To prevent such problems, a mechanism is used that allows the $G$ to get early feedback calculated based on approximated expected future reward through Monte Carlo rollouts~\cite{yu2017seqgan}. These two hurdles for captioning are solved by Dai et al.~\cite{dai2017towards} in \ac{IC}. Compared with images, video has rich content and additional temporal dimension, which make it very difficult to extrapolate the algorithms developed for the images for the video captioning problem~\cite{yang2018video}.
 
The \ac{VC} based video captioning is categorized into three types.
\subsubsection{\ac{VC} using Adversarial \ac{GAN} }
Yang et al.~\cite{yang2018video} published the first work on video captioning using a \ac{GAN}. In their approach, an \ac{LSTM} is used in the caption generation and is expanded by adding a $D$ module which acts as an adversary with respect to sentence generation. The goal is to maximize the conditional probability of an output sequence $(y_{1}, y_{2} \cdots y_{m})$ given an input sequence
$(v_{1},v_{2}. \cdots, v_{n})$. The conditional probabilities over the sentences can be defined as follows:
	\begin{equation}\label{vvvv11}
	p\left(y|v \right)=p\left(y_{1}, y_{2} \cdots y_{m} | v_{1}, v_{2} \cdots v_{n}\right). 
	\end{equation}
 The authors then used the \eqref{vvvv12} as an objective for faster convergence and are given by
\begin{multline}
\label{vvvv12}
p\left(y|v \right)=p\left(y_{1}, y_{2} \cdots y_{t1} | v_{1}, v_{2} \cdots v_{t}\right)\\
=\stackrel[i=1]{t}{\prod} P(y_{i}| v, y_{1},y_{2}, \cdots y_{m-1}),
\end{multline}
where $t$ and $t1$ are the length of the video and the generated text. In $G$, the authors used \ac{LSTM} as the encoder and \ac{CNN} as a decoder. Soft-argmax function~\cite{luvizon2019human} is employed in the last layer of the $G$ which multiplies the exponential term in the soft-max function with a very large constant value. The multiplication makes the values either 0 or 1. Then the obtained array values are multiplied with the array index and summed, which provides the index of the array with the maximum value in the array, thus performing the arg-max on the output. The design used in encoding and decoding of~\cite{yang2018video} overcomes the training difficulties that arise in the works of~\cite{goodfellow2014generative,salimans2016improved}. They employed \ac{Soft-Attn} in the $G$ network to improve the accuracy of \ac{VC}. \\ 
Pan et al.~\cite{pan2017create} proposed a \ac{VC} based on \ac{TGANs-C}. The inputs to the $G$ are the noise vector plus caption embedding. The authors made a $D$ network that is very effective not only able to discriminate a fake from a real, one but also in testing whether the video correctly matches the caption by incorporating video, motion, and frame discriminators. The authors incorporated three losses namely-video-level and frame-level matching-aware loss to correct the label of real or fake video/frames and align video/frames with the correct caption, respectively, and temporal coherence loss to emphasize temporal consistency. The authors in~\cite{xiao2019diverse} paid attention to sentence diversity by utilizing the \ac{CGAN}. study~\cite{zhu2020understanding} proposed \ac{STraNet}, an object-based \ac{VC} algorithm using adversarial learning. The \ac{STraNet} shows the ability to represent precisely by describing concurrent objects and their activities in detail. The \ac{STraNet} model added an adversarial $D$ to the caption generation to enhance the inter-relationship between the vision to text.

\subsubsection{\ac{VC} using adversarial \ac{GAN}s with policy gradient \ac{RL}}: As the caption contains discrete tokens, it is difficult to apply the backpropagation directly. To alleviate this, the GAN system is regarded as an adversarial learning system with \ac{RL} used to update the parameters of \ac{G}.
Yan et al.~\cite{yan2021video} developed \ac{VC} by using conventional \ac{GAN}. The generator employs a \ac{CNN}-\ac{RNN} framework, combined with a hierarchical attention mechanism based on object and frame level. In the $D$, the first network is used to extract the difference between the real from the fake, and the second network is used to compare whether the generated caption matches the video content, which aims to make the generated captions consistent with the video content. The policy gradient \ac{RL} algorithm is used to update the parameters of the $G$.
Xiao and shi~\cite{xiao2022diverse} proposed a diverse captioning module that comprises two parts. To obtain high-quality descriptions the model is trained with \ac{XE} loss with \ac{LSTM}. Followed by a bidirectional-\ac{LSTM} for better visual representation Finally, they employed temporal attention and at last, used a hierarchical \ac{LSTM} to generate descriptions based on visual features and text attention. \ac{CGAN} is employed and the inputs to the \ac{GAN} is a latent variable produced by noise vector and the outputs of text and visual attention networks $L_{t}$. This captioning scheme receives a reward at the end. similar to Yan et al.~\cite{yan2021video}, the authors used The policy gradient \ac{RL} algorithm. The objective of the $G$ is to generate a sequence of words $y_{1:T}$=$\left\lbrace y_{1}^{'}, y_{2}^{'},\cdots, y_{n}^{'}\right\rbrace $ from the start state $s_{0}$ to maximize the expected end result:
\begin{equation}\label{key1}
\mathbb{E}\left[R_{T}|s_{0},\theta  \right] =\sum_{y_{l}^{'}\in \gamma}G_{\theta}\left(y_{l}^{'}|s_{0} \right) Q_{G_{\phi}}\left(s_{0},y_{l}^{'} \right) 
\end{equation}
where $s_{0}$ is the initial state, $R_{T}$ is the reward at the end of a sentence, $\gamma$ is the vocabulary, $G_{\theta}$ is the $G$ policy which decides the anticipating word, $QG_{\theta}(s,a)$ are the action value pair of the sequence. The $D$ considers the probability of the real as a reward. 
\begin{equation}\label{key2}
QG_{\theta}(s= y_{1:T-1}, a=y_{t}^{'}=D_{\eta}(y_{1:T},\hat{v})
\end{equation}
where $\hat{v}$ is the visual features considered. As $D$ provides a reward value for the end of the sequence. For obtaining the long-term reward at each time step, the objective function of previous tokens (prefix) and also the result of future outcomes are considered. It is similar to giving up intermediary results for the final victory~\cite{silver2016mastering, yu2017seqgan}. In order to realize this~\cite{xiao2022diverse} proposed a Monte Carlo rollout for sampling the last T-t tokens to evaluate the action value Q $G_{\theta}(s= y_{1:T-1}, a=y_{t}^{'})$ given by \eqref{eq35}.

\begin{align}
                \begin{cases}{ll}
                  \frac{1}{K} \sum_{n=1}^{K}D_{\eta}(y_{1:T}^{n},\hat{v}), (y_{1:T}^{n} \in MC(y_{1:T}; K) & \text{if}~~ t < T\\
                  D_{\eta}(y_{1:T},\hat{v}) & \text{if}~~ t=T
                \end{cases},
             \label{eq35}
\end{align} 
\color{black}
where $y_{1:t}^{n}$ and $y_{t+1:T}^{n}$ are sampled on the Monte Carlo Rollout and the current state. To reduce the instability, the pre-trained network and $D$ have been given a warm start-up. The main contribution of authors is that formulated a new metric called \ac{DCE}. Hua et al.~\cite{hua2022adversarial} proposed a \ac{GAN} based \ac{VC} by exploiting the \ac{GCN}. Mask RCNN is used to detect, segment, and track objects simultaneously. Later motion features of keyframes are selected. Then, a novel object scene relation graph is exploited to describe spatial and temporal details between the objects and between the objects and environments. To mitigate the sequence mismatch problem, a control gate and self-looping structure were added to \ac{GCN} to improve the text correctly representing the video content. A trajectory-based feature representation is used in place of the previous data-driven method to extract motion and attribute information, so as to analyze the object's motion in the time domain and establish the connection between the visual content and language under small datasets. Finally, an adversarial reinforcement learning strategy and a multi-branch discriminator are designed to learn the relationship between the visual content and corresponding words so that rich language knowledge is integrated into the
model. 

\subsubsection{\ac{VC} using Adversarial \ac{GAN} with Manager-worker policy \ac{RL}}
Hemalatha and Chandrasekher~\cite{munusamy2022video} proposed \ac{VC} based on a semantic contextual \ac{GAN}. This model utilizes a $D$ which distinguishes the ground truth description from the generated description for the input video, and a $G$ which produces a description for the given video. The two blocks in $G$ are the manager block that sets goals for the worker and the worker block that produces descriptions based on the goal, visual features, and semantic features extracted from a multilayer perceptron Network (MLP) (as in~\cite{shekhar2020domain}). The $G$ and $D$ parameters are learned through \ac{RL}. A multi-channel \ac{CNN} based discriminator is used. Word2vec~\cite{mikolov2013efficient} embedding vector is used to represent words in the description. The semantically weighted word-embedding vectors are mixed with the features from the 2D-\ac{CNN} and 3d-\ac{CNN} and are fed to the CNN. The features from the $D$ are leaked and fed to the $G$. The manager is trained to generate the goal vector $g_{t}$ which will be used by the worker in sentence generation. The discriminator generates the reward only for a complete sentence. Hence for the intermediate steps the expected reward is generated using Monte-Carlo search. The $G$ and the $D$ are initially trained with the maximum likelihood (ML) method. During ML training, the gradients are computed. While training the $G$, the parameters of the $D$ are fixed and vice versa. The manager and worker are trained in an interleaved manner during the training.
\color{black}
\begin{tcolorbox}[enhanced,attach boxed title to top center={yshift=-3mm,yshifttext=-1mm},
colback=blue!5!white,colframe=blue!75!black,colbacktitle=red!80!black,
  title=Summary,fonttitle=\bfseries,
  boxed title style={size=small,colframe=red!50!black} ]
  To our knowledge there are only five published papers on \ac{VC} using adversarial networks. In all cases, Maximum likelihood estimation is used for the gradient computation and \ac{RL} is used as the text is non-differentiable. The goal-based rewards and semantic-based rewards are used to improve the accuracy of captioning. 
\end{tcolorbox}
\color{black}

\subsection{Non--Autoregressive}
\label{Sec:NoAU}
Autoregressive decoders generate one word at each time step by conditioning on the previously produced words. Most of the works of sequence-sequence based \ac{VC} used autoregressive decoding~\cite{bahdanau2014neural,vaswani2017attention}. The flaw in autoregressive decoding is that it suffers from latency which is not tolerable in a real-time applications. The latency is further amplified if the description at the decoder is fine-grained~\cite {gella2018dataset,kasai2020non,yang2019non}. The latency must be as small as possible in real-time situations such as \ac{NMT}. As an alternative, researchers focused on Non-Auto regressive decoding techniques to achieve significant inference speedup~\cite{wang2019controllable,shao2020minimizing} in which the words are produced in parallel. The speed-up of the scheme is at the expense of performance degradation. The performance degradation is compensated in the works of~\cite{lee2018deterministic,ghazvininejad2019mask,gu2017non} by iterative refining of the sentences conditioned on parts or whole of the previous outputs rather than producing in one-shot. The loophole in these methods, instead of taking $N$ prior concurrent words, a stochastic random input is given as the decoder input. This is leading to translation errors due to insufficient context which could greatly influence future predictions and thereby generates inefficient captions. 


\subsection{Multimodal Video Captioning}
\label{Sec:MM}
Multimodal \ac{VC} concerns the exploitation of latent deep networks pooled from different modalities (audio, speech, video, image, text, source). 
Tian et al.~\cite{tian2019audio} proposed a multimodal CNN (MMCNN) based audio video framework capable of learning decoupled audio-text and visual-text deep feature hierarchies. Tian et al.~\cite{tian2018attempt} designed a multimodal CNN based audio-visual \ac{VC} framework and introduced a modality--aware module for selecting the modality during the sentence generation. Xu et al.~\cite{xu2019semantic} developed a semantic Filtered Soft-split-Aware model to improve \ac{VC} by fusing semantic concepts with audio-augmented features extracted from the input videos. Moreover, several topic-guided \ac{VC} methods that jointly create topics and topic-oriented captions using multimodal features in a multi-task framework have been proposed~\cite {chen2017generating, chen2017video, chen2015microsoft}. Incorporating multi-modality information within a sequence-sequence model has been employed to leverage information from both audio and visual modalities for improving the efficiency of captioning~\cite{ramanishka2016multimodal,jin2016describing,joshi2019tale}. Varma et al.~\cite{varma2021efficient} proposed a multi-modal \ac{VC} model that uses the audio, external knowledge, and attention mechanism to enhance the captioning process. In~\cite{zheng2021stacked}, the historical information along with motion and object features using the stacked multi-modal attention network are used for \ac{VC}.

\subsection{Dense Video Captioning}
\label{Sec:dvc}
One downside of single sentence \ac{VC} methods~\cite{gao2017video, pei2019memory,qi2018sports} is that they generate sentences with very little information. Dense Video Captioning (\ac{DVC})~\cite{xu2019joint,suin2020efficient} has emerged to detect and describe events via a richer storytelling approach~\cite{krishna2017dense}, which makes it useful in applications such as \ac{CBIR} and video recommendation~\cite{deng2021sketch}. The \ac{DVC} can be divided into: \textit{(i)} event proposal blocks, and \textit{(ii)} caption generation proposal blocks. Indeed, the first \ac{DVC} proposed by~\cite{krishna2017dense} in which the event proposal module spots events with a multi-scale version of deep action proposals for understanding action and denotes them by LSTM hidden states. To incorporate future context along with past context, Wang et al.~\cite{wang2018bidirectional} proposed a novel bidirectional process to encode both past and future context for localizing event proposals. Early \ac{DVC} methods considered event proposal and caption generation blocks as independent entities. The two blocks are either trained separately or alternatively, which in turn affects the generated descriptions. To address this problem, Zhou et al.~\cite{zhou2018end} proposed an end to end \ac{DVC} with a masked transformer. Specifically, a differentiable masking scheme is employed to ensure consistency between the two blocks during training. Transformers have been used in \ac{DVC} to alleviate the limitations of RNNs when modeling long-term dependencies in videos. Despite their superior accuracy, \ac{VC} techniques employing masked transformers~\cite{zhou2018end} suffers from unsatisfactory run time performance. To directly address this, Yu et. al.~\cite{yu2021accelerated} proposed an Accelerated Masked Transformer which reduced run time by $2 \times$ when compared to the reference Masked Transformer model. The Accelerated Masked Transformer~\cite{yu2021accelerated} introduced a lightweight anchor-free event proposal integrated with a local attention mechanism stage and the single-shot feature masking strategy along with an average attention mechanism in the caption stage.  Lu et al.~\cite{lu2021environment} proposed a lightweight \ac{DVC} model based on the transformer framework to improve execution efficiency for video-caption generation on edge cameras. They used an attention mechanism to hierarchy extract the most relevant visual and text features for generating captions. The attention mechanism consists of an object attention module and event attention module. 



WS-DEC~\cite{chen2021towards}, which is a weakly supervised \ac{DVC}, improved the captioning accuracy where temporal boundary annotations are not available. The conventional \ac{DVC} considers the event localization proposal and caption generation proposal in a feed forward manner. In contrast, in WS-DEC, there is a mutual exchange of information between the event localization and captioning modules. The event localizer sends two types of information to the caption generation block. The first information is pertaining to the key words learned during video-text alignment to improve the captioning accuracy and the second type of information is by introducing an induced set attention block, the concept features are extracted and thus WS-DEC can obtain richer information compared to the previous works. 

\ac{SGR}~\cite{deng2021sketch}, first, describes the whole video by generating multi-sentences paragraph, and then improves the quality of the refining stage by refinement enhanced training and dual-path cross attention. \ac{ViSE}~\cite{aafaq2022dense} exploits both visual information and linguistic content in the event detection and caption generation process. After extracting $n$-grams along with their associated weights, the caption embedding is learned and semantic space is constructed by a Sen2Vec model. Then, a visual-semantic joint embedding network (VSJM-Net) is used to conceptualise the visual information and the semantic space. Lu and Han~\cite{lu2021environment} proposed a lightweight \ac{DVC} method using transformer which is made robust to environmental changes (i.e., input drift). Keeping the advances in pre-trained multimodal networks, Zhu et al.~\cite{zhu2022end} formulated \ac{DVC} task as a single sequence-to-sequence network using the transformers. Table~\ref{SUDVCVC} provide a summary of several studies on \ac{DVC} task for \ac{VC}.

\begin{table*}[htbp]
\centering
\caption{Summary of \ac{DVC} task for \ac{VC}.} \label{SUDVCVC}
\begin{threeparttable}
\begin{tabular}{>{\raggedright}p{2.8cm}>{\centering}p{1cm}>{\centering}p{3cm}>{\centering}p{3cm}cccc}
\toprule 
\textbf{Study}  & \textbf{Year} & \textbf{Method} & \textbf{Dataset}  & \textbf{METEOR} & \textbf{CIDEr} & \textbf{ROUGE} & \textbf{B@4}\tabularnewline
\midrule
Krishna et al.~\cite{krishna2017dense} & 2017  & {CIPF}\tnote{1}& ActivityNet Captions& 9.46  & 24.56 & {--} & 3.98  \tabularnewline
\midrule 
Shen et al.~\cite{shen2017weakly} & 2017 & {LFCN}\tnote{2} & \ac{MSR-VTT} & 28.3  & 48.9  & 61.1  & 41.4 \tabularnewline
\midrule
\multirow{2}{4cm}{\raggedright{Li et al.\cite{li2018jointly}}} & \multirow{2}{1cm}{ \centering{2018}} &{DVCGT proposals\tnote{3}} &  \multirow{2}{2cm}{\centering{ActivityNet captions}} & 10.33  & 25.24 & {--}  & 1.62\tabularnewline
\cmidrule{3-3} \cline{5-8} \cline{6-8} \cline{7-8} \cline{8-8} 
&  &{DVCLT proposals}\tnote{4} &  & 6.93 & 12.61 & {--} & 0.73\tabularnewline
\midrule
Zhou et al.\cite{zhou2018end} & 2018 & Transformer model & ActivityNet Captions& 9.56  & {--} & {--} & 2.23 \tabularnewline
			\midrule
Wang et al.~\cite{wang2018bidirectional} & 2018 & {BSST}\tnote{5} & ActivityNet Captions& 9.65  &  {--} &  {--} &  {--} \tabularnewline
		\midrule
Zhang et al.~\cite{zhang2020dense} & 2020 & {GPas}\tnote{6}& ActivityNet Captions& 11.04  & 28.20 & 21.30 & 1.53 \tabularnewline
			\midrule
Wang et al.~\cite{wang2020event} & 2020 & {TL-NMS}\tnote{7} & ActivityNet Captions& 10.58  & 39.73 &  {--} & 1.96 
		\tabularnewline
			\midrule
Chadha et al.~\cite{chadha2020iperceive} & 2020 & {iPerceive \ac{DVC}}\tnote{8} & ActivityNet Captions& 12.27  &  {--} &  {--} &  2.98 \tabularnewline
			\midrule
Wu et al.~\cite{wu2021weakly} & 2021 & {Weakly supervised ECG}\tnote{9} & ActivityNet Captions& 70.6 & 14.25 &  {--} & 1.33 \tabularnewline
			\midrule
Chen et al.~\cite{chen2021towards} & 2021 & {WS-DEC}\tnote{10} & ActivityNet Captions& 7.03   & 19.53 & 12.79 & 1.23 \tabularnewline
			\midrule
Lee and Kim~\cite{lee2021dvc} & 2021 & \ac{DVC}-Net & ActivityNet Captions& -- & 15.80 &  {--} & 1.26 	\tabularnewline
			\midrule
Wang et al.~\cite{wang2021end} & 2021 & {P\ac{DVC}}\tnote{11}  & ActivityNet Captions& 15.93 & 27.27 &  {--} & 11.80 \tabularnewline
			\hline
\multirow{2}{2.2cm}{\raggedright{Yu and Han~\cite{yu2021accelerated}}} & 2021 & \multirow{2}{2.5cm}{AMT (L=3, d=1024)\tnote{12}} & ActivityNet Captions & 5.82 & 10.87 & {-- } & {-- }\tabularnewline
\cmidrule{4-8} 
 &  &  & Youcook II  & 2.43 & 4.88 & {-- } & {-- } \tabularnewline
			\midrule
Lu and Han~\cite{lu2021environment} & 2021 & {Environment enable \ac{DVC}} & ActivityNet Captions& 6.35 & -- & -- & 0.84  \tabularnewline
\midrule
\multirow{2}{2cm}{Zhu et al.~\cite{zhu2022end}} & 2022 & End2end Sequence & Youcook II  & 3.49  & 0.25 & 7.00 & 2.96\tabularnewline
\cline{4-8} 
 &  &  & VITT & 8.10  & 0.25 & 9.26 & 1.29\tabularnewline
 \bottomrule
Wang et al.~\cite{wang2022semantic} & 2022 & Semantic aware pre-training  & ActivityNet Captions& 11.70 & 52.16 & -- & 3.10 \tabularnewline \bottomrule 
\end{tabular}
\begin{tablenotes}
\item[1] Contextual information past \& future events;
\item[2] Lexical fully convolutional neural networks;
\item[3] Dense Video Captioning with ground truth proposals;
\item[4] Dense Video Captioning with learnt proposals;
\item[5] Bidirectional Single Stream Temporal Action Proposals;
\item[6] Graph-based partition-and-summarization;
\item[7] Temporal-linguistic non-maximum suppression;
\item[8] Applying Common-Sense Reasoning to Multi-Modal Dense Video Captioning and Video Question Answering;
\item[9] Weakly supervised Event Caption Generation;
\item[10] Weakly supervised dense event captioning task;
\item[11] End-to-end dense video captioning with parallel decoding;
\item[12] Accelerated Masked Transformer(L=model depth, d=hidden dimensionality);
\end{tablenotes}
\end{threeparttable}
\end{table*}
\subsection{Paragraph Video Captioning}
\label{Sec:PVC}
Generating longer descriptions (but not dense captions) has been explored using paragraph captioning. Yu et al.~\cite{yu2016video} proposed a paragraph \ac{VC} by stacking sentence generator and paragraph generator. In this method, an \ac{RNN} is used for modeling pattern changes over time, a multi-modal layer is used to fuse features from different modalities and an attention model based on spatio-temporal features is used to obtain visual elements. Sah et al.~\cite{ sah2017semantic} evaluated textual summaries of video using recurrent networks, in which the key frames are segregated in the video based on image quality. Then, these key frames are transformed to text annotations based on sequential encoder and decoder design. In another study, Park et al.~\cite{park2019adversarial} developed the paragraph \ac{VC} by employing the ground-truth event segments to produce the coherent paragraphs. A coherent paragraph generation is proposed by Lei et al.~\cite{lei2020mart} using memory augmented transformer. The incorporation of memory module in the transformer architecture produces a highly summarized memory state from the video segments and from the sentence history. Moreover, Song et al.~\cite{song2021towards} discarded the event detection stage and generate paragraphs for the untrimmed videos. The temporal attention mechanism is augmented with the dynamic video memories to improve the sentence coherence. In language perspective, a diversity driven training strategy is employed. Liu and Wan~\cite{liu2021video} produced sentence level captions from the video clips and later summarize these captions to produce the final paragraph caption. This scheme does not rely on ground-truth event segments. In another research, Yamazaki et al.~\cite{yamazaki2022vlcap} proposed a coherant paragraph captioning by minimizing the contrastive learning Visual-Linguistic (VL) loss. Furthermore, Li et al.~\cite{li2022taking} constructed emotion and logic driven multilingual dataset EMPVC for the paragraph captioning to establish the logical associations between sentences and discover more accurate emotions related to video contents as studies in \cite{zhang2021real}. Grounding of language models with features associated with emotion (e.g., facial expressions) has been used to successfully produce more emotionally appropriate language associated with visual content~\cite{huber2018emotional}.
\subsection{Question and Answering}
\label{Sec:vqa}
The Open-domain Question Answering~\cite{ karpukhin2020dense,guu2020realm,lewis2020retrieval} requires a deep learning model to answer any questions employing large-scale documents. Open means not providing the model with documents comprising the right answers directly but requires the model to retrieve documents related to the question from the massive corpus and then generate the correct answer based on them. Therefore, a novel mechanism is required to handle cross-modal interactions in open-domain and can be readily deployed in \ac{VC} tasks. The question-answering task employs unsupervised pre-training models such as BERT and RoBerta~\cite{kim2021bert}, in which the knowledge about the world is gained in an abstract way in the model weights — making it tough to conclude what expertise has been stored and where it is saved in the model. The network's size also has an impact on the amount of storage available and, consequently, the model's accuracy. The conventional method is to train ever-larger networks, which can be extremely slow or costlier, in order to gather more global knowledge. REALM~\cite{guu2020realm} adds a knowledge retriever to the language representation model that first gets a different passage of text from a separate document library, like the Wikipedia text corpus. \par
The REALM retrieval system catered for filling the missing words in a given sentence. REALM employs a reward-based retrieval system that makes good predictions for the missing words are rewarded else it should be discouraged. Just by using question-answer pairings and no information retrieval mechanism, ORQA~\cite{lee2019latent} is the first open-domain question-answering method in which the reader and retriever are simultaneously learned from beginning to end. In this work, the latent variable considered is the evidence retrieval from all of Wikipedia. It is not feasible to train from the scratch, the authors train the retriever using an \ac{ICT}. In \ac{ICT}, a sentence is considered as a pseudo question, while its context is taken as pseudo evidence. The \ac{ICT} needs an option for the corresponding pseudo-evidence among the candidates in a batch.


\begin{table*}
\caption{Comparison of different studies on \ac{VC} (sorted by publication year, and "-" indicates the item is not reported by the corresponding study).} \label{BCSDSSSCOMM}
\centering
\begin{adjustbox}{width=0.90\textwidth}
\begin{tabular}{llllllllllll}
\toprule 
\multirow{2}{*}{\textbf{Study}}      & \multirow{2}{*}{\textbf{Year}} & \multirow{2}{*}{\textbf{Task}} & \multirow{2}{*}{\textbf{Method}} & \multirow{2}{*}{\textbf{Dataset}} & \multicolumn{7}{c}{\textbf{Performance}} \\ \cmidrule{6-12}
          &    &   &  &  & \textbf{METEOR}  & \textbf{CIDEr} & \textbf{ROUGE}  & \textbf{B@1} & \textbf{B@2} & \textbf{B@3} & \textbf{B@4}\\ \midrule
\multirow{2}{*}{Yang et al.~\cite{yang2017catching}} & \multirow{2}{*}{2017} & \multirow{2}{*}{\ac{VC}} & \multirow{2}{*}{DMRM +DA with SS\footnote{}} & \multirow{1}{*}{MSVD} & 33.6 & 74.8 & -- & -- & -- & -- & 51.1\tabularnewline
\cmidrule{5-12} 
 &  &  &  & \multirow{1}{*}{M-VAD} & 6.9 & -- & -- & -- & -- & -- & --\tabularnewline
 \midrule
 Tang et al.~\cite{tang2017richer}  &  2017 & \ac{VC}  & ResNet152-F2F & MSR-VTT   &29.0 & 48.9  & 61.3  & --  &   --  & --  & 41.4  \\
 \midrule 

\multirow{2}{*}{Song et al~\cite{song2017hierarchical}} & \multirow{2}{*}{2017} & \multirow{2}{*}{\ac{VC}} & \multirow{2}{*}{hLSTMat} & \multirow{1}{*}{MSVD} &   33.6&  73.8 & --  &82.9 & 72.2 & 63.0 &53.0 \tabularnewline
\cmidrule{5-12} 
 &  &  &  & \multirow{1}{*}{MSR-VTT} &26.3  & -- &--   &-- & -- & -- & 38.3 \tabularnewline
   
 \hline
\multirow{4}{*}{Wang et al.~\cite{wang2018reconstruction}} & \multirow{4}{*}{2018} & \multirow{4}{*}{\ac{VC}} & RecNetglobal & \multirow{2}{*}{MSVD} & 34 & 79.7 & 69.4 & -- & -- & -- & 51.1\tabularnewline
\cline{4-4} \cline{6-12} 
 &  &  & RecNetlocal &  & 34.1 & 80.3 & 69.8 & -- & -- & -- & 52.3\tabularnewline
\cline{4-12} 
 &  &  & RecNetglobal & \multirow{2}{*}{MSR-VTT} & 26.2 & 41.7 & 59.1 & -- & -- & -- & 38.3\tabularnewline
\cline{4-4} \cline{6-12} 
 &  &  & RecNetlocal &  & 26.6 & 42.7 & 59.3 & -- & -- & -- & 39.1\tabularnewline 
   \hline 
\multirow{2}{*}{Zhao et al.~\cite{zhao2018video}} & \multirow{2}{*}{2018} & \multirow{2}{*}{\ac{VC}} & \multirow{2}{*}{Tube
features} & \multirow{1}{*}{MSVD} & 0.326 & 0.522 & 0.693 & 0.776 & 0.671 & 0.554 & 0.438\tabularnewline
\cline{5-12} 
 &  &  &  & \multirow{1}{*}{Charades} & 0.190 & 0.180& -- & 0.507 & 0.313 & 0.197 & 0.133\tabularnewline
 \hline 
Daskalakis et al.~\cite{daskalakis2018learning}  &  2018 & \ac{VC}  & Triplestream123 & MSVD & 0.3380 & 0.6328 & 0.6962 & 0.7811 & 0.664 & 0.5593 & 0.4502\tabularnewline
\hline 
 \multirow{2}{*}{Lee and Kim\cite {lee2018multimodal}} & \multirow{2}{*}{2018} & \multirow{2}{*}{\ac{VC}} & \multirow{2}{*}{SeFLA\footnote{}} & \multirow{1}{*}{MSVD} & -- &  94.3 &  -- & 84.8 & 70.8 &  70.8&  50.0\tabularnewline
\cline{5-12} 
 &  &  &  & \multirow{1}{*}{MSR-VTT} & -- & -- & -- & -- & -- &--  & 41.8 \tabularnewline
 \hline 
Mun et al.~\cite{mun2019streamlined}   &  2019 & \ac{DVC}  & SDVC\footnote{} & ActivityNet Captions   & 8.82 & 30.68  &  --  &  17.92  &  7.99  &  2.94  &  0.93 \\
   \hline 
\multirow{2}{*}{Tang et al.~\cite{tang2019rich}} & \multirow{2}{*}{2019} & \multirow{2}{*}{\ac{VC}} & \multirow{2}{*}{ResNet152-F2F} & \multirow{1}{*}{MSVD} &   35.7&  84.3 & --  &82.8 & 71.7 & 62.4 & 52.4\tabularnewline
\cline{5-12} 
 &  &  &  & \multirow{1}{*}{MSR-VTT} & 29.0 & 48.9& 61.3  &-- & -- & -- & 41.4 \tabularnewline
\hline 
\multirow{2}{*}{Chen and Jiang~\cite{chen2019motion}} & \multirow{2}{*}{2019} & \multirow{2}{*}{\ac{VC}} & \multirow{2}{*}{MGSA\footnote{}} & \multirow{1}{*}{MSVD} &   35.0 &  86.7  & --  &  & &  & 53.4 \tabularnewline
\cmidrule{5-12} 
 &  &  &  & \multirow{1}{*}{MSR-VTT} & 28.6 & 50.1 &  -- &-- & -- & -- &  45.4 \tabularnewline
 \midrule
\multirow{2}{*}{Hu et al.~\cite{hu2019hierarchical}} & \multirow{2}{*}{2019} & \multirow{2}{*}{\ac{VC}} & \multirow{2}{*}{HTM\footnote{}} & \multirow{1}{*}{MSVD} & 35.2   & 91.3  &72.5  & 86.8 & 75.0 & 65.1 & 54.7 \tabularnewline
\cmidrule{5-12} 
 &  &  &  & \multirow{1}{*}{Charades} & 18.4 & 23.7 &  -- &-- & -- & -- & 14.5 \tabularnewline
 \midrule 
 \multirow{2}{*}{
 Huang et al.~\cite{huang2020multimodal}} & \multirow{2}{*}{2020} & \multirow{2}{*}{\ac{DVC}}   & \multirow{2}{*}{MASSvid\footnote{}-BiD}   & Youcook II & 18.32 & 1.23  &39.03    &  --  &   -- &   -- &  12.04  \\ \cmidrule{5-5}
  &   &  & & \ac{ViTT}  &   10.76 &  0.80 &  31.49  & 22.45   &    --  & --  &  --\\  \midrule
 \multirow{2}{*}{Akbari et al.~\cite{akbari2020neuro}} & \multirow{2}{*}{2020} & \multirow{2}{*}{\ac{VC}} & \multirow{2}{*}{R3\footnote{}-Transformer} & \multirow{1}{*}{Youcook II} & 10.19 & 50.22 & 28.17 & 25.74 & 13.67 & 7.61 & 4.50\tabularnewline
\cmidrule{5-12} 
 &  &  &  & \multirow{1}{*}{Activity Net} & 9.78 & 29.68 & 20.42 & 23.53 & 11.65 & 6.54 & 4.01\tabularnewline
 \midrule 
\multirow{2}{*}{Jin et al.~\cite{jin2020sbat}} & \multirow{2}{*}{2020} & \multirow{2}{*}{\ac{VC}} & \multirow{2}{*}{SBAT\footnote{}} & \multirow{1}{*}{MSVD} & 35.3 & 89.5 & 72.3 & -- & -- & -- & 53.1\tabularnewline
\cmidrule{5-12} 
 &  &  &  & \multirow{1}{*}{MSR-VTT} & 28.9 & 51.6& 61.5 &-- & -- & -- & 42.9\tabularnewline
 \midrule
Fang et al.~\cite{chen2019motion}  &  2020 & \ac{VC}  & Video CMS Transformer & \ac{V2C}   &28.5  &  62.0 &  54.6  &  60.8  &   48.4 &  39.1  & 34.1  \\  \midrule

\multirow{2}{*}{Shi et al.~\cite{shi2020video}} & \multirow{2}{*}{2020} & \multirow{2}{*}{\ac{VC}}  & \multirow{2}{*}{N/A}   &MSVD &  33.0   & 71.0 & -- &  --  & --    &  --  & 51.7  \\  \cmidrule{5-12} 
  &   &  & & MSR-VTT  &26.7   & 45.3  & -- & -- &    --& -- & 41.8 \\  \midrule
  Shi et al.~\cite{shi2020learning}&2020   & VPC\footnote{}   &  N/A&  Youcook II  & 18.31  & 1.12   &  41.56  &    &  &  16.04  &  10.42 \\ 
 \midrule
Iashin and Rahtu~\cite{iashin2020multi}&2020   & \ac{DVC}  &  MDVC, no missings\footnote{} &  ActivityNet Captions  & 11.72  &  -- &   -- &   -- &  --  &  5.83  &  2.86 \\  \midrule
Lin et al.~\cite{lin2020semi}&2020   & \ac{DVC}  &  SC-SSL&   VATEX \& MSR-VTT  &  0.364  & 0.888 &    0.725 &   -- &  --  &  5.83  &  0.572 \\  \midrule
 \multirow{2}{*}{Xiao et al.~\cite{xiao2020exploring}} & \multirow{2}{*}{2020} & \multirow{2}{*}{\ac{VC}}  & \multirow{2}{*}{N/A}   &MSVD & 35.1 & 82.9 & --  & -- & --& -- & 51.3 \\ \cmidrule{5-12} 
  &   &  & & MSR-VTT  & 35.1 & 82.9 & -- & 81.7 & 70.5 & 61.0 & 51.3\\   \midrule
\multirow{2}{*}{Wu et al.~\cite{wu2020hierarchical}} & \multirow{2}{*}{2020} & \multirow{2}{*}{\ac{VC}}  & \multirow{2}{*}{MemNet}   &MSVD & 34.85 &84.27 & --  & -- & --& -- & 51.62 \\ \cmidrule{5-12} 
  &   &  & & MSR-VTT  & 27.1 & 41.9 & -- & --& -- & -- & 38.5\\ \midrule
  
  Shi et al.~\cite{shi2020learning}&2020   & VPC  &  N/A&  Youcook II  & 18.31  & 1.12   &  41.56  &    &  &  16.04  &  10.42 \\ 
 \midrule
Suin and Rajagopalan~\cite{suin2020efficient}&2020   & \ac{DVC}  & N/A &  ActivityNet Captions  & 6.23  &  13.78 &   -- &   -- &  --  &  1.35  & 2.88 \\  \midrule
\multirow{2}{*}{Hou et al.~\cite{hou2020video}} & \multirow{2}{*}{2020} & \multirow{2}{*}{\ac{VC}} & \multirow{2}{*}{Progressive visual reasoning} & \multirow{1}{*}{MSVD} & 34.7 & 80.1 & 71.3 & -- & -- & -- & 47.9\tabularnewline
\cmidrule{5-12} 
 &  &  &  & \multirow{1}{*}{MSR-VTT} & 27.9 & 45.3 & 60.1 & -- & -- & -- & 40.4\tabularnewline
 \midrule
  Hou et al.~\cite{hou2020joint}&2020   & \ac{VC}  &  \makecell{Joint
Commonsense and \\ Relation Reasoning} &  MSVD & 36.8 & 96.8 & -- & -- & -- & -- & 57.0 \\  \midrule
  
Boran et al.~\cite{boran2021leveraging} & 2021  & \ac{DVC}  & N/A &  ActivityNet Captions  &14.46  & 15.54  &     --&   --  &   --  &   --  &  8.88 \\  \midrule
Aafaq et al.~\cite{aafaq2021cross}  & 2021 & \ac{DVC}  &  SC-Net \footnote{}&   ActivityNet Captions  & 10.93 & 14.68  &  22.32  &  21.89  &  10.31  &  5.21  &  2.74 \\  \midrule
\multirow{2}{*}{Wang et al.~\cite{wang2021cross} } & \multirow{2}{*}{2021} & \multirow{2}{*}{\ac{VC}}   & \multirow{2}{*}{CMG\footnote{}}   &MSVD &  38.8  & 107.3 & 76.2 &  --  & --    &  --  & 59.5 \\  \cmidrule{5-12} 
  &   &  & & MSR-VTT  &  29.6  &  53.0 &62.9   & 83.5  &  70.7   & 57.4& 44.9\\ \midrule 
  
\multirow{3}{*}{Chen and Jiang~\cite{chen2021motion}} & \multirow{3}{*}{2021} & \multirow{3}{*}{\ac{VC}}   & \multirow{3}{*}{RRA\footnote{}+ MGCMP\footnote{}+ ATGD\footnote{}}   &MSVD & 36.9   & 98.5  & 74.5 &  --  & --    &  --  &55.8  \\ \cmidrule{5-12} 
  &   &  & & MSR-VTT  &  28.9  & 51.4 &  62.1 & --   &   --   & --& 41.7  \\ \cmidrule{5-12}
   &   &  & & VATEX  & 23.5   & 57.6  & 50.3  & --   &   --   & --&  34.2 \\  \midrule
   \multirow{2}{*}{Sun et al.~\cite{sun2021visual}} & \multirow{2}{*}{2021} & \multirow{2}{*}{\ac{VC}}  & \multirow{2}{*}{VADD\footnote{}}   &MSVD & 34.8  &  91.5 & 72.1 &  --  & --    &  --  & 51.5 \\ \cmidrule{5-12}
  &   &  & & MSR-VTT  &  28.2  & 49.7 &  61.7  & --   &   -- & -- &42.4   \\  \midrule
   Estevam et al.~\cite{estevam2021dense}& 2021  &  \ac{DVC} & N/A&   ActivityNet Captions  & 8.65 &  --  &  --   &  --   & --    &  2.55  & 4.57 \\ \midrule
  Liu et al.~\cite{liu2021aligning}   & 2021 &  \ac{VC} & UVC-VI\footnote{3} & MSR-VTT &   27.8 &  45.5  &  59.5  & --    & --   &  --   &   38.9\\ \midrule
 \multirow{2}{*}{Zhu et al~\cite{zhu2021video}} & \multirow{2}{*}{2021} & \multirow{2}{*}{\ac{VC}}   & \multirow{2}{*}{VCCV\footnote{}}   &MSR-VTT &  29.3  &  48.7 & --  &  --  & --    &  --  & 42.8 \\ \cmidrule{5-12}
  &   &  & & Charades  &   18.0 &  21.0 &    --& --   &   --   & --&  16.5 \\ \midrule
Ji and Wang~\cite{ji2021multi}  &  2021 & \ac{VC}  &  MIMLDL\footnote{} & MSR-VTT   & 27.1&  -- &  59.5  & --  &   --  & --  & 27.1 \\
   \midrule
  Zhang et al.~\cite{zhang2021guidance}  &  2021 & \ac{VC}  & GMNet\footnote{} & MSVD   & 33.5 & 83.1 & 70.7 & -- & -- & -- & 52.1 \\
   \hline
    Bin et al.~\cite{bin2021multi}  &  2021 & \ac{VC}  & PAC\footnote{} & VidOR-MPVC\footnote{}   & 33.5 & 83.1 & 70.7 & -- & -- & -- & 52.1 \\
   \midrule
  \multirow{2}{*}{Ji et al.~\cite{ji2022multimodal}} & \multirow{2}{*}{2022} & \multirow{2}{*}{VPC}   & \multirow{2}{*}{MGNN\footnote{}}   &Youcook II &  21.77  &  136.59 & 42.40 &  --  & --    &  --  & 11.74 \\ \cmidrule{5-12}
  &   &  & & ActivityNet Captions  &   10.80 &  -- &    --& --   &   --   & --&  2.08 \\ \cmidrule{1-12}
 \multirow{2}{*}{Ji et al.~\cite{ji2022attention}} & \multirow{2}{*}{2022} & \multirow{2}{*}{\ac{VC}}   & \multirow{2}{*}{ADL\footnote{}}   &MSVD & 35.7   &81.6  & 70.4 &  --  & --    &  --  & 54.1 \\ \cmidrule{5-12}
  &   &  & & MSR-VTT  &   26.6 &  44.0 &   60.2& --   &   --   & --&  40.2\\ \cmidrule{1-12}
  
\multirow{4}{*}{Seo et al.~\cite{seo2022end}} & \multirow{4}{*}{2022} & \multirow{4}{*}{  M\ac{VC}\footnote{}}   & \multirow{4}{*}{MV-GPT\footnote{}}   & Youcook II & 27.09  &  2.21 & 49.38 &  --  & --    &  --  &  21.88\\ \cmidrule{5-12}
  &   &  & & \ac{ViTT}  &  26.75   &1.04  &  34.76 & 37.89   &   --   & --& -- \\ \cmidrule{5-12}
  &   &  & & MSR-VTT  &  38.66   &0.60  &  64.00 & --   &   --   & --& 48.92 \\ \cmidrule{5-12}
   &   &  & &  ActivityNet Captions  &  12.31   & -- &  -- & --   &   --   & --&   6.84 \\ \cmidrule{1-12}
\end{tabular}
\end{adjustbox}

\begin{tablenotes}
\tiny{
\item[1] Dual Memory Recurrent Model with Semantic Supervision and Decoder Attention; \item[2] SEmantic Feature Learning and Attention-Based Caption
Generation; \item[3] Streamlined \ac{DVC}; \item[4] Motion Guided Spatial Attention; 
\item[5] Hierarchical Temporal Model; 
\item[6] MASS pre-training with video; \item[7] Relative Role Representations; 
\item[8] Sparse Boundary-Aware Transformer; 
\item[9] Video Procedural Captioning;
\item[10] Multi-modal \ac{DVC} module; 
\item[11] Semantic Contextualization Network; \item[12] Cross-Modal Graph; \item[13] Recurrent Region Attention; \item[14] Motion Guided Cross-Frame Message Passing; \item[15] Adjusted Temporal Graph Decoder; \item[16] Visual-aware Attention Dual-stream Decoder; \item[17] Unpaired \ac{VC} with Visual Injection; \item[18] \ac{VC} in Compressed Video; \item[19] Multi-instance multi-label dual learning;
\item[20] Guidance Module Net; \item[21] Perspective-Aware Captioner; \item[22] Multi-Perspective \ac{VC}; \item[23] Multimodal Graph Neural Network; \item[24] Attention-based Dual Learning ; \item[25] Multimodal \ac{VC}; \item[26] Multimodal Video Generative Pre-training; 
} 
\end{tablenotes}

\end{table*}

\subsection{Event Captioning from Video}
\label{Sec:EVNECAP}
Dense event captioning intends to detect and detail all events of interest comprised in a video. This captioning technique tries to improve the inter-task association between event detection and event captioning. Earlier techniques employ dense temporal annotations for event detection which is a time-consuming process. In order to avoid this problem, Duan et al.~\cite{duan2018weakly} introduced a \ac{WS-DEC} which does not depend upon temporal segmentation while training the model. In \ac{WS-DEC}~\cite{duan2018weakly,rahman2019watch}, the localization and captioning modules can only receive information from each other. The \ac{WS-DEC}~\cite{duan2018weakly} method utilized an iterative technique, where the event captioner as well as sentence localizer in turn feeds results to one another. Its fundamental premise is that the output of the sentence localizer will converge to a location that is ideal for the event captioner by optimizing the reconstruction losses.\par
Rahman et al.~\cite{rahman2019watch} focused on the issue of \ac{WS-DEC} in videos and demonstrate that, when paired with video, audio may provide performance that is approximately on par with that of a state-of-the-art visual model. Moreover, Chen et al.~\cite{chen2020learning} developed a multimodal interaction process that utilizes a Channel-Gated Modality Interaction method to calculate pairwise modality interactions, which better accomplishes cross-model information in videos. Li et al.~\cite{li2020grounding}, on the other hand, developed a framework for \ac{MED}~ and~\ac{MEC}. Both \ac{MED} and \ac{MEC} may use the grounded concepts to detect the zero-shot video event and generate multimedia event captioning. The semantic representation of events is enhanced by combining \ac{TF-IDF} with semantic word vector representation of keywords from the events captions. Then, they formulated a Statistical Mean-shift outlier model to remove outliers for generating robust grounded visual features.\par
As a part of submission to ActivityNet challenge 2020, Wang et al.~\cite{wang2020dense} proposed a two plug-and-play modules, \ac{TSRM} and \ac{CMG}. By utilizing the relationships between events in the perspective of both temporal structure and semantic meaning, \ac{TSRM} enhances the event-level description. \ac{CMG} is developed to successfully combine linguistic and visual information in hierarchical \ac{RNN}. Moreover, Zhang et al.~\cite{ zhang2022unifying} also unified event detection and event detection tasks by using the transformer architecture. The authors suggested a unique pre-training task termed Masked Event feature modeling to learn the video-event representation for event detection and used two pre-training tasks, Masked Language Modeling and Masked Video Feature Regression, to learn the video-text representation for event captioning. \par
As these two sub-tasks share the same model architecture and parameters, the authors trained the unified model with the three pre-training tasks to strengthen the cooperation among the two sub-tasks. In another study, Ji et al.~\cite{ji2021hierarchical} proposed a novel hierarchical context-aware model for dense video event captioning to exploit both the local and global context concurrently. The authors applied different mechanisms such as a flat attention module between the source and local context; a cross-attention module for the source to select the global context.\par

\subsection{Other Relevant Studies}
\label{Sec:FURTHEST}
As discussed earlier, usually two different methods are used for encoding and decoding the features of video frames. However, Nabati and Behrad~\cite{nabati2020multi} proposed a sequence-learning model for multi-sentence \ac{VC}. Specifically, they used two \ac{LSTM} modules for encoding and decoding the features of video frames into the sentence words. The first \ac{LSTM} encodes the visual descriptor of video frames and the second \ac{LSTM} determines the output sentence by sequentially generating word probabilities. In addition, they proposed a beam search algorithm that constructs a dictionary of objects using YOLOv3~\cite{redmon2018yolov3} object detector to describe objects in multi-sense videos. Moreover, a multi-stage \ac{RL} algorithm is used to remove incorrect sentences.
\ac{EMVC}~\cite{chang2022event} consists of three modules including multi-modal feature extractor, temporal event proposal and event captioning. \ac{EMVC}, first, obtains various features from audio and video by applying a multi-head attention module and stores the obtained information using a hierarchical RNN. Then, during the caption generation, the effects of visual, audio, and linguistic features are regulated by applying a balance gating.
Instead of collecting features on a frame-by-frame basis, \ac{SGN}~\cite{ryu2021semantic} is based on the semantic grouping of people, objects, and actions. It consists of four modules of the visual encoder, phase encoder, semantic grouping filter, and decoder. The visual encoder extracts features from the frames of the video. The phase encoder produces phrases to the immediately generated captions. The semantic grouping filter removes the correlated phrases and the decoder produces the meaningful sentences.\par
Chan et al.~\cite{chan2020active} conducted an empirical of several active learning strategies using two models, i.e., transformer-based and \ac{LSTM}-based models, and showed that traditional uncertainty sampling techniques are not able to significantly outperform random sampling. In addition, they proposed a cluster-based active learning technique, known as cluster-regularized ensemble divergence active learning, for \ac{VC} that is able to increase the diversity of the samples and improve the performance. In~\cite{baraldi2017hierarchical}, a boundary aware \ac{LSTM} cell is proposed to discover discontinuity points between frames and segments, and enable the encoding layer to enhance the temporal connectivity. To achieve this, it reinitializes the hidden state and memory cell after estimating an action change. In another research, Sun et al.~\cite{sun2022cross} used multiple languages for \ac{VC}. After extracting features from multiple languages, the high-level sense semantics are learned for each video. In addition, entity words are learned by applying a leap sampling technique to better represent the video content.\par
The \ac{VC} model in~\cite{tang2022visual} consists of three branches including language, visual semantic, and optimization branches. Firstly, it integrates language and visual semantic branches with a multi-modal feature-based module. Then, a multi-objective training strategy is used to optimize the model. The outcome of these three branches are fused via a weighted average to predict the word for each frame. The proposed visual encoding technique in~\cite{aafaq2019spatio} extracts features enriched with spatio-temporal dynamics of the scene and semantic attributes of the videos to generate rich captions using a gated recurrent unit. Short Fourier transforms are applied hierarchically to the \ac{CNN} features of the whole video to process the activation of \ac{CNN} features. \par
Most of the \ac{VC} techniques operate directly on video and they neglect completely additional contextual information. In order to effectively capture this information, Man et al.~\cite{man2022scenario} proposed a \ac{VC} based on a scenario-aware recurrent transformer. The dependencies in hybrid models of RNN-CNN is less studied in the literature. An attempt is made in~\cite{huang2022accelerating} to the GPU-capacity-guided pipelining and EdgeTPU-capacity for maximizing GRU and SRAM utilization. The proposed \ac{VC} method by Hosseinzadeh and Wang~\cite{hosseinzadeh2021video}, first, forecasts the features of future frames in the semantic space of convolutional features, and then merges the contextual information into those features, and applies it to a captioning module that largely improved the efficiency of the \ac{VC}. Moreover, Lebron et al. ~\cite{lebron2022bertha} provided a new strategy for training a new \ac{VC} evaluation score. This method learns to correlate system-generated captions to human references based on human evaluations. The majority of metrics attempt to compare the system-generated captions to a single or collection of human-annotated captions. This work proposes a new way based on a deep learning model for evaluating these systems. The model relies on BERT, a language model that has proven to be effective in a variety of NLP tasks. The goal of the model is to learn and make a human-like assessment. This is accomplished by analyzing a dataset of human judgments of captions generated by the machine. The dataset is made up of human evaluations of captions generated by the algorithm during different years of the TRECVid video-to-text task. To lessen the uncertainty in the training and test data, numerous human judgments per caption are collected as the human decision is often erroneous. MV-GPT~\cite{seo2022end} is a pre-training framework that learns from unlabeled videos. It uses future utterances as an auxiliary text to solve the lack of captions in unlabeled videos. MV-GPT trains both multi-modal video encoder and sentence decoder jointly, and uses raw pixels and transcribed speech to directly generate a caption. Iashin et al.~\cite{iashin2020multi} obtained a temporally aligned textual description of the speech by applying automatic speech recognition (ASR) system. Besides, they formulated the captioning task as a machine translation task and converted the multi-modal input data into textual descriptions via a transformer technique. 

\section{Applications}
\label{Sec:App}


\ac{VC} has applications in video tagging and retrieval, video understanding, video subtitling, video title generation, action recognition, event detection, content-based video retrieval, accessibility, video indexing, human-robot interaction, video summarization and video recommendation. In this section, we review these applications and methods that have been proposed to address them.

\subsection{Video Tagging, Indexing and Retrieval}
\label{Sec:sec:sec:VISE}
With vast media collections available online there has developed a need for automatic understanding along with efficient retrieval of these data. Commercial search engines like \emph{Google}, \emph{Microsoft Bing}, and \emph{Yahoo!}, offer video search based on indexing textual metadata related to videos. However, the search performance can be unsatisfactory using the limited amount of manually annotated textual metadata. Manual annotations can be inconsistent and incomplete. To reduce human labeling costs, and create more consistent as well as complete textual data, semantic-based indexing has been developed. The deep features may serve as signatures for semantic details of video helpful in many search and retrieval tasks~\cite{podlesnaya2016deep}. The graph-based storage structure used in this model for video indexing allows to retrieval the content aptly with complicated spatial and temporal search queries.

Tags are descriptive keywords that are added to videos. Video tagging is used to index and catalog videos based on content making it easier to search. While different from \ac{VC}, video tagging is nevertheless relevant and has been used in \ac{VC} systems. Several related innovations have been introduced to build video tagging algorithms, including introducing POS tags to produce syntactically correct text~\cite{perez2021improving}, semantic tags to bridge the gap between the vision and text~\cite{huang2021semantic}, and integrating a video scene ontology with a \ac{CNN}~\cite{ilyas2019deep}.  
 
The strategy used in video retrieval is, given a text query and a pool of candidate videos, to select the video which corresponds to the text query. Garg et al.~\cite{garg2021video} proposed combining tagging with a neural network to retrieve relevant videos~\cite{rohrbach2017movie}.
The \ac{LSMDC} dataset challenge~\cite{rohrbach2017movie} comprises four video-to-language tasks. The movie retrieval task involves ranking 1,000 movie clips for a given natural language query. Dong et al.~\cite{dong2019dual} proposed zero example video retrieval in which an end user searches for unlabeled videos. Using a Tree-augmented Cross-modal Encoding technique, Yang et al.~\cite{yang2020tree} presented video retrieval via more complex questions by mutually learning the language structure of queries along with the temporal representation of videos. Cao et al.~\cite{cao2022visual} proposed a two-stage \ac{VC} approach, in the first stage, video retrieval was used to find sentences related to a given video from the text corpus, and in the second stage, the retrieved captions are used to guide caption generation.\par

\ac{CBVR} systems serve a critical role in improving human–computer interaction. Several modern medical applications, for example, record the health state of patients in real-time and preserve the video for further examination. While tracking things, surveillance agencies mostly depend on video footage. All of the recorded videos are preserved in a database for later investigation, and indeed the database is also relatively large. The ultimate focus of video storage is to examine videos for decision-making as well as comparison’s purposes. Video retrieval is somewhat more important for good data analysis since it aids in the retrieval of relevant videos and improves decision-making. The \ac{CBVR} system overcomes the challenges of the text query-based video retrieval systems experience, such as increased computational and time demands. A conventional \ac{CBVR} design comprises of three key phases: query acquisition, video retrieval, and grading. The \ac{CBVR} system evaluates the query using advanced image processing algorithms and extracts the videos related to the query as soon as the query is sent to it. Finally, the videos are organized into categories based on their degree of importance. Moreover, since it is not essential to process all the frames in a video, only the key frame must be identified and obtained. The essential frame's features are extracted and used in the video recovery procedure. This strategy thus saves both memory and time while also optimizing work efficiency.\par
 As a consequence, the current requirement for an efficient \ac{CBVR} system is to retrieve relevant videos in a reasonable amount of time with a higher accuracy rate. Taking this as a challenge, Prathiba and Kumari~\cite {prathiba2021content} proposed a useful \ac{CBVR} system that takes into account both audio and video aspects when delivering appropriate videos to the viewer. The work~\cite{prathiba2021content} is accomplished through the use of two modules: video and audio. Video frame extraction using shot detection, key frame detection, plus feature extraction are three crucial steps in the video module. Audio denoising as well as feature extraction are part of the audio module. The feature database is created by combining the features gathered from both modules. The attributes are clustered using the kernelized fuzzy C-means (KFCM) technique, which speeds up video retrieval even further. The Euclidean distance seen between the query and the set of features is then calculated. The closer the video is to the question, the more relevant it is. Pyramid regional graph representation learning (PRGRL)~\cite{zhao2021pyramid} takes control of both local and global techniques, and also compensates for their shortcomings by proactively measuring the "importance" of a video's sectors in its frames.
 
\subsection{Video Understanding}
\label{Sec:sec:sec:VUNDER}
\ac{VC} can be thought of as a downstream product of video understanding, with some form of understanding being a prerequisite, in many cases implicitly, for captioning. Video understanding aims to recognize and localize different actions or events in a video. In this regard, Diba et al.~\cite{diba2020large} proposed a spatio-temporal ``Holistic Appearance and Temporal Network" (HATNet). The HATNet combines 2D and 3D architectures by joining intermediate representations that capture appearance-based and temporal features. Structured \ac{GNN} models have been used to capture spatio-temporal interactions explicitly under the supervision and implicitly as nodes of spatio-temporal graphs~\cite{arnab2021unified}, as have perturbation-based models~\cite{li2021towards}. Huang et al.~\cite{ huang2021tada} proposed temporally-adaptive convolutions (TAdaConv) arguing that by calibrating the convolution weights for each frame according to their local and global temporal context it combines spatial and temporal modeling abilities. Global-local video features can also be learned contrastively as leveraged by Zeng et al.~\cite{zeng2021contrastive}.



\subsection{Video Subtitling}
\label{Sec:sec:sec:VIDSUBNNY}
Video subtitling is a direct application of \ac{VC} that can be used in video production for translating foreign language programs or for making content more accessible. A vast majority of video content is not manually described by humans, therefore machine-generated subtitles~\cite{brousseau2003automated,alvarez2016automating}, while less accurate, can be very useful. 
Soe et al.~\cite{soe2021evaluating} showed that using AI in a semi-automated workflow could support but not replace a human annotator. Guerreiro et al.~\cite{guerreiro2021towards} proposed a monolingual model that effectively showed consistent performance in 100 different languages. 
\subsection{Video Title Generation}
\label{Sec:sec:sec:VIDTITG}
Title generation can be thought of as a very condensed form of summarization or subtitling. Zeng et al~\cite{zeng2016generation} addressed this by applying two methods for automatic video title generation. The first approach employed a primer with a highlight detector, jointly training a model for title generation and highlight localization. The second approach promoted sentence diversity, such that the generated titles were also distinct and expressive. To achieve sentence diversity, the authors incorporated novel sentence augmentation techniques.
\subsection{Action Recognition}
\label{Sec:sec:sec:ACCTION}
Video action recognition is one of the key research areas used for video understanding. It is used for classifying human action categories in videos~\cite{pareek2021survey}. One of the main problems with video action recognition models is that they are often limited to detecting only a few classes, whereas captioning as an alternative is more expressive. Wang et al.~\cite{wang2021actionclip} proposed a video action recognition task using action clips in a multimodal learning framework with semantic supervision. The main advantage of this model is that it enabled zero-shot action recognition without labeled data or parameter tuning. A bottleneck in video action recognition is that most models can only handle a small number of input frames and are therefore unable to capture complex long-term dependencies. Zhi et al.~\cite{zhi2021mgsampler} proposed a motion-guided frame sampling, considering the cumulative motion distribution to ensure the sampled frames cover all the important segments and have high motion salience. Equivalent motion saliency cannot be captured well with a simple spatio-temporal convolution due to the requirement for a very large network. In another research, Kwon et al.~\cite{kwon2021learning} proposed a method to capture motion dynamics based on spatio-temporal self-similarity (STSS). STSS represents each local region as similar to its neighbors in space and time. It can be deployed in any neural architecture and can be trained end-to-end without requiring supervision.
\subsection{Event Detection}
\label{Sec:sec:sec:EVEDER}
Event or anomaly detection is the task of identifying abnormal events in the video. Given the nature of rare events, it is very time-consuming to manually annotate content and therefore automated algorithms are attractive, whether it be objected detection or person re-identification. Huang et al.~\cite{huang2021abnormal} introduced a temporal contrastive network to address the problem of finding rare events in videos. The algorithm was unsupervised, utilizing deep contrastive self-supervised learning to extract higher-level semantic features and anomaly detection with multiple self-supervised tasks. Pan et al.~\cite{pang2022fall} used the real-world fall (RWF) dataset,in which events are captured with mobile devices or cameras. This dataset recorded fall events, with the goal of helping to identify injuries. In their work, temporal local information was extracted via semi-supervised learning, they introduced a multiple instance (MI) module to extract small time-scale information. The multiple instance learning determines a segment that may possibly contain falls in a video by minimizing the sum of two losses of the multiple instance learning (MIL) loss and cross-entropy loss.
\subsection{Accessibility}
\label{Sec:sec:sec:BLIN}
\ac{VC} can be used as an assistive technology in physical environments, for example helping individuals who are blind travel independently~\cite{giudice2008blind}. In~\cite{lakshan2021blind}, a blind navigation system was formulated that detected thin structured wires as obstacles. This was realized by taking the video stream of a path taken by a blind individual from a single-lens camera. The system consisted of three blocks an image information extraction block, a tracking block, and a mapping block. In the former block, the difference of Gaussians (DoG) was used to derive keylines. Later, these keylines were used to create an edge map for each frame. In the second stage, the previous edge map was fit into the new edge map by employing a warping function. Each keyline in the new edge map was matched against the ones in the previous map in order to separate real obstacle edges from spurious ones. In ~\cite{gunethilake2021blind}, an obstacle identification using an SSD mobile net architecture was proposed, with an audio alert to warn of the distance from an obstacle.
\subsection{Human-Robot Interaction}
\label{Sec:sec:sec:HUMAN}
\ac{HRI}~\cite{kong2018human} is a rapidly developing field of research with applications as broad as search and rescue, mine and bomb detection, scientific exploration, law enforcement, entertainment, and elder care. In many applications, it would be beneficial if a robot could communicate with a human collaborator by providing an explanation of its understanding in natural language. Temporal models have been employed to this end in \ac{NLVD}~\cite{cascianelli2018full}. The egocentric view of robotic vision and exocentric view of general video are fused in \ac{VC} for better~\ac{HRI} interaction~\cite{kang2021video}.

\subsection{Video Description and Summarization}
\label{Sec:sec:sec: VIDDESC}
Video description has to capture all key aspects of the underlying events in a video and specify with a story-like description with multiple sentences. The generated descriptions must be relevant, coherent and concise. Zhu et al.~\cite{zhu2021saying} proposed a video description of videos with occlude information. The authors described the video employing the natural language dialogues between the two agents. The dialogue generation between two agents may be generative, i.e., the agent generates questions along with answers freely, or discriminative, i.e., agents opt for the questions and answers coming out of the candidates. Existing caption datasets for video descriptions are scarce. Hence, Monfort et al.~\cite{monfort2021spoken} proposed \ac{S-MiT}, which comprises 500k spoken captions; each ascribed to a unique short video describing a wide range of assorted events. Moreover, Xiong et al.~\cite{xiong2018move} developed a concise and coherent paragraph description of the video using self-critical sequence training of \ac{RNN} by giving rewards at sentence and paragraph level. \par
Video summarization~\cite{elfeki2022multi} aims to fuse segments of videos that include key visual information~\cite{yang2015unsupervised}. Video summarization techniques rely on low-level features such as color, and motion~\cite{zhang1997integrated} or objects~\cite{zhang1997integrated} and their relationships to select key segments. Meanwhile, others utilize text inputs from user studies to extract the key segments~\cite{song2015tvsum}. Video summarization aims to provide this information by generating the gist of a video, benefiting both the users and companies that provide video streaming and search (with increased user engagement). It should be noted that video summarization is an efficient approach to facilitate both video browsing and searching~\cite{li2021exploring}.

\subsection{Video Recommendation}
\label{Sec:sec:sec:VIDDRECC}
The video recommendation system suggests a video to a user. These systems typically feature two blocks, one is a candidate block that extracts videos as candidates from a large-scale video repository, and the second is a block ranking these candidates to select the best match for the user. DL-based video recommender systems exploit multiple features including user demographics and behaviors, video titles, and video tags for suggesting recommendations. Pu et al.~\cite{pu2020multimodal} formulated a recommender system by integrating semantic features in a multi-modal topic learning algorithm to improve the efficiency of the recommender system. Dong et al.~\cite{dong2018feature} boosted the performance by taking data augmentation at frame level and video level, and later incorporated the features in a multi-modal recommender system. 
Liu et al.~\cite{liu2021concept} proposed concept-aware denoising \ac{GNN}, which is a micro video recommender system, formed based on a tripartite graph to link user nodes with video nodes, and video nodes with associated concept nodes, retrieved from captions and comments of the videos.\par
Video recommendation is a particularly important tool for online video providers like \emph{YouTube} and \emph{Hulu} since it helps viewers discover videos. User interaction data such as clicks, views, comments, or ratings are not available when a video is first published to a service, this is referred to as the cold-start issue. BlerinaLika et al.~\cite{lika2014facing} addressed the cold-start problem by adopting a three-phase approach. The first is the collection of demographic data of new users. The second approach is finding the user's neighbors based on people with a common interest. The third is classifying new users into a group and based on the classification, computing ratings for new items. The final ratings are modified with a weighting scheme, where developers can pay attention to the specific attributes.

\section{Discussion}
\label{Sec:Diss}
In this survey, we have provided a comprehensive review of \ac{VC} methods. We first categorized the \ac{VC} methods into template and sequence-to-sequence methods, and then, divided each category into sub-categories and presented their corresponding representative methods. Then, we described datasets, and performance metrics, and discussed applications. Table~\ref{BCSDSSSCOMM} provides a comprehensive comparison of \ac{VC} methods in the literature. In the following sub-sections, we list the most important research gaps in the domain (see sub-section~\ref{Sec:RG}) and future research directions (see sub-section~\ref{Sec:FR}).



\subsection{Research Gaps}
\label{Sec:RG}
Despite the recent and substantial progress in \ac{VC} research, there are a number of unsolved challenges. Here we discuss these gaps and potential solutions. \ac{VC} methods are still not very effective at capturing sequences of actions, especially in shorter video clips~\cite{pasunuru2017multi} and many also neglect the multi-modal nature of videos~\cite{jin2019low}. 
The proposed encoder-decoder models in~\cite{venugopalan2016improving, rohrbach2015dataset} do not use high-level video concepts, resulting in-poor results. Sequence-sequence models~\cite{aafaq2019video} poorly represent the objects in the video because the encoder is used to generate low-level features and not the objects. In supervised \ac{VC} algorithms, the generated captions are evaluated with the ground truth caption using a loss function calculated on a word-to-word basis. However, changing a single word in a sentence can have a significant impact on its meaning. \par
In \ac{RNN}s, a hidden state is computed based on the previous hidden state such that it cannot be parallelized. Sequential computation introduces a high cost, especially for long sequences. The use of pre-trained networks for extracting object interactions in temporal frameworks suffers from several drawbacks. Firstly, they cannot derive discriminant multi-object features. Second, models only trained for recognizing human actions fail to capture the temporal information in other non-human objects in the scene, e.g., animals, and vehicles. Moreover, different visual features that are fed into captioning modules at the same time often represent different time periods, resulting in confusion during training~\cite{zhu2020ovc}. Third, training a model that captures more subtle fine-grained visual attributes is difficult. Furthermore, object occlusions and unclear boundaries make vision-to-language translation difficult~\cite{li2019visual}. 
Finally, the majority of the proposed techniques for \ac{VC} in the literature mainly focus on English-only video tasks~\cite{gan2022vision}. The lack of video datasets in other languages is one of the reasons for this bias. More attention needs to be paid to faithfully modeling under-represented languages, and evaluating performance in languages other than English.

\subsection{Future Research Directions and Outlooks}
\label{Sec:FR}

\subsubsection{Uncertainty Quantification in Video Captioning}
\label{Sec:sec:UQVC}
Quantifying the uncertainty of machine learning and deep learning models plays a critical role in producing reliable predictions. In general, the sources of uncertainty can be grouped into \textit{aleatoric} (data) and epistemic (model) uncertainties~\cite{abdar2021review,zhou2021survey,psaros2022uncertainty, abdar2022need}. \ac{VC} has been widely used in many sensitive fields such as medical imaging~\cite{abdar2021barf, abdar2022hercules, abdar2023uncertaintyfusenet} and other safety-critical scenarios~\cite{secchi2008quantifying}. 
However, our findings indicate that although quantifying the uncertainty of \ac{VC} methods is important, to the best of our knowledge, there are no methods that fully characterize uncertainty in an interpretable way. We would suggest the research community continues to direct efforts in this area and develops new uncertainty quantification methods.

\subsubsection{Curriculum Learning in Video Captioning}
\label{Sec:sec:CURRMLAA}
\ac{CL} is an effective learning strategy, which is inspired by the human learning ability, that trains a learning algorithm from easier to harder samples of datasets~\cite{wang2021survey,graves2017automated}. The key question in developing a good curriculum is how to measure the difficulty of each example~\cite{kong2021adaptive}. This strategy is able to improve the convergence rate and generalization capability of learning algorithms in tackling various scenarios such as \ac{NMT} and \ac{NLP}~\cite{zhan2021meta}. Our review shows that there exist only a few studies conducted in \ac{CL} for \ac{VC}. Thus, we list this field as potential research for further investigation.

\subsubsection{Continual Learning in Video Captioning}
\label{Sec:sec:ECONNTT}
Continual Learning is referred to a group of machine learning techniques that attempt to continually learn from a stream of data while its distribution shifts over time~\cite{doan2022efficient}. Catastrophic forgetting, e.g., the learning algorithm should able to learn information from new data without forgetting the previously learned information, is the main challenge of this strategy that must be addressed~\cite{mirzadeh2022architecture,mirzadeh2020linear,mirzadeh2020understanding}. This is another interesting field that can be used for \ac{VC}, thus we list it as a research gap that requires further investigations. 

\subsubsection{Explainability in Video Captioning}
\label{Sec:sec:sec:EXPVC}
Many DL models are black boxes due to their complex structures and a large number of parameters. This makes it difficult for users to interpret their predictions. In addition, it is crucial for many real-world problems such as healthcare and safety-critical industrial problems to interpret the model prediction~\cite{ghassemi2021false,miller2019explanation}. In this regard, explainable artificial intelligence (EXI)~\cite{gunning2019xai,vilone2021notions} have been proposed. Our findings in this review paper indicate that interpreting the predictions of the \ac{VC} methods can improve their performance.

\subsubsection{Transformers in Video Captioning}
\label{Sec:sec:transformers}
Transformers~\cite{lin2021survey,yuan2021incorporating} are prominent types of DL models, originally proposed as for sequence-to-sequence tasks in machine translation. Transformers require minimal inductive biases and they are able to process multiple modalities~\cite{khan2021transformers}. In addition, they are robust to domain shifts and perturbations~\cite{naseer2021intriguing}. 
Raghu et al.~\cite{raghu2021vision} showed that transformers extract more uniform features across all the layers. Recently, a few studies have used transformers for \ac{VC}. For example, the proposed transformer model in~\cite{iashin2020better} generalizes the transformer architecture for a bi-modal input (audio and visual). We believe that transformer-based methods still require more investigations in dealing with the \ac{VC} problems.

\subsubsection{Knowledge Distillation in Video Captioning}
\label{Sec:sec:KNOWDISS}
\ac{KD}~\cite{gou2021knowledge,wang2021knowledge} is a type of compression and acceleration that is able to learn a small student model from a large teacher model in a target domain. In other words, \ac{KD} is able to solve the problems caused by the lack of labeled samples via transferring knowledge learned from one model to another. Nonetheless, the key challenge of \ac{KD} is selecting the effective information for transferring~\cite{kobayashi2022extractive}. Our review reveals that only a few studies applied \ac{KD} for \ac{VC}. For example, Pan et al.~\cite{pan2020spatio} used \ac{KD} mechanism to avoid unstable performance caused by the variable number of objects. Hence, we suggest designing \ac{KD}-based frameworks for \ac{VC}.

\subsubsection{Active Learning in Video Captioning}
\label{Sec:sec:sec:KNOWDISS}
\ac{AL} methods are effective in dealing with the lack of labeled samples. They label data samples in an iterative manner until achieving sufficient labeled samples with promising outcomes~\cite{kwak2022trustal,ren2021survey,aljundi2022identifying}. In other words, \ac{AL} identifies the most informative samples that should be sued for annotations. It has been adopted in many real-world applications such as satellite image segmentation~\cite{desai2022active} and medical image analysis~\cite{budd2021survey}. We have found that fewer studies focused on \ac{AL}-based \ac{VC}. For example, Chan et al.~\cite{chan2020active} proposed a cluster-based active learning technique for \ac{VC} that is able to increase the sample diversity and improve the model performance. 
\subsubsection{Text-to-Video in Video Captioning}
\label{Sec:sec:sec:TEX@VIDEO}
There are billions of (alt-text, image) pairs from HTML pages on the Internet, enabling great research interest in \ac{T2I} modeling. However, \ac{T2V} is very complex than \ac{T2I} due to the additional temporal dimension that makes the generated data high dimensional, and also the generated video must be both photo-realistically diverse and should look natural~\cite{balaji2019conditional}. \ac{T2V} is widely used in generating synthetic data for machine learning tasks, domain adaptation, multimedia applications, text-to-video retrieval, and also important research direction to improve the overall effectiveness of interactive video search systems. The two challenges of \ac{T2V} are that one is the generated video must be natural and temporally coherent and the second is the content of the generated video must match with the input text. The invention of \ac{GAN}, \ac{VAE}, \ac{c-GAN} boosted the research in \ac{T2I} and \ac{T2V}. Li et al.~\cite{li2018video} proposed \ac{T2V} based on \ac{CGAN}. The shortcoming of the scheme is that they used 3D transposed convolution layers in the \ac{GAN} which produce only a fixed-length video. Second, the videos generated are poor in resolution and thirdly the text-to-video synthesis has shown poor generalization on large datasets. The above problems are rectified in~\cite{balaji2019conditional} by employing \ac{RNN} for generating variable length video. The poor resolution of the video is solved by employing RESNET. Further, to strengthen the associations between the conditioned text and the generated video Text-Filter conditioning GAN (TFGAN)~is employed. The same problems in~\cite{li2018video} are addressed in another way by~\cite{deng2019irc}. Deng et al.~\cite{deng2019irc} have improved the temporal coherence and the visual quality of the video by employing recurrent cells with 2D transconvolutional layers. The 2D transconvolutional layers put more emphasis on the details of each frame than on 3D and improve the videos with better visual quality. The mutual-information introspection is incorporated in GAN to match the generated video to text semantically. Semantic consistency is measured by introspecting the semantic distance between the generated video and the corresponding text and trying to minimize it to boost semantic consistency. In \ac{T2VR}, the text modality and the video modality need to be represented in a shared common space for \ac{T2V} similarity matching. For video encoding, 2D-RESNET models are pre-trained on ImageNet to generate visual features. The text is encoded by running multiple text encoders independently
with their output concatenated later~\cite{li2019w2vv++} or by stacking the encoders~\cite{li2018renmin}. Loco et al.~\cite{lokoc2020w2vv++} deployed a lightweight W2VV++ model in \ac{T2VR}.
The advantage of this scheme is that 
pipelined architecture of W2VV++ facilitates to the removal of a specific
encoder with ease which made the proposed \ac{T2VR} very interactive in real-time. Chen et al.~\cite{chen2020fine} represented the shared common space for \ac{T2V} by developing a Hierarchical Graph Reasoning model that decomposes videos and texts into hierarchical
semantic levels of events, actions, and entities. Later, they obtained textual embeddings
via attention-based graph reasoning and matches the text with videos at different fine-grained levels. Mun et al.~\cite{mun2020local} proposed \ac{T2V} scheme is effective in capturing relationships of semantic phrases in the text and video segments by modeling local and global contexts. Hu et al.~\cite{hu2022lightweight} point out the drawbacks of \ac{MHA} as it pays more attention to weak features. So they proposed Lightweight Attention Feature Fusion in place of \ac{MHA} for exploiting diverse,
multi-level (off-the-shelf) features for \ac{T2VR}. Dong et al.~\cite{dong2022reading} formulated the shared common space between the text and video in \ac{T2VR} problem inspired by the reading strategy of humans. The visual representation learning of videos is analyzed in two stages of the reading, one is previewing and another is intensive reading. The previewing of the video is to analyze the overview and the intensive reading stage is to extract in-depth information for obtaining the fine-grained features. In another study, Fadime Sener~\cite{sener2022transferring} developed \ac{T2V} by training on large text corpora and video models are trained on scarce parallel data and can be used for zero-shot queries (data that is unseen prior).

\subsubsection{Diverse Texts in Video Captioning}
\label{Sec:sec:sec:DIVVDERTSTEX}
Generating high-quality texts with high diversity
is an important requirement in computer vision. In general, a video may contain complex texts and visual information that is very difficult to represent comprehensively. The existing method employs conventional \ac{VC} schemes to address this problem, which produces a global caption for the entire scene in a frame. The problems in the existing work are first, it is difficult to know which parts of text in a video to be paraphrased. Second, it is very difficult to exploit the complex relationship between diverse texts in the video. Third, the Generation of multiple captions with diverse content. Shi et al.~\cite{ shi2018toward } proposed a novel Auto-captioner method. The authors used the important tokens which are given more attention and take them as anchors. Later, for each identified anchor, its relevant texts are mapped to obtain the \ac{ACG}. Finally, based on different \ac{ACG}s, the authors produced multi-view caption generation to improve the content diversity of generated captions. The existing neural models are not able to model the input data dynamically during the generation to produce diverse texts. It is due to the lack of capturing inter-sentence coherence The above problem is rectified by~\cite{shao2019long} using a 
planning-based hierarchical Variational model
to prevent the inter-sentence incoherence problem. The authors planned a hierarchical generation process, which models the process of human writing. To obtain this, the author segments the text generation into a sequence of dependent sentence generation sub-tasks where each sub-task depends specifically on an individual group and
the previous context. By this scheme, the input data
can be modeled better and thus obtains inter-sentence coherence. Zhang et al.~\cite{ zhang2022divergan } proposed \ac{T2I} using Driver GAN. The single-stage schemes used in \ac{T2I} suffer from the lack of diversity, yielding similar output to diverse texts. To prevent this, the authors introduced a channel-attention module (CAM) and a pixel-attention module (PAM), which give more importance to each word in the given sentence while allowing the network to assign larger weights to the significant channels and pixels semantically aligning with the salient words. The authors achieved text diversity, without harming quality and semantic consistency. Du et al.~\cite{ du2022diverse } proposed an improvement over existing works by introducing context-aware variations into the encoder based on Gaussian process priors to produce diverse texts. This improves inter-sentence coherence which can help to preserve more semantic information
from source texts. During generation, the decoder generates diverse outputs conditioning on sampled different context variables. In other words, by learning a stochastic function on top of one deterministic encoder, the proposed approach offers many versions of random context variables for a decoder to generate diverse texts.
\section{Conclusion}
\label{Sec:Co}
In this work, a detailed survey on deep learning-based Video Captioning (\ac{VC}) methods along with datasets and evaluation metrics is presented. Following the detailed survey, the performance of the \ac{VC} methods using different datasets and evaluation metrics is comprehensively evaluated. This survey covers the wide applicability of \ac{VC} in other fields such as video tagging, content-based image retrieval, video recommender systems, blind navigation, etc. Some research gaps are highlighted in this to serve as a good starting point for researchers who are interested in \ac{VC} and shed some light on this ever-growing field.


\section{Acknowledgment}
We thank Professor Jianfeng Gao at Microsoft Research, Redmond, USA for checking the article and providing valuable
feedback and comments. 
\section*{Acronyms}
\begin{acronym}[1234567]
\setlength{\itemsep}{-\parsep}
  \acro{VC}{Video Captioning} 
 \acro{3G}{International Mobile Telecommunications-2000}
 \acro{4G}{Successor to 3G}
 \acro{R-AConvEN}{Retrieval Augmented Convolutional Encoder Network}
  \acro{R-AConvDN}{Retrieval Augmented Convolutional Decoder Network}
 \acro{SVO}{Subjects Verbs and Objects}
 \acro{NLP}{Natural Language Processing}
 \acro{CNN}{Convolutional Neural Network}
\acro{2D-CNN}{Two dimensional Convolutional Neural Network}
\acro{GNN}{Graph Neural Network}
\acro{CBVR}{Content Based Video Retrieval}
\acro{3D-CNN}{Three dimensional Convolutional Neural Network}
 \acro{RNN}{Recurrent Neural Network}
 \acro{AI}{Artificial Intelligence}
\acro{3D-CNN}{Three Dimensional Convolutional Neural Network}
\acro{MP}{Mean Pooling}
\acro{TE}{Temporal Encoding}
\acro{SEM}{Semantic Attribute Learning}
\acro{LSTM}{Long Short-Term Memory}
\acro{EMVC}{Event-centric Multi-modal fusion approach for dense Video Captioning}
\acro{SGR}{Sketch, Ground, and Refine}
\acro{ViSE}{Visual-Semantic Embedding}
\acro{GRU}{Gated Recurrent Unit}
\acro{Self-Attn}{Self-Attention}
\acro{Soft-Attn}{Soft-Attention}
\acro{Hard-Attn}{Hard-Attention}
\acro{Text-Attn}{Text-Attention}
\acro{Local-Attn}{Local-Attention}
\acro{Global-Attn}{Global-Attention}
\acro{Stacked-Attn}{Stacked-Attention}
\acro{Bimodal-Attn}{Bimodal-Attention}
\acro{Temp-Attn}{Temporal-Attention}
\acro{FFN}{Feed Forward Network}
\acro{MAM}{Multi-level Attention}
\acro{MHA}{Multi-head attention}
\acro{HATT}{Hierarchical attention strategy}
\acro{AAMF}{Attention-based Averaged Multimodal Fusion}
\acro{AMF}{Attention-based Multimodal Fusion}
\acro{HRNAT}{Hierarchical Representation Network
with Auxiliary Tasks}
\acro{AVSSN}{Attentive Visual Semantic Specialized Network}
\acro{DL}{Deep Learning}
\acro{GPaS}{Graph-based Partition-and-Summarization}
\acro{GCN}{Graph Convolutional Network}
\acro{NMT}{Neural Machine Translation}
\acro{GAN}{Generative Adversial Network}
\acro{CGAN}{Conditional Generative Adversial Network}
\acro{DVC}{Dense Video Captioning}
\acro{TGANs-C}{Temporal GANs conditioning on Captions}
\acro{CGAN}{Conditional Generative and Adversial Networks}
\acro{BLEU}{Bilingual Evaluation Understudy }
\acro{METEOR}{Metric for Evaluation of Translation with Explicit Ordering }
\acro{ROUGE-L}{Recall Oriented Understudy of Gisting Evaluation }
\acro{CIDEr}{Consensus based Image Description Evaluation }
\acro{VS}{Vocabulary Size}
\acro{PNS}{Percentage of Novel Sentences}
\acro{DC}{Diversity of Captions}
\acro{DCE}{Diverse Captioning Evaluation}
\acro{V-QA}{Video Question Answering}
\acro{CBIR}{Content Based Image Retrieval}
\acro{MSVD}{Microsoft Research Video Description Corpus }
\acro{MSR-VTT}{ Microsoft Research-Video to Text}
\acro {M-VAD}{Montreal Video Annotation Dataset}
\acro {V2C} {Video-to-Commonsense} 
\acro {ViTT} {Video Timeline Tags}  
\acro {OOC} {Object-Oriented Captions dataset}
\acro{BF}{Brevity Factor}
  \acro{UQ}{Uncertainty Quantification} 
  \acro{CBVR}{Content Based Video Retrieval}
  \acro{LCS}{Longest Common Subsequence}
  \acro{F}{F-Measure}
  \acro{TF-IDF}{Term Frequency-Inverse Document Frequency}
  \acro{BoW}{Bag of Words}
  \acro{LSA}{Latent Semantic Analysis}
  \acro{BERT}{Bidirectional Encoder Representations from Transformers}
  \acro{STraNet}{Structured Trajectory Network via Adversarial Learning}
  \acro{STAT}{Spatial-Temporal Attention Mechanism}
  \acro{POS}{Part-of-Speech}
  \acro{GMMP}{\ac{GCN} Meta-learning with Multi-granularity \ac{POS}}
  \acro{MPII-MD}{Max Plank Institute for Informatics Movie Description Dataset}
  \acro{SVCDV}{Sports Video Captioning Dataset-Volleyball}
  \acro{HRI}{Human Robot Interaction}
  \acro{NLVD}{Natural Language Video Description}
  \acro{AMT}{Amazon Mechanical Turkers}
  \acro{DVS}{Descriptive Video Service}
  \acro{TACoS-Mlevel}{Textually Annotated Cooking Scenes-Multilevel}
  \acro{FN}{Filteration Network}
  \acro{S-MiT}{Spoken Moments Dataset}
  \acro{AD}{Audio Descriptions}
  \acro{LSMDC}{Large Scale Movie Description Challenge
dataset}
\acro{KD}{Knowledge Distillation}
\acro{CRF}{Conditional Random Fields}
\acro{MRF}{Markov Random Field }
\acro{OA-BTG}{Object Aggregation Bidirectional Ttemporal Graph}
\acro{OVC-Net}{Object-Oriented Video Captioning with
Temporal Graph and Detail Enhancement}
\acro{ORG}{Object Relation Graph}
\acro{TRL}{Teacher Recommended Learning}
\acro{D-LSG}{Discriminant Latent Semantic Graph}
\acro{OSTG}{Object-aware Spatio-temporal Graph}
\acro{VLAD}{Vector of Locally Aggregated Descriptors}
\acro{GCN}{Graph Convolution Network}
\acro{ORMF}{Object Relation Graph and Multimodal Feature Fusion}
\acro{CL}{Curriculum Learning}
\acro{AL}{Active Learning}
\acro{SGN}{Semantic Grouping Network}
\acro{MIXER}{Mixed Incremental Cross-Entropy Reinforce}
\acro{RL}{Reinforcement Learning}
\acro{CCN}{Codec Network} 
\acro{ICT}{Inverse Cloze Task}
\acro{WS-DEC}{Weakly supervised dense Event captioning}
\acro{MED}{Multi event detection}
\acro{MEC}{Multi event captioning}
\acro{TSRM}{Temporal Semantic Relation Module}
\acro{CMG}{CrossModal gating}
\acro{VSLA}{Variational stacked local attention network}
\acro{T2I}{Text to Image}
\acro{T2V}{Text to Video}
\acro{VAE}{Variational Auto encoders}
\acro{IRC-GAN}{Introspective Recurrent Convolutional GAN }
\acro{c-GAN}{conditional generative models }
\acro{T2VR}{Text to Video Retrieval}
\acro{ACG}{Anchor--centered Graph}
\acro {AViD}{Anonymized
Videos from Diverse countries} 
\acro{VTW}{Video titles in the wild}
\acro{ST}{Spatio-Temporal}
\acro{TS}{Temporal-Spatial}
\acro{TGN}{Temporal Graph Network}
\acro{RGN}{Relation Graph Network}
\acro{BTG}{Bidirectional Temporal Graph}
\acro{WMD}{Word Mover’s Distance }
\acro{SPICE}{Semantic Propositional Image Captioning Evaluation}
\acro{MT}{Machine Translation}
\acro{IC}{Image Captioning}
\acro{BoW}{Bag of Words}
\acro{EMD}{Earth Mover's Distance}
\acro{G}{Generator}
\acro{D}{Discriminator}
\acro{XE}{Cross entropy}
\end{acronym}


%





\ifCLASSOPTIONcaptionsoff
  \newpage
\fi

\color{black}
\bibliographystyle{IEEEtran}
\tiny
\bibliography{ref.bib}

\end{document}